%% file: main.tex
\documentclass{article}

\PassOptionsToPackage{numbers, compress}{natbib}
\usepackage{biblatex}
\usepackage[subtle]{savetrees}

\usepackage[preprint,nonatbib]{neurips_2022}

\usepackage[T1]{fontenc}    % use 8-bit T1 fonts
\usepackage{hyperref}       % hyperlinks
\usepackage{url}            % simple URL typesetting
\usepackage{booktabs}       % professional-quality tables
\usepackage{amsfonts}       % blackboard math symbols
\usepackage{nicefrac}       % compact symbols for 1/2, etc.
\usepackage{microtype}      % microtypography

\usepackage{float}

\usepackage[dvipsnames]{xcolor}
\usepackage{graphicx}
\usepackage{subcaption}
\usepackage{ragged2e}
\usepackage{booktabs, multirow} % for borders and merged ranges
\usepackage{soul}% for underlines
\usepackage{changepage,threeparttable} % for wide tables
\usepackage{hyperref}
\usepackage{url}

\usepackage{listings}

\lstdefinelanguage{coqpython}{
    keywords={class, def, case, Repair, Lift, module, Module, Theorem, Proof, Record, Lemma, Definition, Abort, Qed, forall, Inductive, Type, Prop, Set, fun, fix, forall, Require, Import, Fixpoint, match, end, with, as, return, struct, Qed, Defined, let, Parameter, Axiom, Patch, Configure, Preprocess,UNROLL[, ],},
    basicstyle=\linespread{0.95}\scriptsize\ttfamily,
    keywordstyle=\color{blue},
    commentstyle=\itshape\rmfamily,
    showstringspaces=false,
    columns=flexible,
    breaklines=true,
    texcl=true,
    mathescape=true,
    tabsize=4,
    stringstyle=\color{brown},
    escapeinside={(@}{@)},
    morecomment=[n][\itshape\rmfamily\color{violet}]{(*}{*)}
}
\title{Can Transformers Learn to Solve Problems Recursively?}
\author{%
Shizhuo Dylan Zhang$^{1}$ \quad Curt Tigges$^{2}$ \quad Stella Biderman$^{2,3}$ \quad Maxim Raginsky$^1$ \quad Talia Ringer$^1$ \\
$^1$University of Illinois Urbana-Champaign \quad $^2$EleutherAI \quad $^3$Booz Allen Hamilton\\
\texttt{\{shizhuo2,maxim,tringer\}@illinois.edu}\\
\texttt{\{curt,stella\}@eleuther.ai}
}
\begin{document}

\maketitle

\begin{abstract}
Neural networks have in recent years shown promise for helping software engineers write programs and even formally verify them. While semantic information plays a crucial part in these processes, it remains unclear to what degree popular neural architectures like transformers are capable of modeling that information.

This paper examines the behavior of neural networks learning algorithms relevant to programs and formal verification proofs through the lens of mechanistic interpretability, focusing in particular on structural recursion. Structural recursion is at the heart of tasks on which symbolic tools currently outperform neural models, like inferring semantic relations between datatypes and emulating program behavior. 

We evaluate the ability of transformer models to learn to emulate the behavior of structurally recursive functions from input-output examples. Our evaluation includes empirical and conceptual analyses of the limitations and capabilities of transformer models 
in approximating these functions, as well as reconstructions of the ``shortcut'' algorithms the model learns. By reconstructing these algorithms, we are able to  \textit{correctly predict} 91\% of failure cases for one of the approximated functions. Our work provides a new foundation for understanding the behavior of neural networks that fail to solve the very tasks they are trained for.
\end{abstract}

%\newpage
\input{intro}
\input{overview}

\input{asm}
\input{evaluation}
\input{related}
\input{conclusions}
% \section*{References}
\printbibliography
\newpage
\appendix

\section{Experimental Details}

This appendix elaborates on details of our experiments, including
training (Appendix~\ref{app:training}) and the
tasks we choose (Appendix~\ref{app:tasks}).

\subsection{Training Details}
\label{app:training}

To examine the performance of our models for the experiments in Section~\ref{sec:evaluation}, we computed accuracy based on exact match of complete sequences under a greedy decoding set-up, i.e.:

\begin{equation}
    Acc = 
    \frac{\sum_{i\in \mathcal{D}_{test}}\mathbf{}{1}(\hat{Y_i} == Y_i)}{|\mathcal{D}_{test}|}
\end{equation}
In order to address the inherent limitation of transformers in generating longer sequences beyond their training exposure, and to explore whether the learned algorithms in transformers can extrapolate, we implemented a straightforward approach inspired by previous research~\cite{dehghani2019universal}. This involved introducing a simple mechanism where we randomly adjust the positional encoding by adding random lengths of padding to the inputs. By doing so, we challenged the model to generate both the actual sequence and the corresponding padding accurately. 

We experimented with encoder-decoder transformer models with 2 encoder layers, 2 decoder layers, a hidden dimension of 128, and 2 heads. For simplicity of analysis, we used sinusoidal positional encoding and greedy decoding. 

For binary successor tasks, we train the model on binary string pairs from 1 to $n$, where $n$ is the number of examples we would like to use for training. For example, the maximum number we used for training is 131072, which corresponds to a 17-bit string. One can observe that the distribution of depths is non-uniform, as it aligns with the natural occurrence of different depths.

For tree traversal experiments, by default, we split the train and test sets by tree structures, where we wish to understand the model's capability of dealing with unseen topologies of the tree. 

\subsection{More Task Details}
\label{app:tasks}

We evaluated model performance on variants of the two tasks
from Section~\ref{sec:overview}. This appendix elaborates on those tasks.

\subsubsection{Binary Successor}
\label{app:bin}

The binary successor function from Section~\ref{sec:bin} is significant in that it requires no explicit knowledge of any other number format, and yet it entirely grounds binary positive numbers in the semantics of other number formats. 
Once we have \lstinline{s}, for example, we can align
unary and binary positive natural numbers. In fact, it turns out that it is \emph{impossible} to determine how to, in general, move
functions and proofs between these two number formats without
discovering something that behaves like \lstinline{s} along the way~\cite{ringer2021pldi}. And yet the relation is not completely obvious, since the structures of unary and binary natural numbers are different from one another~\cite{ringer2021proof}. This makes \lstinline{s} an apt choice for our  
first task, in that it gets us closer to discovering the semantic relations current neural tools struggle to infer.

From our six input-output examples and our priors described in Section~\ref{sec:bin}, we can
infer the shape of the function we are synthesizing:

\begin{lstlisting}
  Fixpoint s n :=
    match n with (* break into cases *)
    | 01 => ?? (* if n is one, return ?? *)
    | XO b => ?? (* if the LSB is zero, ?? *)
    | X1 b => ?? (* otherwise, ?? *)
    end.
\end{lstlisting}
To fill in those \lstinline{??} holes, we can look at our six input/output examples. The first example already fills one hole:

\begin{lstlisting}
    | 01 => XO 01 (* if n is one, return two *)
\end{lstlisting}
For the other two holes, we can fill in the behavior on the least significant bit first:

\begin{lstlisting}
    | XO b => X1 ?? (* if the LSB is zero, flip ?? *)
    | X1 b => XO ?? (* otherwise, ?? and shift *)
\end{lstlisting}
The first remaining hole is satisfied on all examples by the identity function on \lstinline{b}.
\iffalse
\begin{lstlisting}
      | 01 => XO 01
      | XO b => X1 b
      | X1 b => XO ??
\end{lstlisting}
\fi
The second remaining hole, however, must recurse. Filling in those holes
accordingly gives us the definition of \lstinline{s} from Section~\ref{sec:bin}.

\subsubsection{Tree Traversal}
\label{app:tree}

We can represent the binary trees from Section~\ref{sec:tree} inductively:

\begin{lstlisting}
  Inductive tree :=
  | Leaf (* the base case is an empty leaf *)
  | Branch (v : char) (l r : tree). (* the inductive case is a branch with a character value and two subtrees *)
\end{lstlisting}
We can then write an inorder traversal this way:

\begin{lstlisting}
  Fixpoint inorder t :=
    match t with
    | Leaf => []
    | Branch val l r => (inorder l) ++ [v] ++ (inorder r)
    end.
\end{lstlisting}

The reduction alluded to in Section~\ref{sec:tree}
selects the appropriate case of \lstinline{inorder} and takes one step at a time. That is,
to compute:

\begin{lstlisting}
  inorder (Branch 'a' (Branch 'c' Leaf Leaf) (Branch 't' Leaf Leaf))
\end{lstlisting}
it can reduce one step at a time by selecting the \lstinline{Branch} case and substituting in the right values:

\begin{lstlisting}
  (inorder (Branch 'c' Leaf Leaf)) ++ ['a'] ++ (inorder (Branch 't' Leaf Leaf))
\end{lstlisting}
It can reduce \lstinline{inorder} on the left and right subtrees:

\begin{lstlisting}
  ((inorder Leaf) ++ ['c'] ++ (inorder Leaf)) ++ ['a'] ++ ((inorder Leaf) ++ ['t'] ++ (inorder Leaf))
\end{lstlisting}
Finally, it can reduce the \lstinline{Leaf} cases:

\begin{lstlisting}
  ([] ++ ['c'] ++ []) ++ ['a'] ++ ([] ++ ['t'] ++ [])
\end{lstlisting}
to get \lstinline{['c'; 'a'; 't']}.

\section{How Effectively did the Transformer Simulate a Recursive ASM?}
\label{app:recursive}

\begin{table}\centering
\scriptsize
\begin{tabular}{lrrrrr}\toprule
&\multicolumn{2}{c}{Binary Classification} &\multicolumn{2}{c}{Recursion Depth Classification} \\\cmidrule{2-5}
&Train Acc &Test Acc &Train Acc &Test Acc \\\midrule
Nat C=1 &0.68 &0.66 &0.21 &0.14 \\
Nat C=0.1 &0.62 &0.52 &0.22 &0.13 \\
Rev C=1 &0.67 &0.56 &0.20 &0.13 \\
Rev C=0.1 &0.64 &0.51 &0.21 &0.14 \\
\bottomrule
\end{tabular}
\caption{Sequence pattern-matching accuracy table.}
\label{tab:seq_pattern}
\end{table}
\begin{table}\centering
\scriptsize
\begin{tabular}{lrrrrr}\toprule
&\multicolumn{2}{c}{Encoder Embedding} &\multicolumn{2}{c}{Decoder Embedding} \\\cmidrule{2-5}
&Train Acc &Test Acc &Train Acc &Test Acc \\\midrule
Nat C=1 &0.33 &0.25 &0.68 &0.54 \\
Nat C=0.1 &0.62 &0.52 &0.62 &0.53 \\
Rev C=1 &0.32 &0.24 &0.47 &0.37 \\
Rev C=0.1 &0.33 &0.23 &0.44 &0.32 \\
\bottomrule
\end{tabular}
\caption{Recursion-depth recognition accuracy table.}
\label{tab:depth_pattern}
\end{table}

\begin{figure*}[!tb]
     \begin{subfigure}[b]{.48\columnwidth}
         \centering
        \includegraphics[width=\columnwidth]{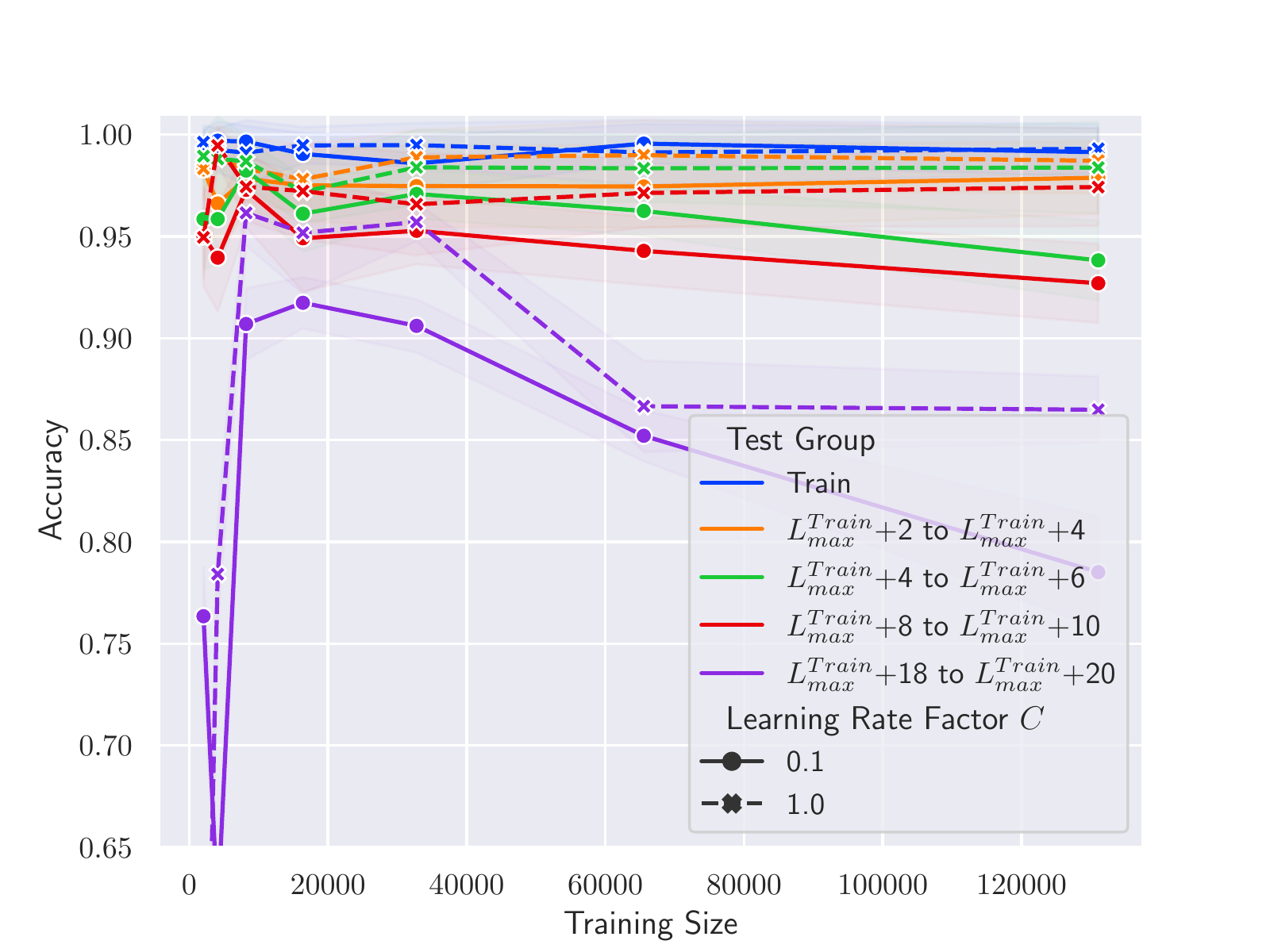}
         \caption{Basic-Depth3}
         \label{fig:basic_depth3}
     \end{subfigure}
     \begin{subfigure}[b]{.48\columnwidth}
         \centering
        \includegraphics[width=\columnwidth]{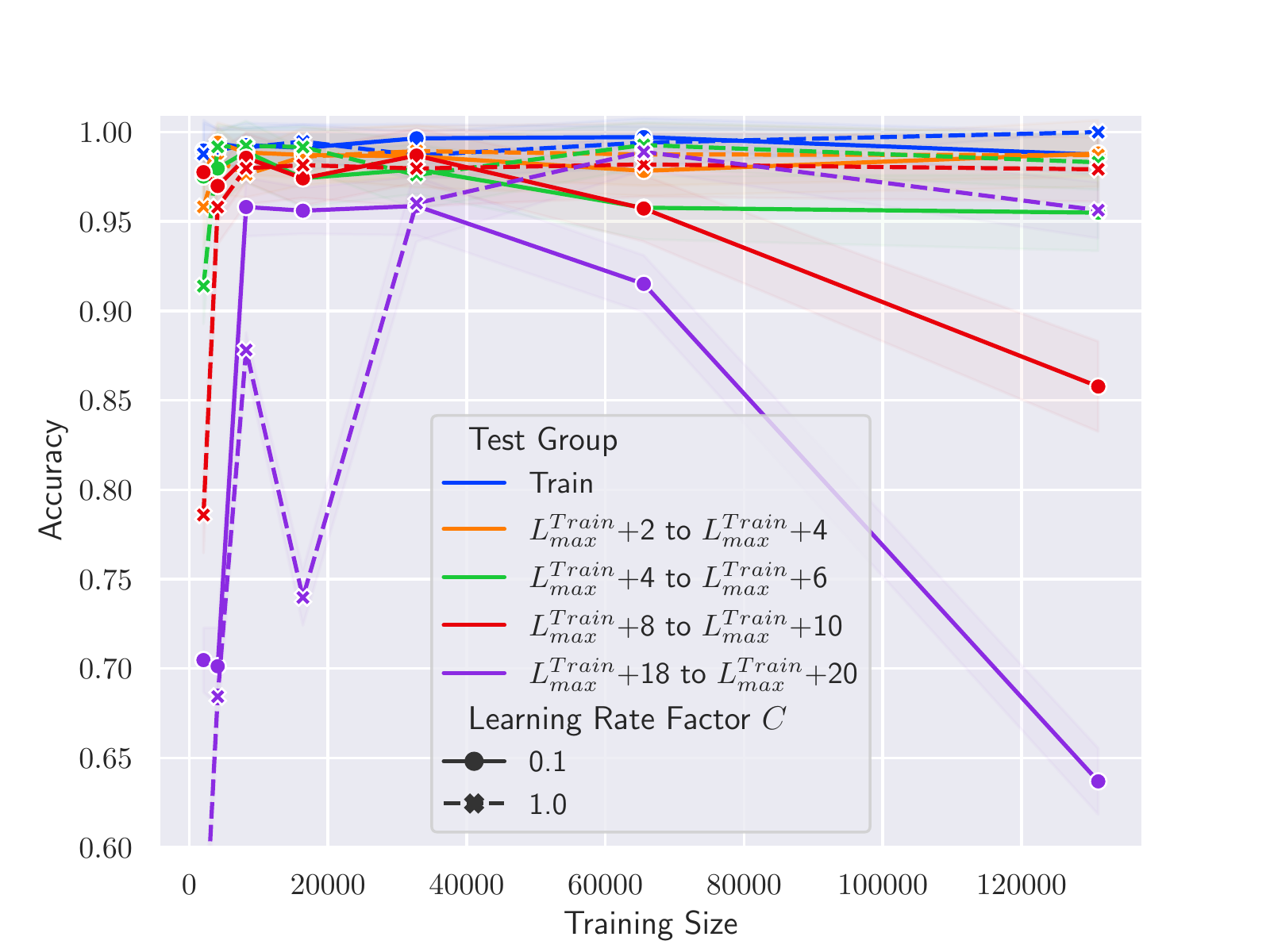}
         \caption{Basic-Depth6}
         \label{fig:basic_depth6}
     \end{subfigure}

     \begin{subfigure}[b]{.48\columnwidth}
         \centering
         \includegraphics[width=\columnwidth]{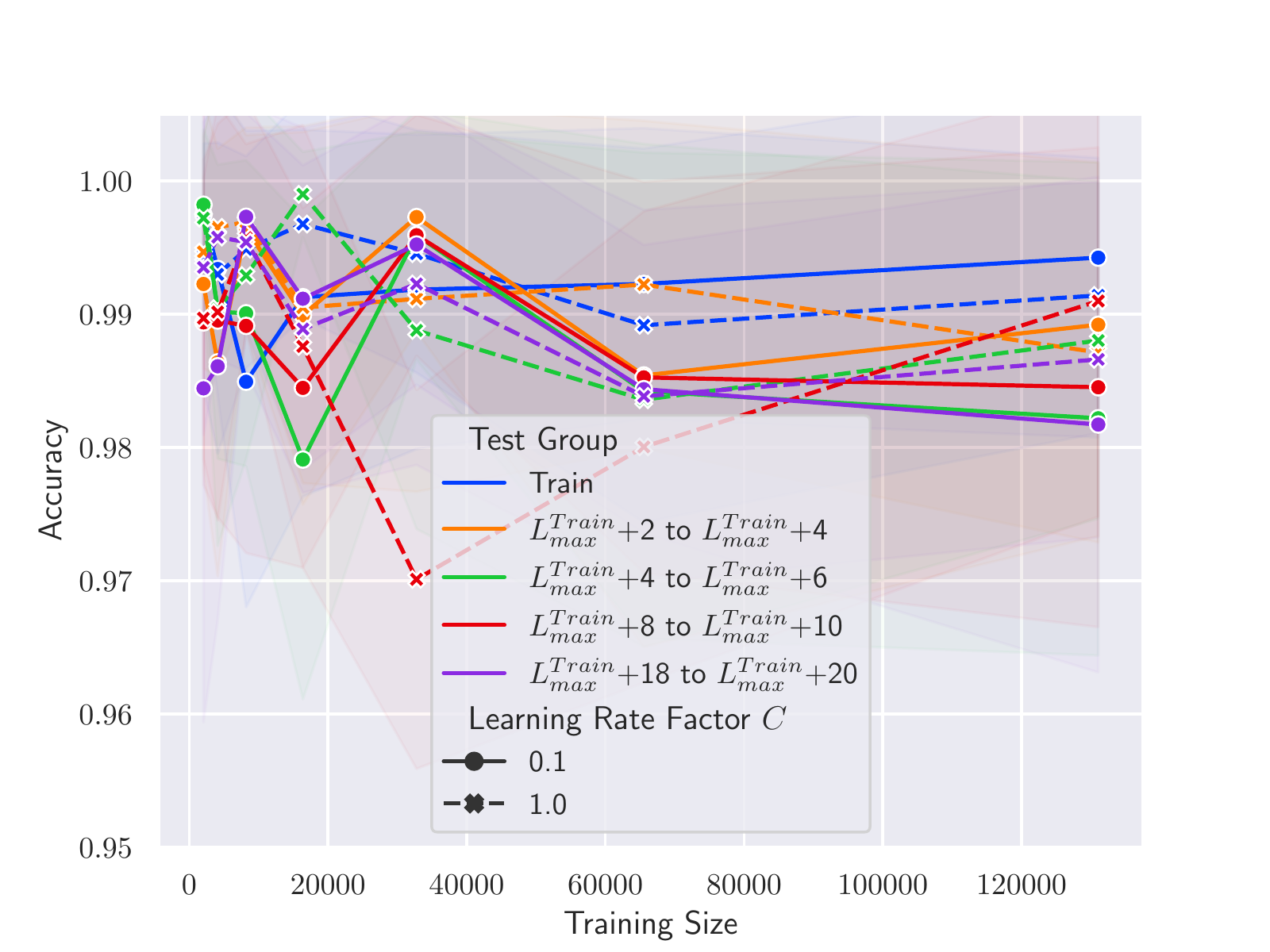}
         \caption{Reversed-Depth3}
         \label{fig:rev_depth3}
     \end{subfigure}
     \begin{subfigure}[b]{.48\columnwidth}
         \centering
         \includegraphics[width=\columnwidth]{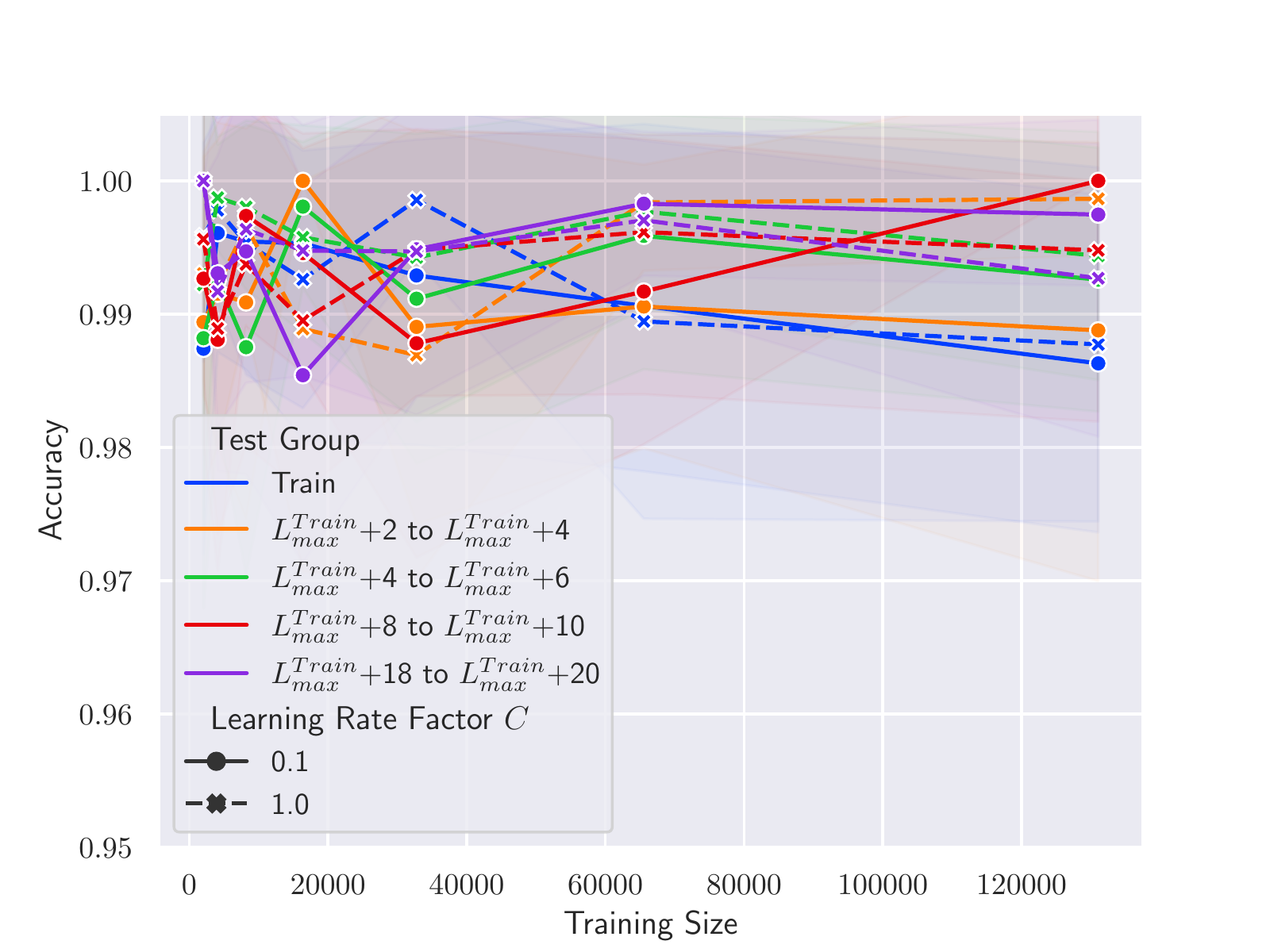}
         \caption{Reversed-Depth6}
         \label{fig:rev_depth6}
     \end{subfigure}
    \caption{Model performance when trained on data with more constrained recursion depths.}
    \label{fig:more_depth}
\end{figure*}

As discussed in Section~\ref{sec:model}, one of the primary objectives of our study was to investigate whether the model implicitly implemented a recursive algorithm by utilizing pattern matching on the most significant bit. To accomplish this, we conducted an initial analysis to examine the embeddings of the encoder and decoder for pattern-matching and recursion-depth-related information.
(Figure~\ref{fig:more_depth} shows model performance---first alluded to in Section~\ref{sec:evalbinsuc}---when models were trained on data with more constrained recursion depths.)

Our ideal outcome was to identify evidence supporting the implementation of a recursive Abstract State Machine (ASM) employing a pattern-matching scheme.
%, as outlined in Section~\ref{sec:bin}.
Alternatively, we expected that a sequentialized algorithm for the recursive task might exhibit indications of tracking the depth of recursion.

To delve deeper into this matter, we designed a straightforward experiment. We sought to train a simple linear classifier capable of distinguishing between cases such as \lstinline{XO bb} and \lstinline{X1 bb} based on the model's internal representations. We computed the average encoder output across the sequence and trained a linear classifier on the fixed encoder representations for both binary classification and recursion-depth classification. The outcomes of these experiments are presented in Table~\ref{tab:seq_pattern}. Regrettably, our attempts to train a classifier that could detect the distributional distinctions in the embeddings between the two cases proved unsuccessful.

Additionally, we aimed to investigate the extent to which the embeddings of individual tokens contained information regarding the recursion depth. To achieve this, we trained a classifier to recognize the recursion depth of each token using both the encoder embeddings and the output embedding of the final decoder layer. The results, presented in Table~\ref{tab:depth_pattern}, demonstrate that the encoder embeddings do not possess positional information, while the decoder embeddings do exhibit a certain degree of positional information in the natural direction. However, we did not observe this phenomenon when considering the embeddings in reverse order. 

In conclusion, our analysis suggests that the model does not explicitly exhibit evidence of implementing a recursive algorithm based on pattern matching or tracking the recursion depth. Moreover, the encoder embeddings lack positional information, while the decoder embeddings exhibit some level of positional information in the natural direction.

% \begin{figure}
%     \centering
%      \begin{subfigure}[tb]{.35\columnwidth}
%          \centering
%          \includegraphics[width=\columnwidth]{recursive_func_transformer_neurips/plots/traversal/traversal_unroll1/layer_0_head_1.pdf}
%          \caption{First encoder layer for traversal 1-step reduction}
%          \label{fig:perturb_nat_orig}
%      \end{subfigure}
%      % \hspace{30mm}
%      \begin{subfigure}[tb]{.35\columnwidth}
%          \centering
%          \includegraphics[width=\columnwidth]{recursive_func_transformer_neurips/plots/attn_bin_rev_unroll/128_2heads_lr1_dec_self_layer0_head0.pdf}
%          \caption{Reverse order, first layer.}
%          \label{fig:perturb_nat_perturbed}
%      \end{subfigure}
%     \caption{Decoder self-attention for one-step unrolling of binary successor task.}
%     \label{fig:unroll}
% \end{figure}

% \section{Depth Generalization}

\section{Algorithm Reconstruction for Binary Successor}
\label{app:reconstruct}

We summarized our results for the binary successor task in Section~\ref{sec:evalbinsuc}. Here, we (1) present more evidence for our findings in the natural order, and (2) provide pseudocode for the reconstructed algorithms in both the natural and reverse orders.

\begin{figure}
     %\begin{subfigure}[t]{.31\textwidth}
         \centering
         \includegraphics[width=0.7\columnwidth]{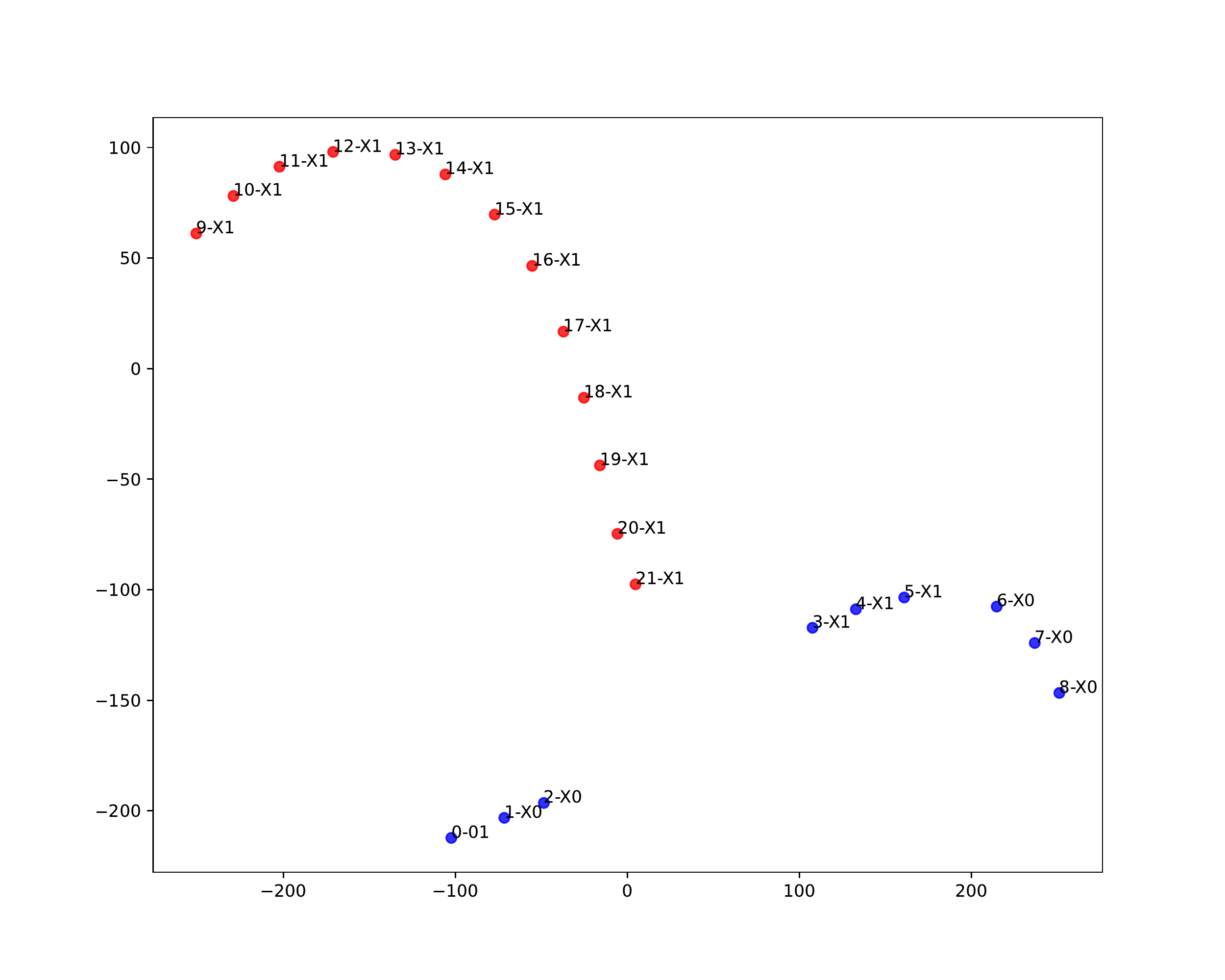}
         \vspace{-0.5cm}
         \caption{T-SNE plot for the encoder embedding.}
         \label{fig:enc_emb}
     \end{figure}

\begin{figure}
     \centering
     % \hspace{3mm}
     \includegraphics[width=0.7\columnwidth]{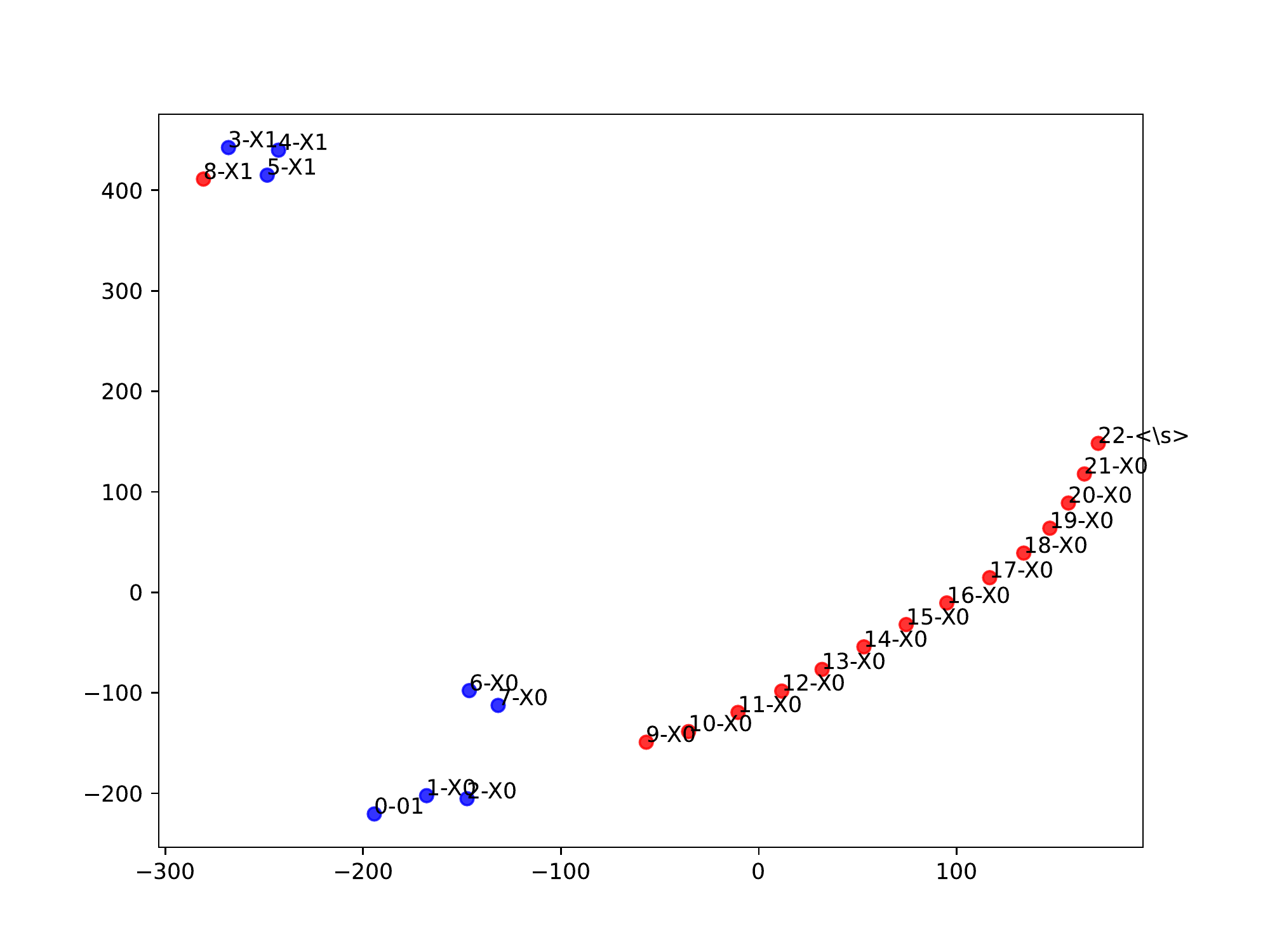}
         \caption{T-SNE plot for the decoder layer 0 self-attention output (before cross-attention). }
         \label{fig:dec_layer0}
     \end{figure}

\begin{figure}
         \centering
         \includegraphics[width=0.7\columnwidth]{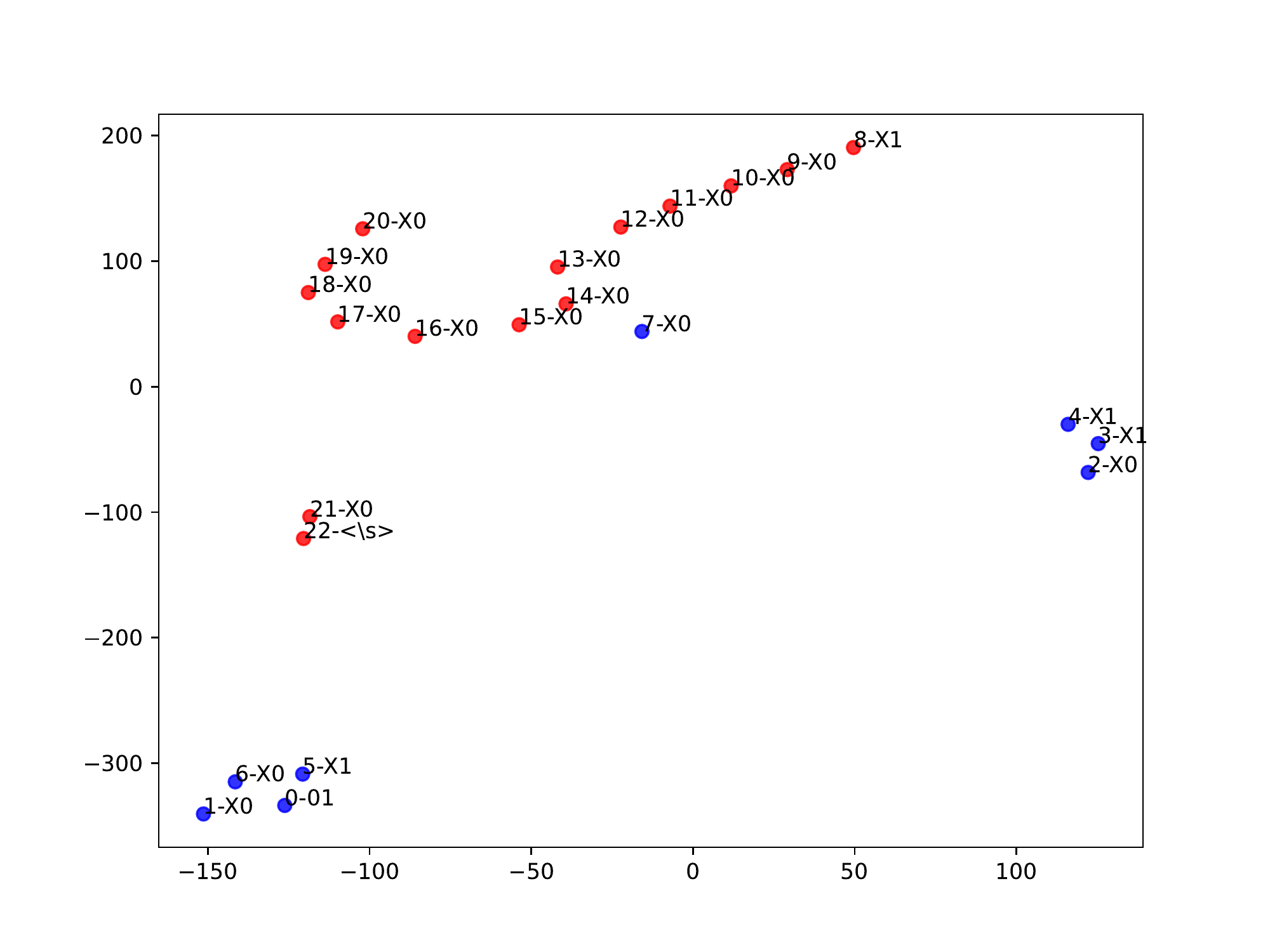}
         \caption{T-SNE plot for the decoder layer 1 self-attention output.}
         \label{fig:dec_layer1}
     \end{figure}

% Figure~\ref{fig:enc_nat}---originally referred to in Section~\ref{sec:evalbinsuc}---shows that the encoder plays a part in modeling semantics by helping differentiate between cases.
Figures~\ref{fig:enc_emb}, \ref{fig:dec_layer0}, and~\ref{fig:dec_layer1} show
T-SNE plots of the internal representation for the respective layers, with recursive segments in red and non-recursive segments in blue.
The encoder (Figure~\ref{fig:enc_emb}) distinguishes between recursive and non-recursive segments. The first decoder layer (Figure~\ref{fig:dec_layer0}) recognizes symbols. Finally, the cross-attention between encoder and decoder (Figure~\ref{fig:dec_layer1}) injects instructions from the encoder about the start of recursion and end of sentence.

% \begin{figure}
%          \centering
%          \includegraphics[width=.6\columnwidth]{recursive_func_transformer_neurips/plots/attn_bin_nat/bin_basic_enc_self_attn.pdf}
%          \vspace{-0.5cm}
%          \caption{Encoder self-attention in the natural order.}
%          \label{fig:enc_nat}
%         % \vspace{-3mm}
% \end{figure}


\begin{figure}
         %\centering
         \includegraphics[width=.8\columnwidth]{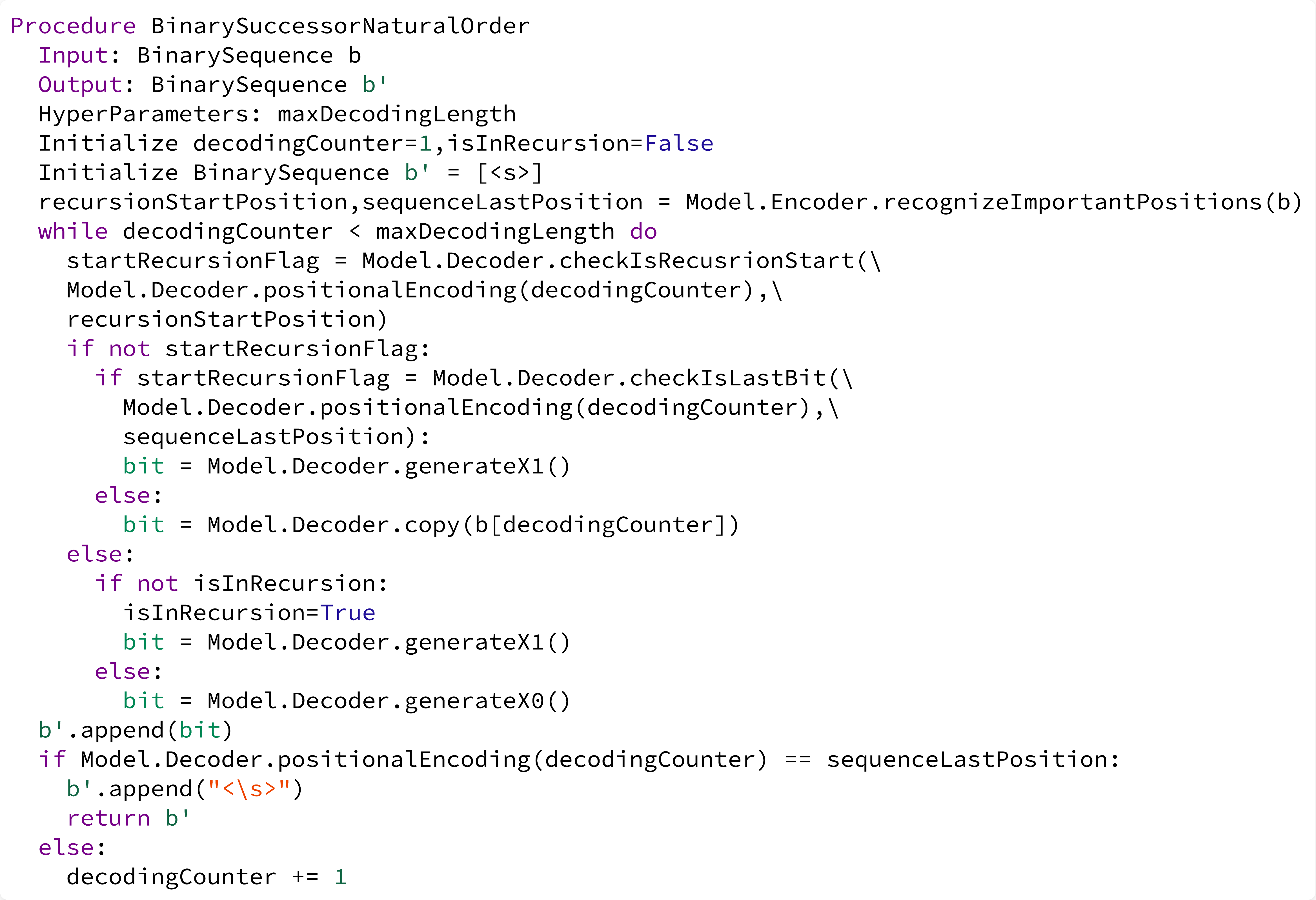}
         \caption{Reconstructed Algorithm: Natural Order}
         \label{fig:algo_nat}
 \end{figure}

\begin{figure}
         %\centering
         \includegraphics[width=.65\columnwidth]{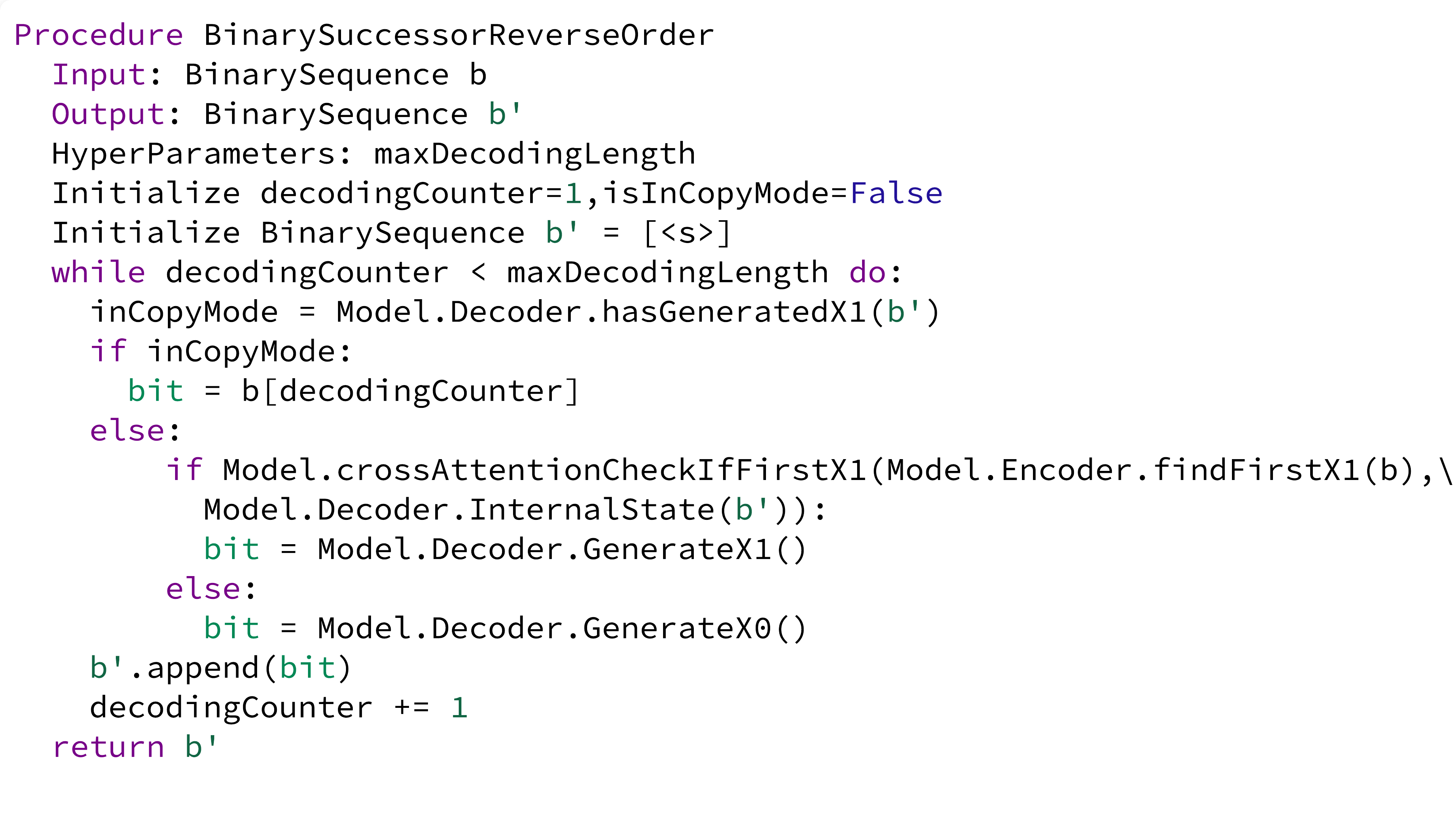}
         \caption{Reconstructed Algorithm: Reverse Order}
         \label{fig:algo_rev}
         %Attention map of error cases in both feeding orders. The reversed-order attention no longer follows the normal pattern because the shortcut is unavailable. For natural order, it still cheats as usual but the cheating solution is wrong in this case. }
\end{figure}

The pseudocode for the reconstructed algorithms that we described informally in Section~\ref{sec:evalbinsuc} can be found in Figures~\ref{fig:algo_nat} and~\ref{fig:algo_rev}, respectively.

     % \hspace{3mm}

\iffalse
\begin{figure}
     \begin{subfigure}[t]{.45\textwidth}
         \centering
         \includegraphics[width=\columnwidth]{recursive_func_transformer_neurips/plots/error/bin_rev_dec_self_attn_error.pdf}
         \caption{Reversed Order}
         \label{fig:err_rev}
     \end{subfigure}
     % \hspace{3mm}
          \begin{subfigure}[t]{.45\columnwidth}
         \centering
         \includegraphics[width=\columnwidth]{recursive_func_transformer_neurips/plots/error/bin_basic_dec_self_attn_error2.pdf}
         \caption{Natural Order}
         \label{fig:err_basic}
     \end{subfigure}
         \caption{Recursion head attention on error cases.}%Attention map of error cases in both feeding orders. The reversed-order attention no longer follows the normal pattern because the shortcut is unavailable. For natural order, it still cheats as usual but the cheating solution is wrong in this case. }

\end{figure}
\fi

\newpage
\section{Algorithm Reconstruction for Tree Traversal}
\label{app:curt}

Here we describe our reconstruction of the algorithms for the tree traversals initially described in Section~\ref{sec:evaltree}.

\subsection{Approach}
\label{sec:tree_approach}
Our overall approach to reverse-engineering the tranformer's behavior in the tree traversal task involved the following:
\begin{enumerate}
    \item decomposing the multi-step inference task into key subtasks,
    \item conducting activation patching in order to determine what parts of the network were most important to accomplishing the subtask, and
    \item examining the attention pattern behavior for the most significant attention heads, as well as the patterns of earlier heads that likely contributed to the given head's behavior.
\end{enumerate}

\subsection{Decomposition of Tasks}
We decomposed overall tree traversal task performance into seven key subtasks and examined the behavior of each. Not all applied to each case (e.g., some were relevant for only preorder or inorder traversal), but overall these subtasks included:
\begin{enumerate}
    \item copy initiation,
    \item midpoint of left tree copy,
    \item end of left tree copy,
    \item insertion of root node from initial sequence,
    \item insertion of the UNROLL symbol,
    \item resumption of copy after insertion, and
    \item  end of sequence.
\end{enumerate}

% 1) copy initiation, 2) midpoint of left tree copy, 3) end of left tree copy, 4) insertion of root node from initial sequence, 5) insertion of the UNROLL symbol, 6) resumption of copy after insertion, and 7) end of sequence. 

\subsection{Activation Patching}
Using counterfactual patching experiments inspired by prior work~\cite{meng2023locating}, we identified the most important attention heads in the transformer's final output. To summarize the technique, we replaced neural activations from a forward pass where the model produced counterfactual answer B with activations from a forward pass when prompted to produce answer A. The activations were patched during the forward pass, and their effects propagated to later layers.

This process was performed one attention head at a time, and the resulting difference in logits between correct answer A and incorrect answer B was measured. The degree to which the patch restored the original logit difference was taken as a measure of significance for the corresponding head. Typically, cross-attention heads most significantly affected the final result, but patching decoder self-attention also sometimes yielded significant, task-dependent changes in the logit difference.

We utilized this technique to determine which attention heads were most important to the final output of the transformer. This did not reveal the full circuit used to perform the task, but did provide suggestive, causal evidence of what the transformer is doing. After identifying these key components of the model, we focused our investigation on the roles and composition behavior of these attention heads, leaving a comprehensive analysis of the entire circuit for future research.

\subsection{Reconstructed Algorithms}

Based on the attention patterns at various steps of the traversal task, our encoder-decoder architecture appears to follow a simple, interpretable set of rules. All traversals begin by providing the complete parenthesized tree sequence to the encoder, and then providing the decoder with a starting <s> token. The remaining portion of the output was produced autoregressively.

\textbf{In-order Partial Reduction Traversal (Example: Reduce Twice, Depth 2-3)}
\begin{enumerate}
\item The network always starts by inserting an "UNROLL[" symbol.
\item Cross-attention attends to the first parenthesized node and writes this token into the final token position.
\item This process continues with the model attending to target tokens and parentheses in sequence until the first parenthesis is closed.
\item The model inserts a closing "]" bracket. The left subtree is now complete.
\item The model attends to the token immediately preceding the closed first parenthesis, and writes this token out, as shown in ~\ref{fig:traverse3}.
\item The model then performs the same copy operation with the right subtree.
\end{enumerate}

\begin{figure*}
\centering
\includegraphics[width=.6\columnwidth]{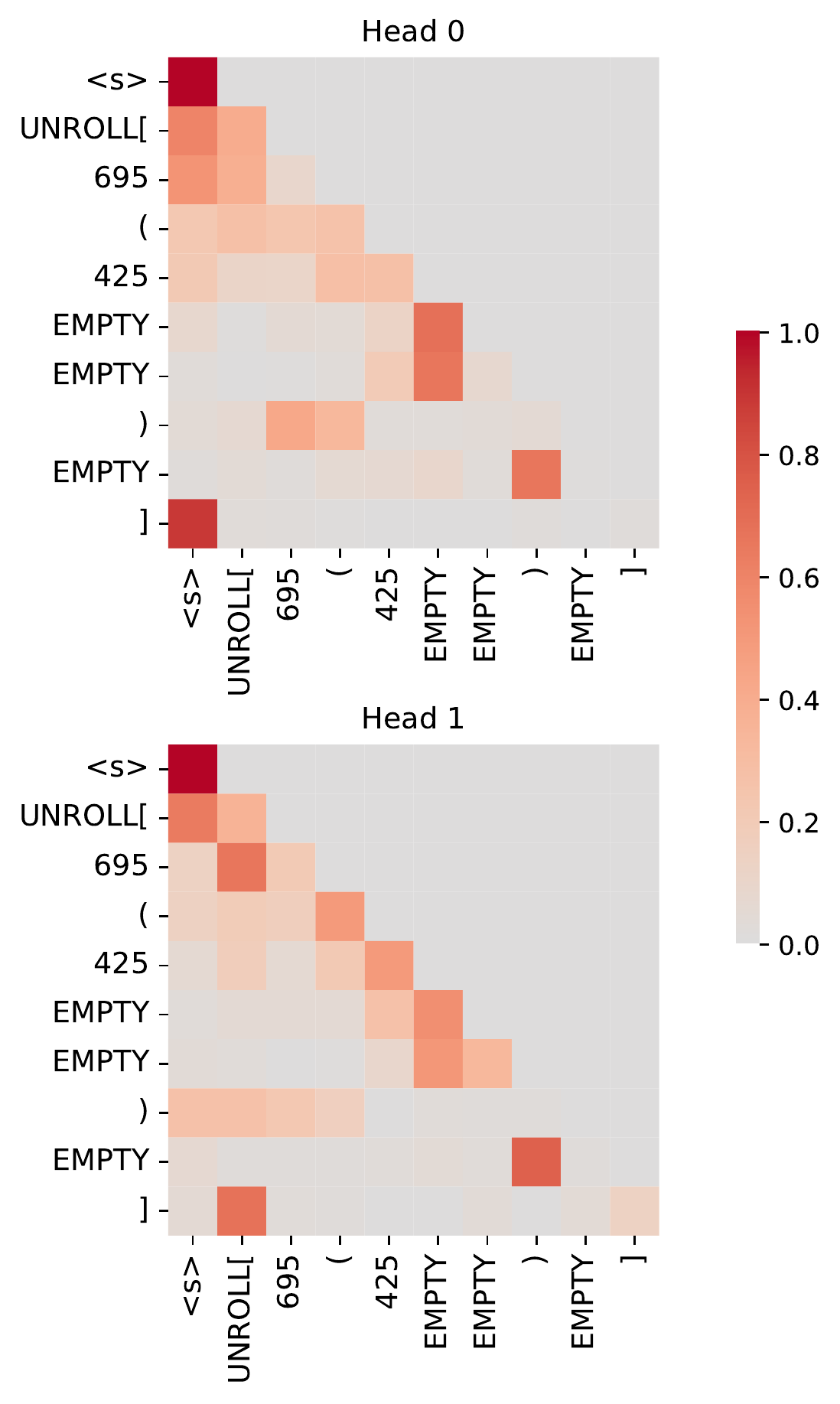}
\caption{Layer 1 decoder self-attention during in-order partial reduction at a higher depth. The model's next token is a non-consecutive node ID (as per in-order traversal, a parent node which occurs before its two children should be copied to the position after its left child and before its right child), which should only be copied once a bracket statement containing an unreduced part of the tree has been copied. Here, the most significant heads attend to the beginning and end of the bracket statement.}
\label{fig:traverse4}
\end{figure*}

For higher depths, the model performs essentially the same operation but additionally uses decoder self-attention to attend to the beginning and end of the relevant bracket level as per Figure~\ref{fig:traverse4}. The significance of this attention pattern seems to suggest that it helps the network track which parent node to copy over. For shallower trees, the model only needs to copy over the root node at the beginning of the encoder sequence, so it doesn't need to track multiple parent nodes; however, for deeper trees, the model needs some indication of where the appropriate parent node can be found. Potentially, the model could use the start of the "UNROLL[]" sequence as a way to point to the token immediately \textit{after} the appropriate parent node, which could then be used as a query to find the node prior to it in the original encoder sequence.

\textbf{Pre-order Partial Reduction Traversal (Example: Reduce Twice, Depth 2-3)}
\begin{enumerate}
\item The model begins by copying node numbers without parentheses.
\item Depending on the depth, the model is trained to output an "UNROLL[" token once two reductions have occurred. When writing out this token, cross-attention patterns of the key heads indicate that the model is attending to A). the parenthesis immediately prior to the node after which the "UNROLL[" token is to be inserted, and B). the following opening parenthesis, and C). the final closing parenthesis of the subsequence that will be copied into the "UNROLL[...]" statement. This seems to suggest that the model is keeping track of the parenthetical depth when deciding where to insert the "UNROLL[...]" wrapper.
\item In parts of the output sequence into which an EMPTY token would normally be inserted with a simple copy operation, decoder self-attention is more significant to the final logit difference. Looking at attention patterns once again, we see that the model is attending to the preceding "UNROLL[" token, if present, and is likely using it as a basis for whether to output this token or not. (Subsequences inside the wrapper should contain "EMPTY" tokens, while those outside should not.)
\item The model completes the sequence with the end token ("\lstinline{<\s>}"). It is less clear what the model is doing here, but the most significant contributing heads attend to a combination of the final brackets and EMPTY tokens as well as the \lstinline{<\s>} token of the original encoder output.
\end{enumerate}

\begin{figure}
              \begin{subfigure}[b]{\columnwidth}
         \centering
         \includegraphics[width=\columnwidth]{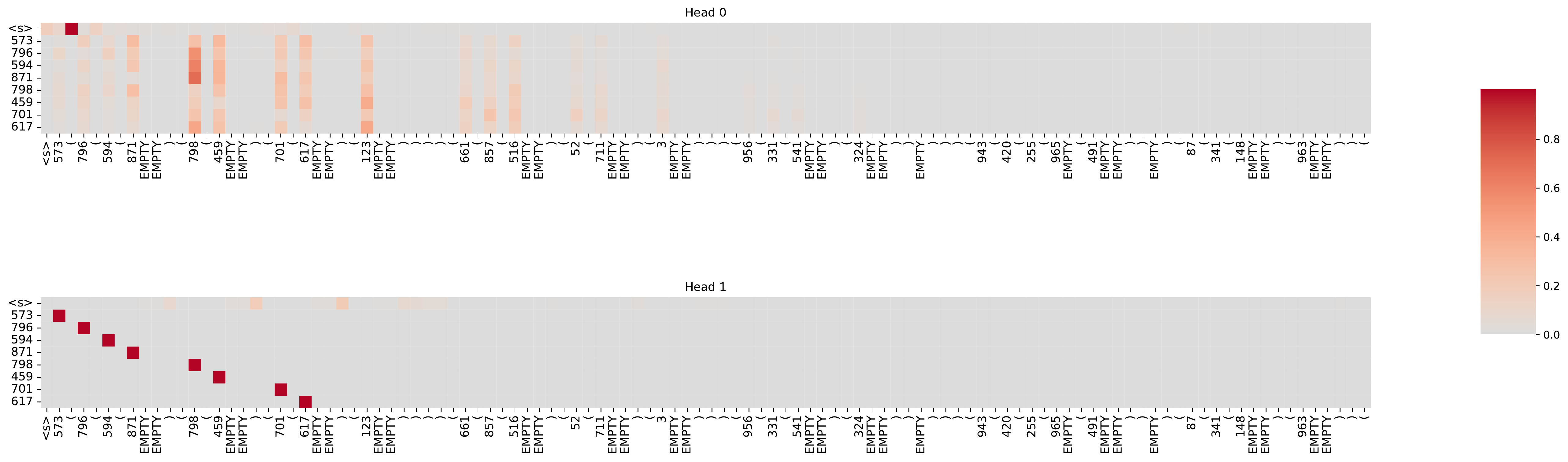}
         \caption{Layer 0 cross-attention. The model attends only to nodes, not to parentheses or EMPTY tokens.}
         \label{fig:traverse5}
     \end{subfigure}
          \begin{subfigure}[b]{\columnwidth}
         \centering
         \includegraphics[width=\columnwidth]{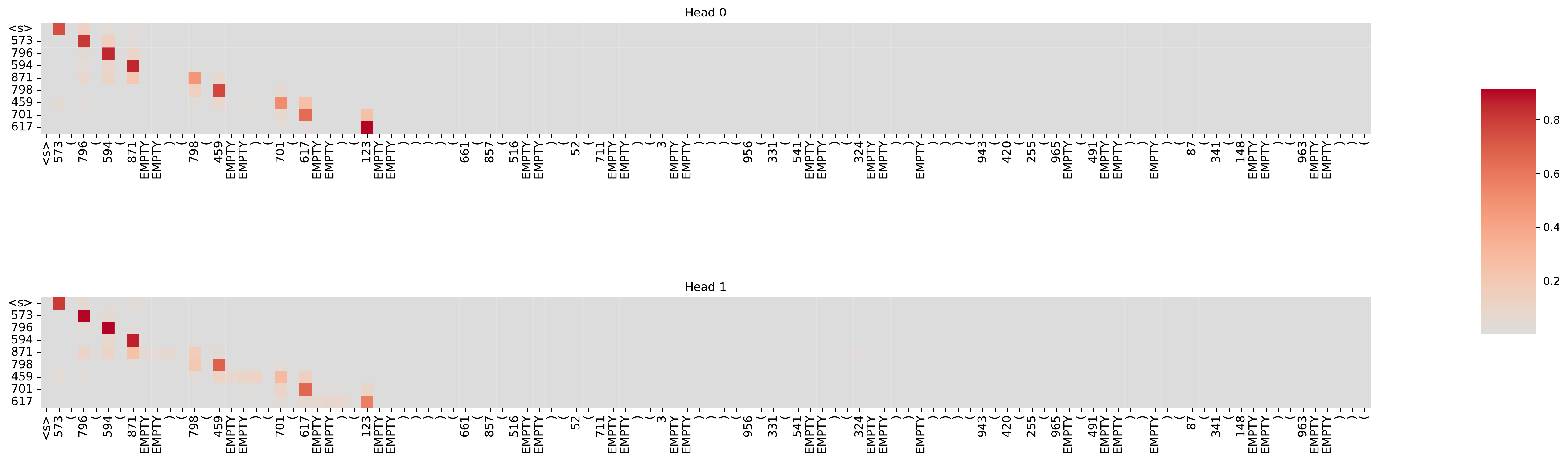}
         \caption{Layer 1 cross-attention. Attention is consolidated on the next node to be copied, skipping all non-node items.}
         \label{fig:traverse6}
     \end{subfigure}
     \caption{Cross-attention on the preorder full traversal task.}
     \label{fig:traverse5and6}
\end{figure}

\textbf{Preorder Full Traversal}
\begin{enumerate}
    \item The model conducts a straightforward copy operation of all node IDs in the encoder sequence, omitting brackets, EMPTY tokens, and parens.
    \item Lower-level heads seem to attend to "nodes in general", with attention being paid to many different nodes both ahead of and behind the current output position, ignoring all non-node tokens, as seen in Figure~\ref{fig:traverse5}.
    \item Later heads attend more exclusively to the next token to be copied from the encoder sequence, as seen in Figure ~\ref{fig:traverse6}.
\end{enumerate}

\subsection{Separation of Tasks in Attention Heads}
\begin{figure*}
     \centering
     \begin{subfigure}[b]{\columnwidth}
         \centering
         \includegraphics[width=\columnwidth]{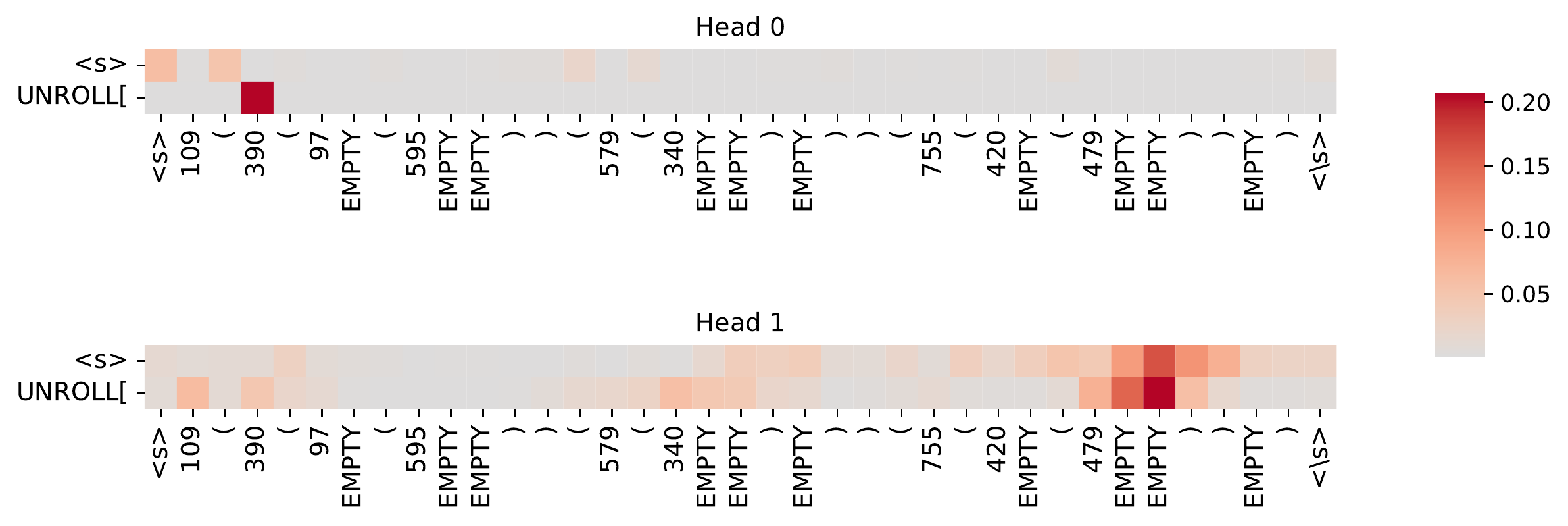}
         \caption{Layer 0 cross-attention at the start of sub-tree to be unrolled. The model attends to distant parentheses and EMPTY tokens as well as the next tokens to be copied.}
         \label{fig:traverse1}
     \end{subfigure}
          \begin{subfigure}[b]{\columnwidth}
         \centering
         \includegraphics[width=\columnwidth]{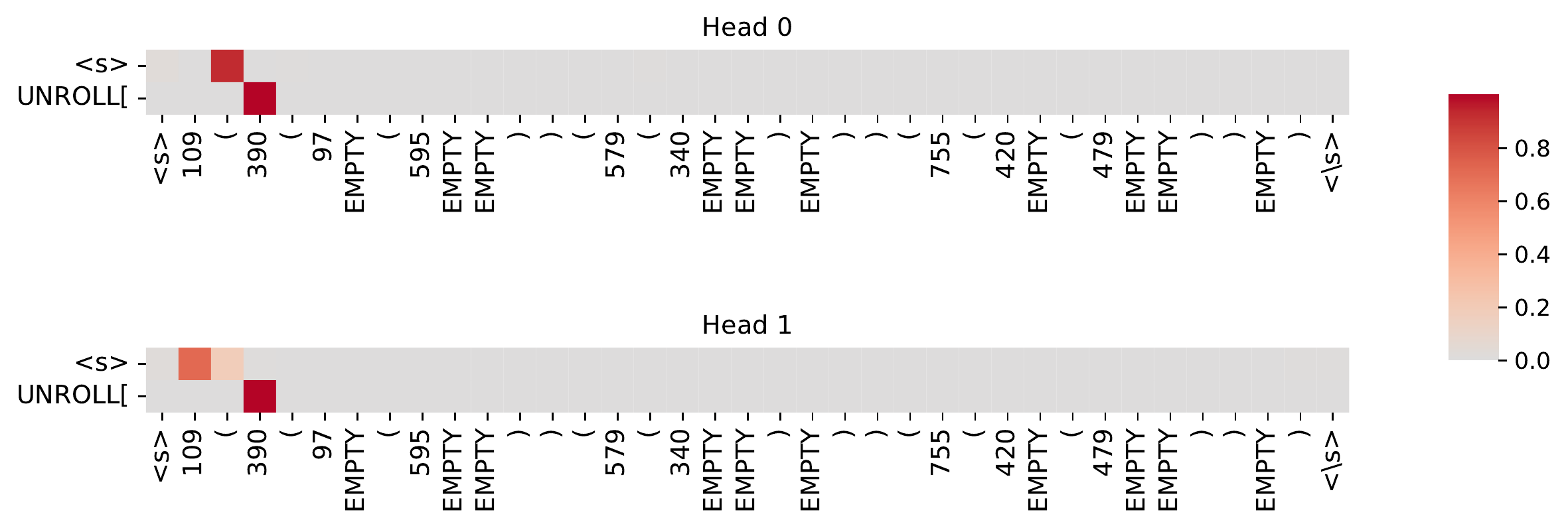}
         \caption{Layer 1 cross-attention at the start of sub-tree to be unrolled. At this point the model has consolidated its attention primarily on the token to be copied.}
         \label{fig:traverse2}
     \end{subfigure}
          \begin{subfigure}[b]{\columnwidth}
         \centering
         \includegraphics[width=.6\columnwidth]{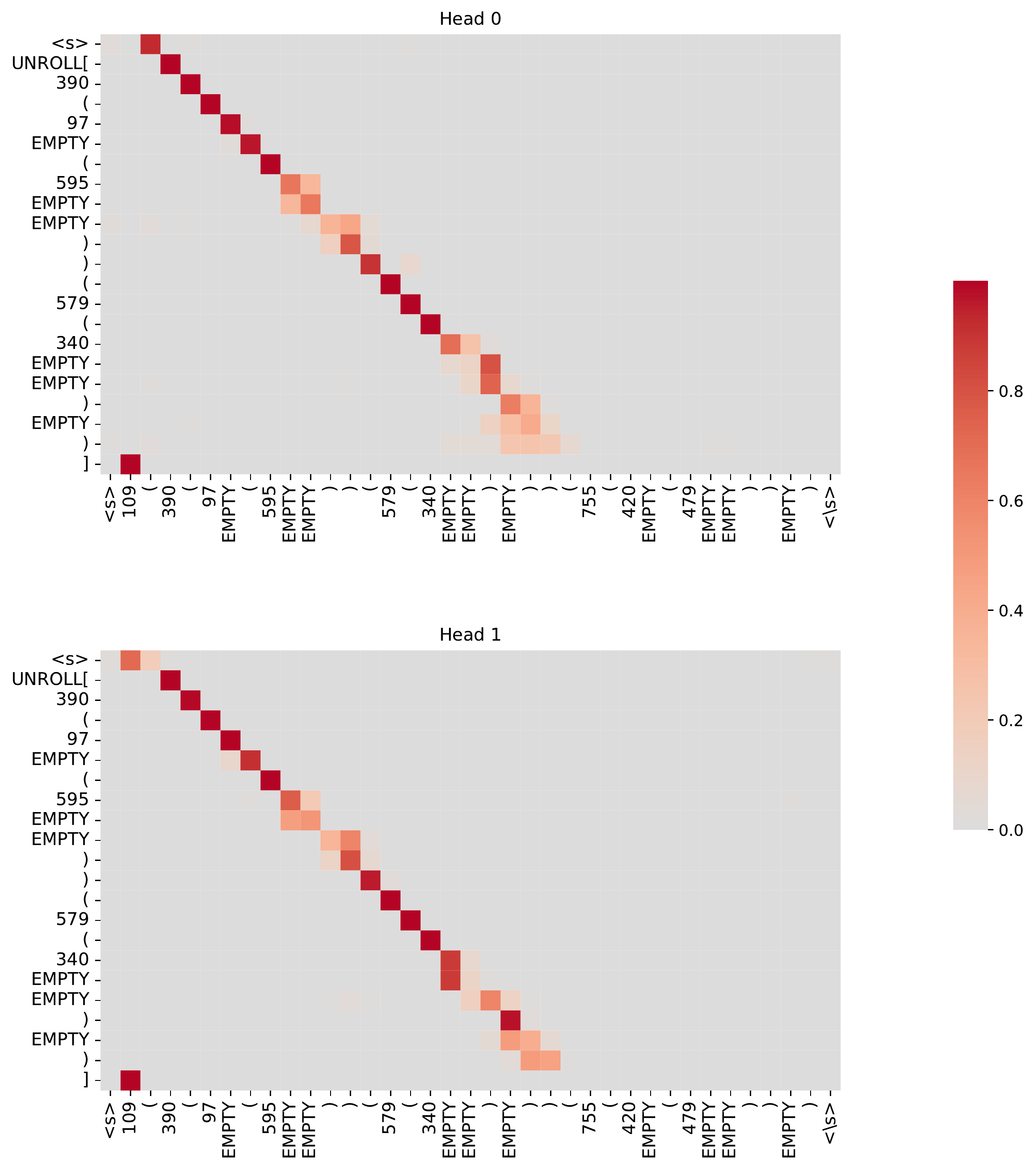}
         \caption{Layer 1 cross-attention at the end of the left sub-tree to be unrolled. The model looks back to the root node whose value is to be copied after unrolling operation on the left sub-tree.}
         \label{fig:traverse3}
     \end{subfigure}
     \caption{Cross-attention during inorder traversal task.}
\end{figure*}

The analysis of attention patterns reveals important insights. In the in-order traversal task, Layer 0 cross-attention heads show distinct behavior: one head focuses on forward parenthesis, brackets, and EMPTY tokens, while the other attends to encoder sequence tokens linearly (Figure~\ref{fig:traverse1}). At Layer 1 (Figure~\ref{fig:traverse2}), both heads correctly attend to the token to be copied, suggesting they write out the next correct token based on Layer 0's output. Brackets and symbols play a crucial role in signaling behavior changes, particularly in steps requiring the insertion of non-consecutive symbols. Decoder self-attention heads attend to previous UNROLL symbols when determining the inclusion or omission of EMPTY tokens. For in-order traversal, higher depths involve attending to bracket statements for copying root nodes, unlike binary trees with a single root node. This behavior may be utilized by decoder cross-attention heads to copy the first node within the UNROLL statement.

Our analysis indicates that the individual encoder self-attention heads do not appear to play a significant independent role in the transformer model's performance. This is demonstrated by our counterfactual patching experiments, which resulted in minimal changes to the network's predictions. On the other hand, decoder cross-attention appears to be a crucial component of the transformer circuit. As the sequence progresses, however, decoder self-attention becomes increasingly important and the model takes into account the structure of the previously-generated content (especially at higher depths, where keeping track of multiple levels of parent nodes becomes vital). 

\section{Extrapolation of Different Model Architectures}
\label{app:extrap}

We were initially not sure whether to evaluate the tasks of interests on an encoder-only, decoder-only, or encoder-decoder transformer model. To decide, we ran preliminary experiments to evaluate these three architectures under our setup for a related toy experiment, as well as on one of our two tasks.

Our toy experiment focused on string reversal. We trained models on strings of lengths ranging from 10 to 37, and tested up to length 50. %We plotted the performance drop as the extrapolation length increased.
We used three different architectures: a 2-layer encoder-decoder transformer model, a 4-layer encoder-only model, and a 4-layer decoder-only model. We trained until we achieved perfect accuracy on both the training and in-domain validation datasets. We employed the same ``random-padding'' strategy as in the main experiment.

\begin{figure}
         \centering
         \includegraphics[width=0.9\columnwidth]{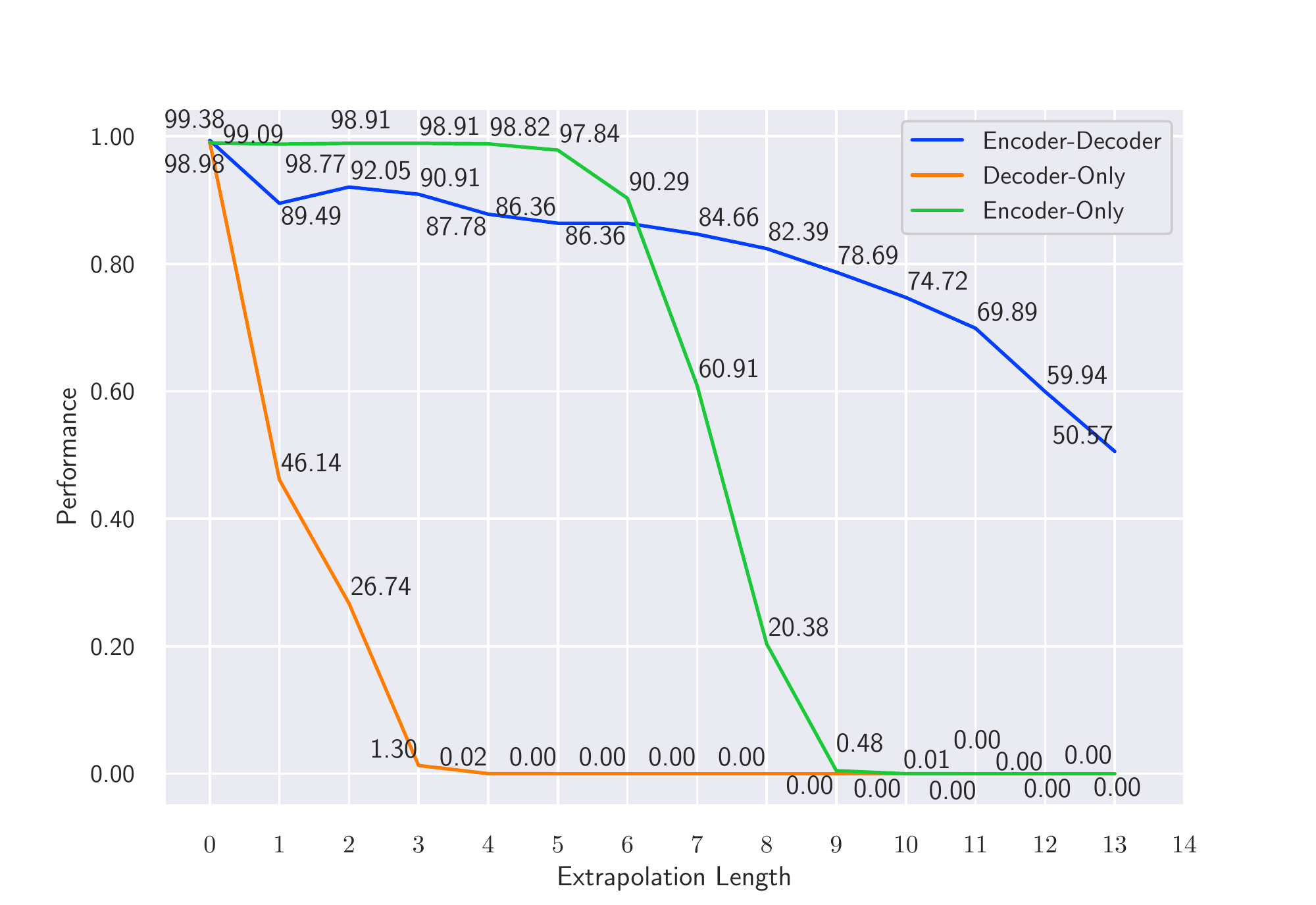}
         \vspace{-0.3cm}
         \caption{Performance of Extrapolation Study}
         \label{fig:extrapolation1}
\end{figure}

\begin{figure}
         \centering
         \includegraphics[width=\columnwidth]{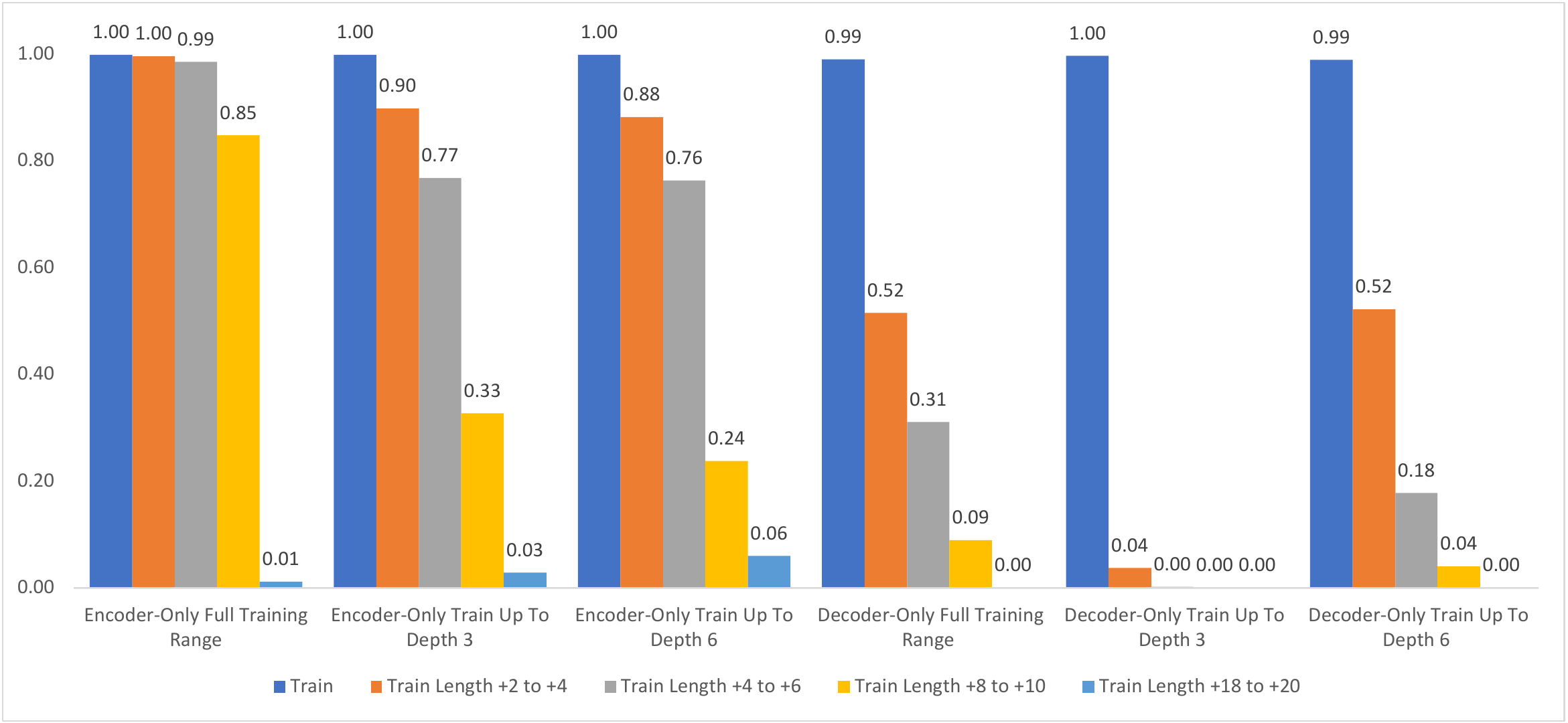}
         \caption{Accuracy of encoder-only and decoder-only transformers on natural order binary successor task.} 
         \label{fig:enc_only_dec_only}
\end{figure}

The encoder-decoder model exhibited the best extrapolation performance among the three architectures (Figure~\ref{fig:extrapolation1}). Both the encoder-only and decoder-only models experienced a rapid decline in performance as the extrapolation length increased.
Interestingly, we also observed that the model failed to extrapolate when trained on a fixed-length ``learnable token'' setup, even though we ensured that the positional embeddings were seen during training. Moreover, the model also struggled to extrapolate when padding was not applied.

The results of the encoder-only and decoder-only transformers on the natural order binary successor task (each with 4 layers) are presented in Figure~\ref{fig:enc_only_dec_only}.

\section{Binary Successor Performance By Depth}
\label{app:depth}
\begin{figure*}
     \centering
     \begin{subfigure}[b]{.48\columnwidth}
         \centering
         \includegraphics[width=\columnwidth]{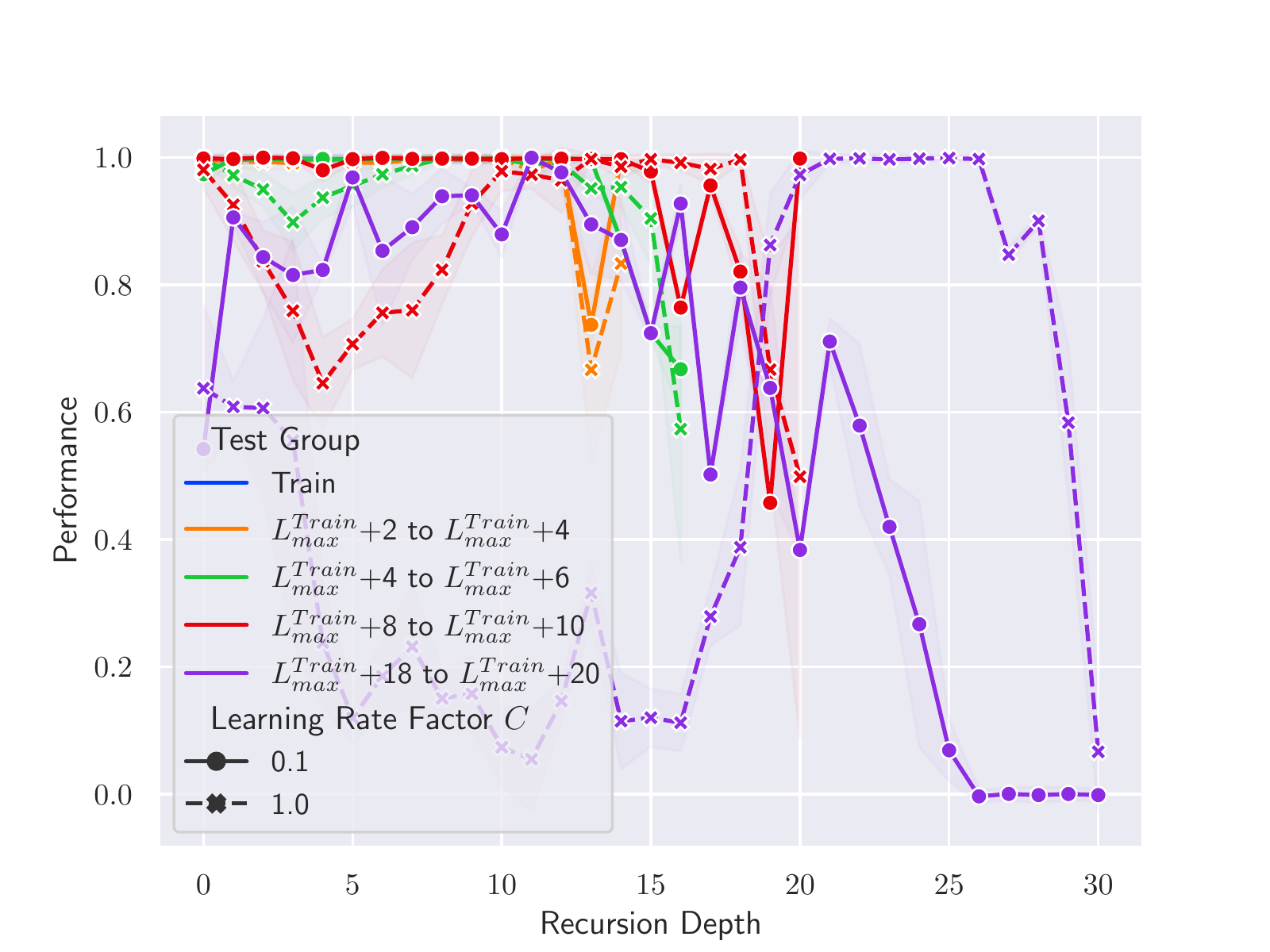}
         \caption{Trained on Binary Strings Up to 2048}
     \end{subfigure}
     \begin{subfigure}[b]{.48\columnwidth}
         \centering
         \includegraphics[width=\columnwidth]{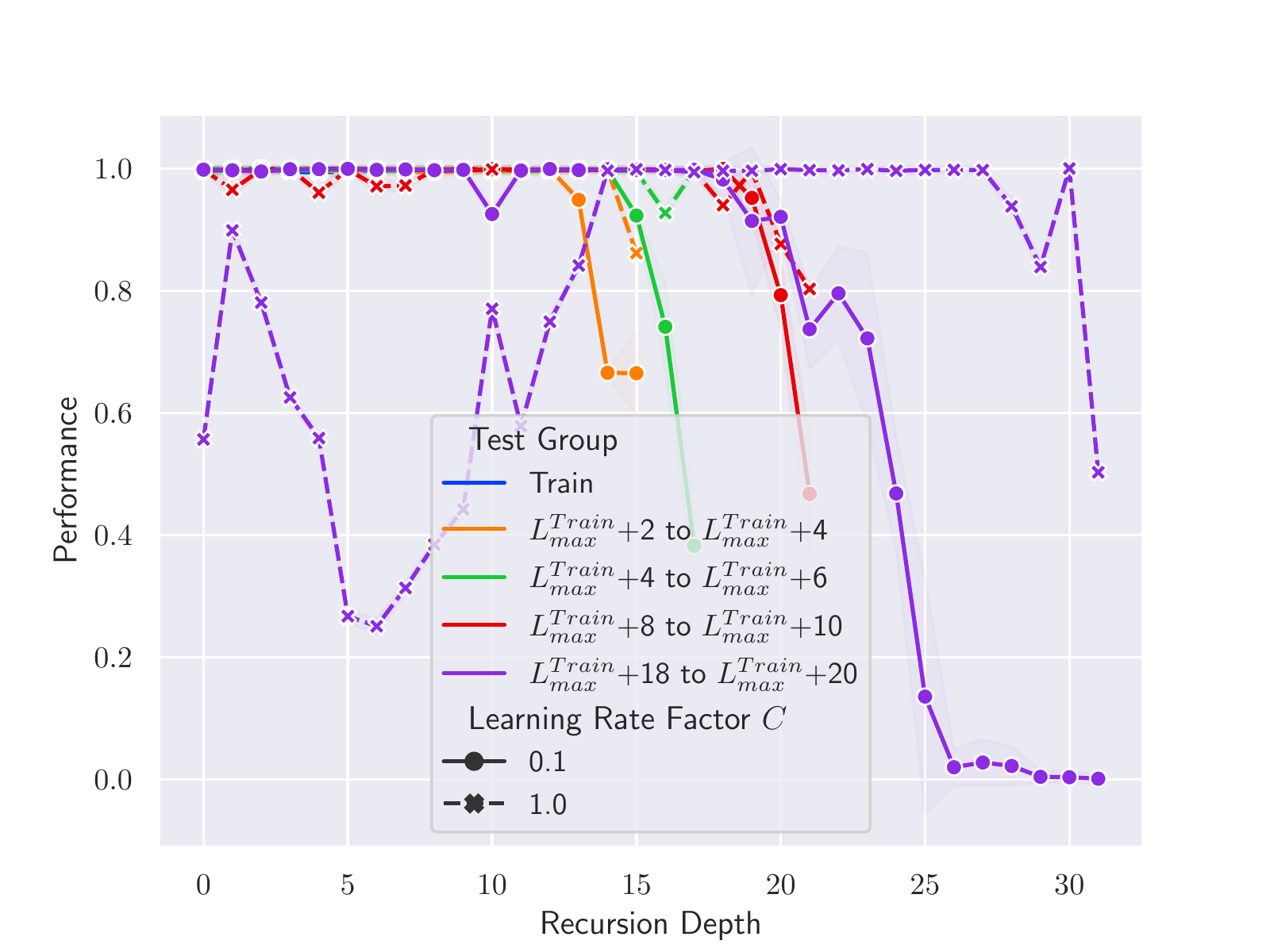}
         \caption{Trained on Binary Strings Up to 4096}
     \end{subfigure}
     \begin{subfigure}[b]{.48\columnwidth}
         \centering
         \includegraphics[width=\columnwidth]{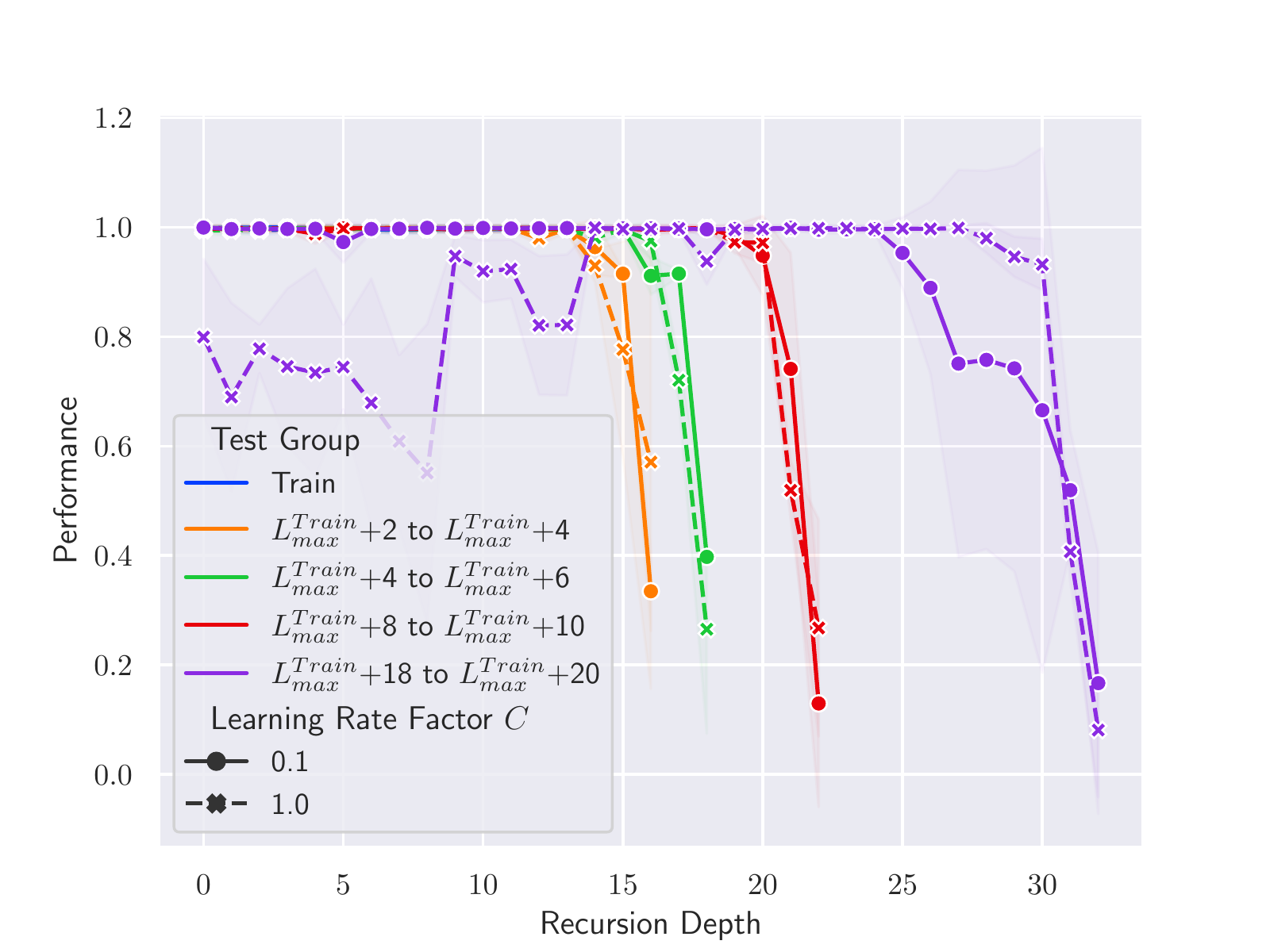}
         \caption{Trained on Binary Strings Up to 8192}
     \end{subfigure}
     \begin{subfigure}[b]{.48\columnwidth}
         \centering
         \includegraphics[width=\columnwidth]{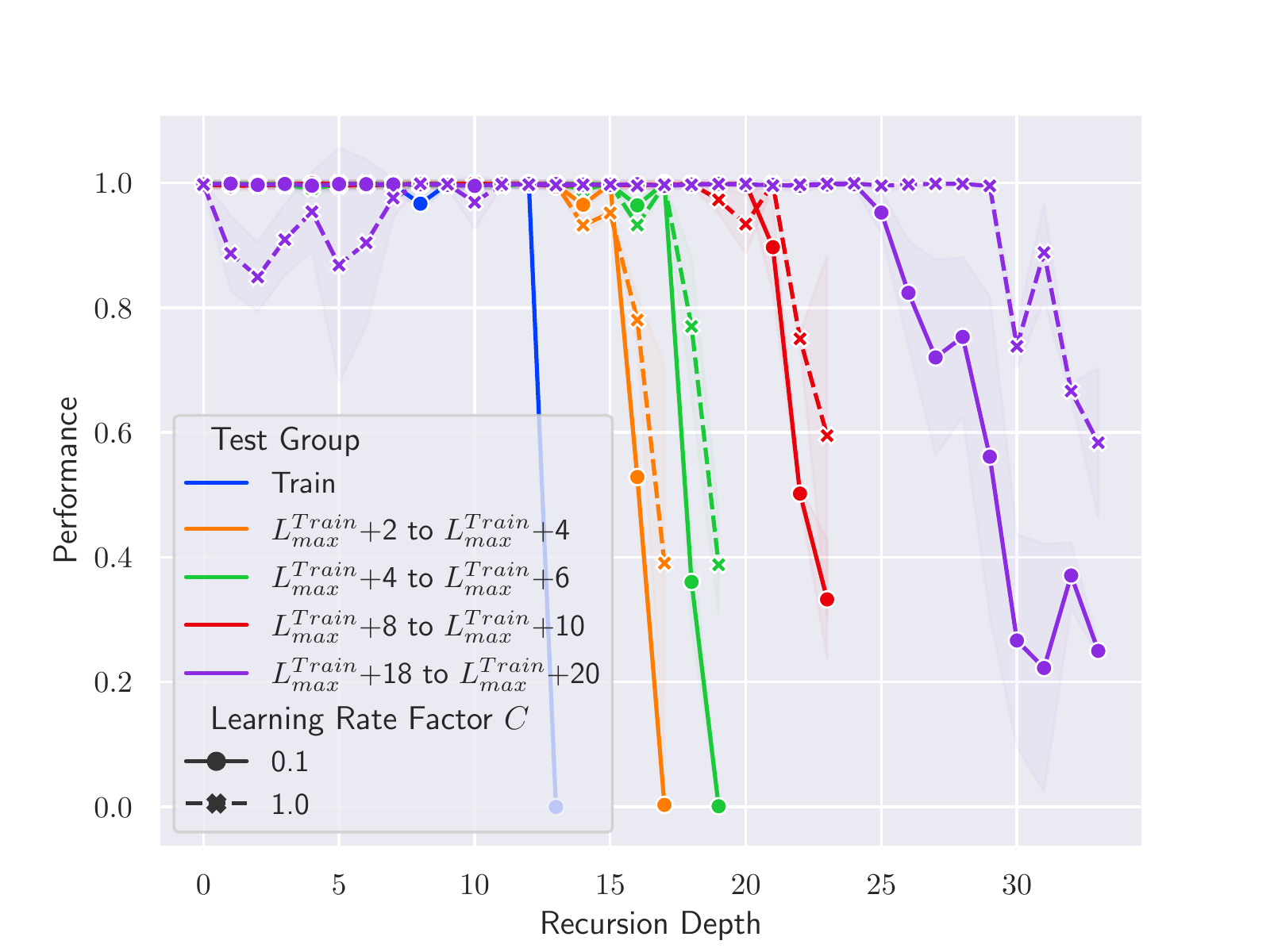}
         \caption{Trained on Binary Strings Up to 16384}
     \end{subfigure}
     \begin{subfigure}[b]{.48\columnwidth}
         \centering
         \includegraphics[width=\columnwidth]{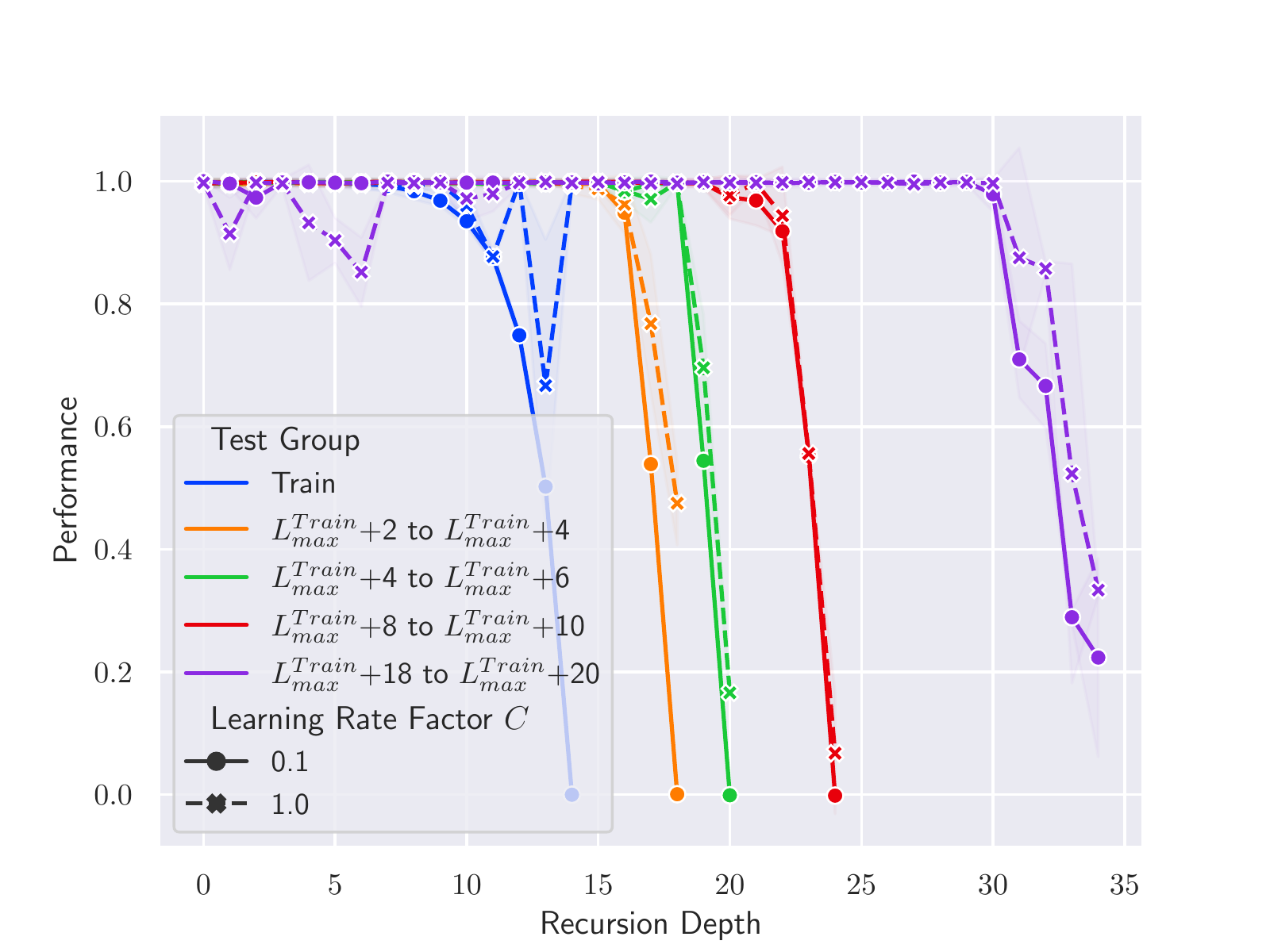}
         \caption{Trained on Binary Strings Up to 32768}
     \end{subfigure}
     \begin{subfigure}[b]{.48\columnwidth}
         \centering
         \includegraphics[width=\columnwidth]{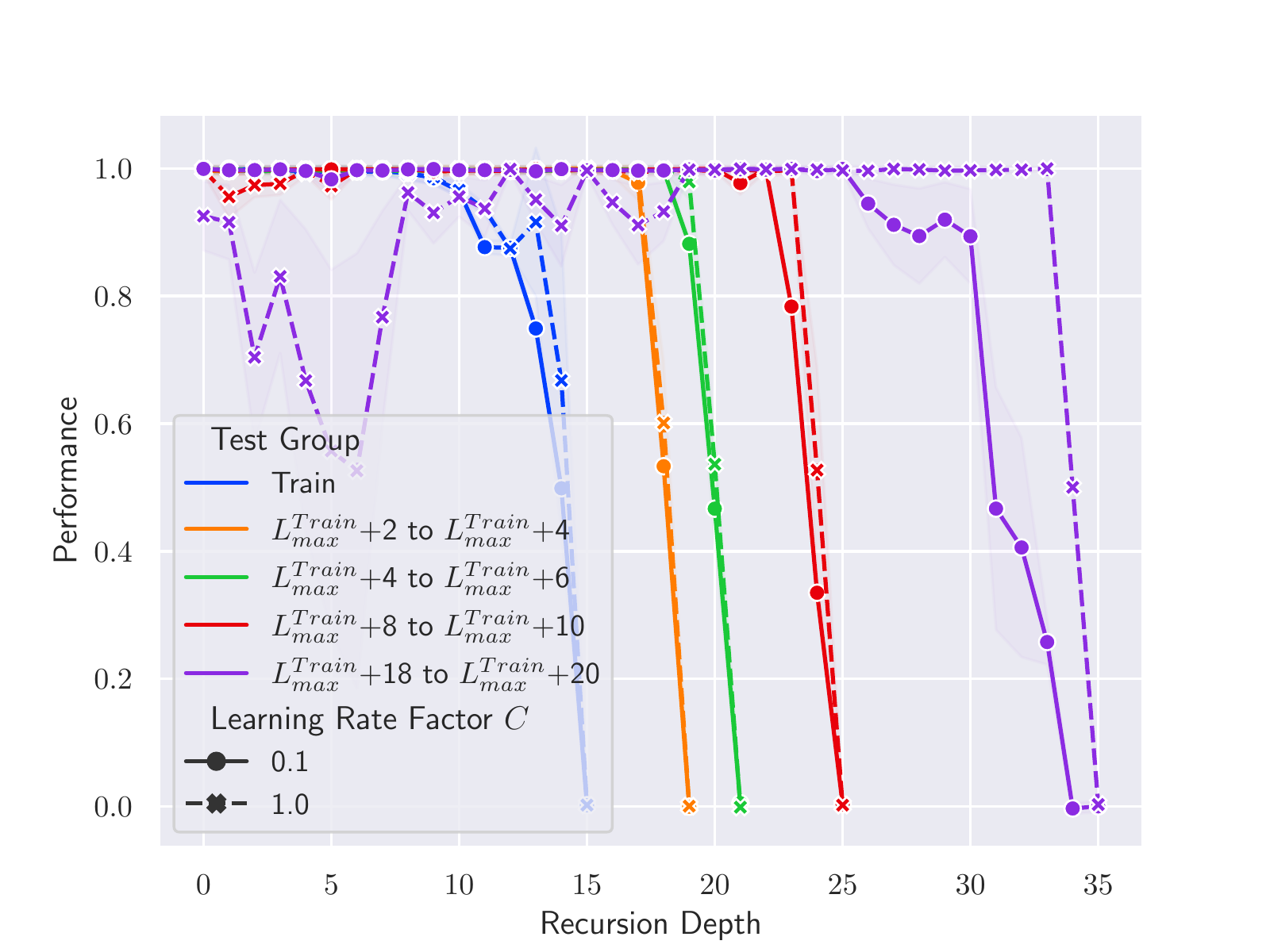}
         \caption{Trained on Binary Strings Up to 65536}
     \end{subfigure}
     \begin{subfigure}[b]{.48\columnwidth}
         \centering
         \includegraphics[width=\columnwidth]{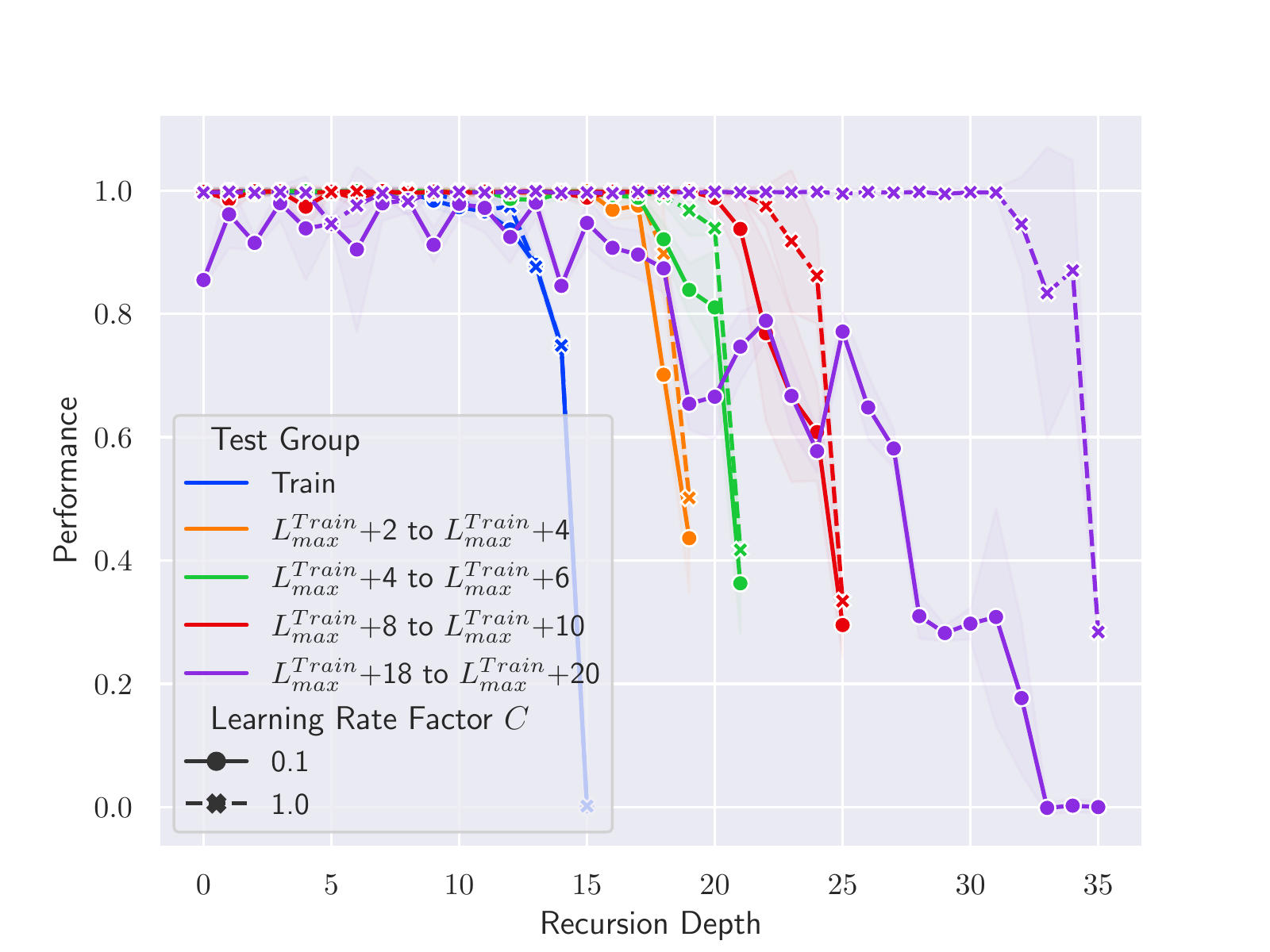}
         \caption{Trained on Binary Strings Up to 131072}
     \end{subfigure}
        % \vspace{-3mm}
    \caption{Accuracy versus recursion depth: natural order.}
    \label{fig:nat_perf_depth}
\end{figure*}

\begin{figure*}[!tb]
     \centering
     \begin{subfigure}[b]{.48\columnwidth}
         \centering
         \includegraphics[width=\columnwidth]{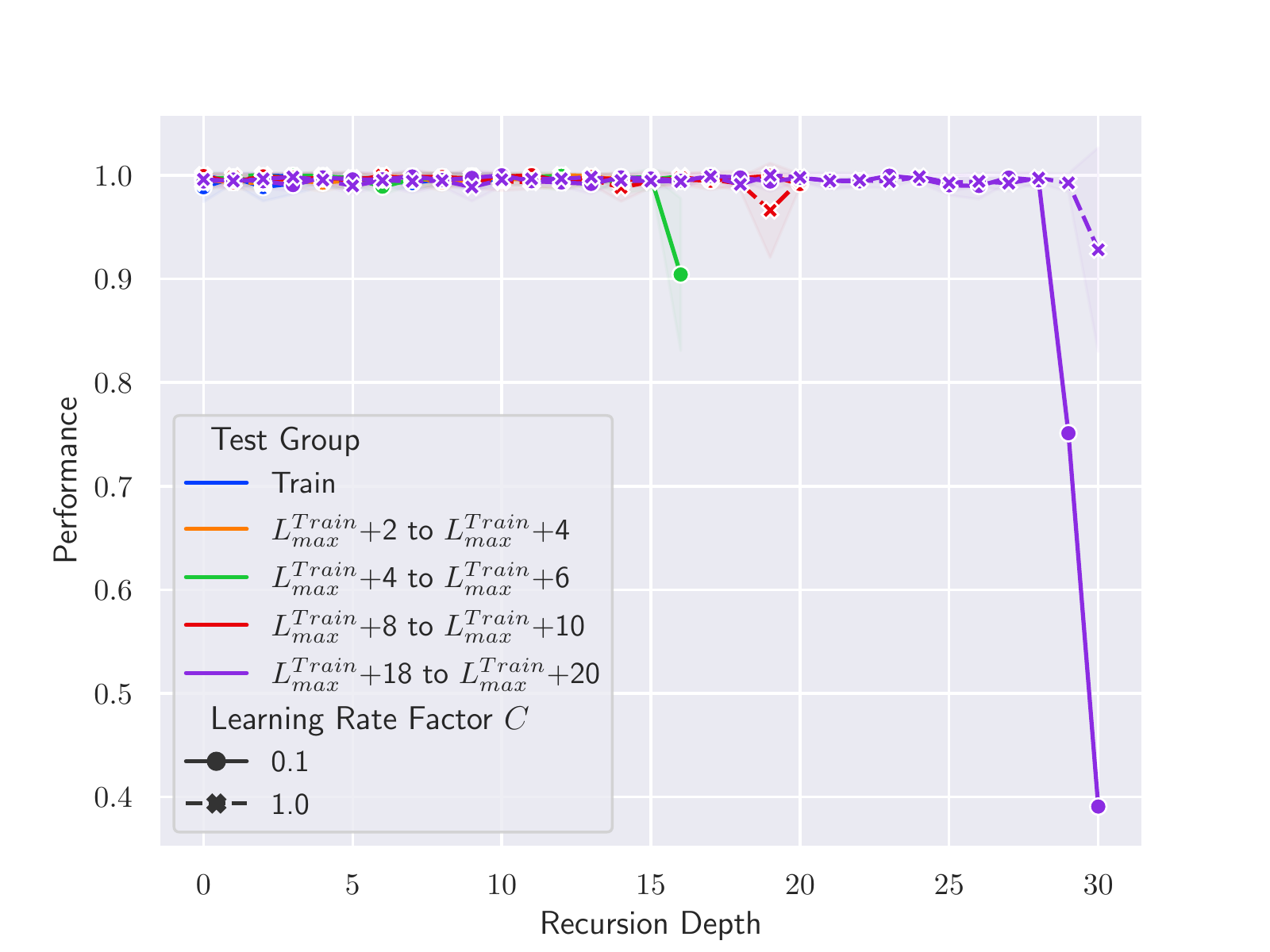}
         \caption{Trained on Binary Strings Up to 2048}
     \end{subfigure}
     \begin{subfigure}[b]{.48\columnwidth}
         \centering
         \includegraphics[width=\columnwidth]{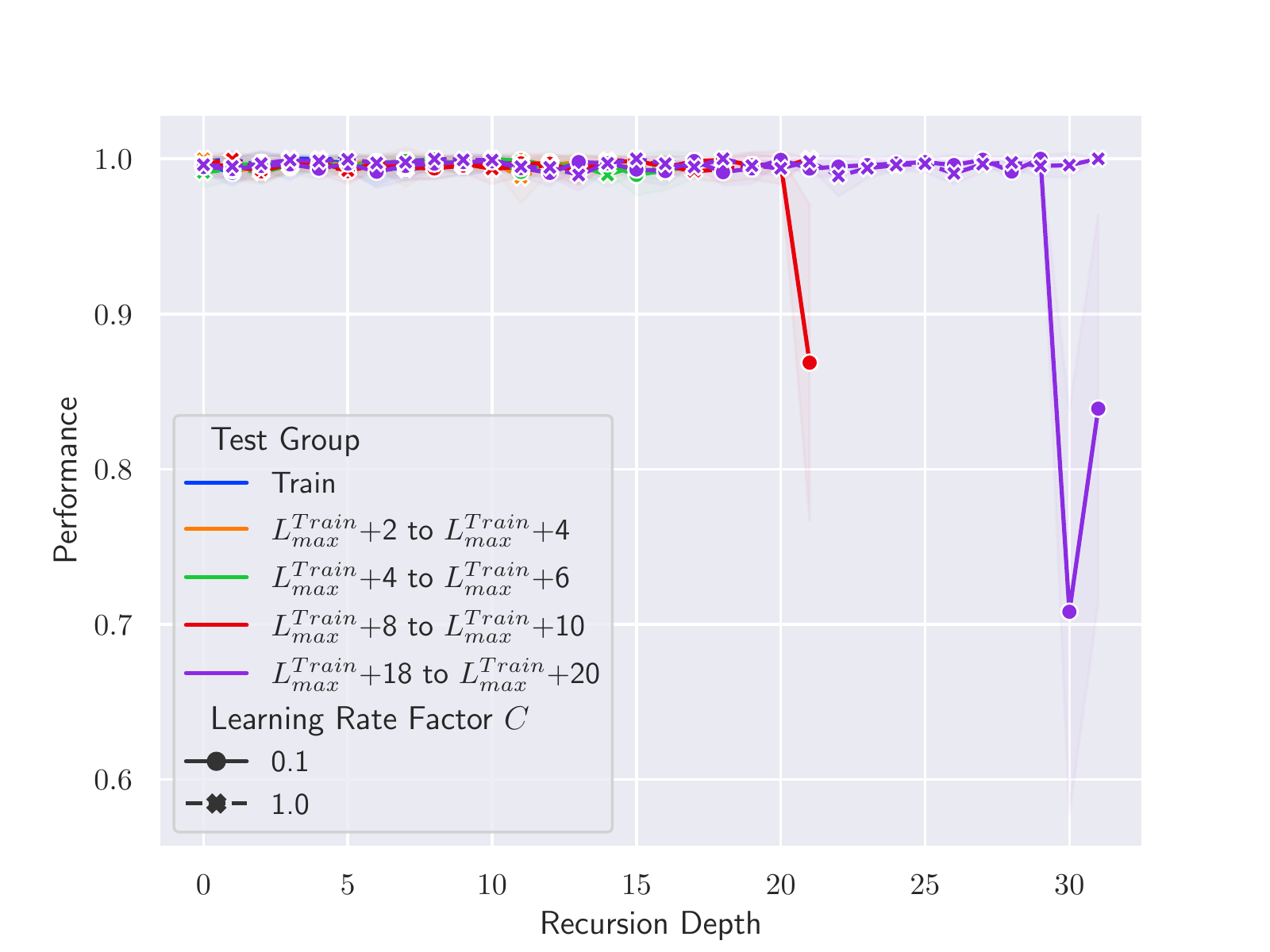}
         \caption{Trained on Binary Strings Up to 4096}
     \end{subfigure}
     \begin{subfigure}[b]{.48\columnwidth}
         \centering
         \includegraphics[width=\columnwidth]{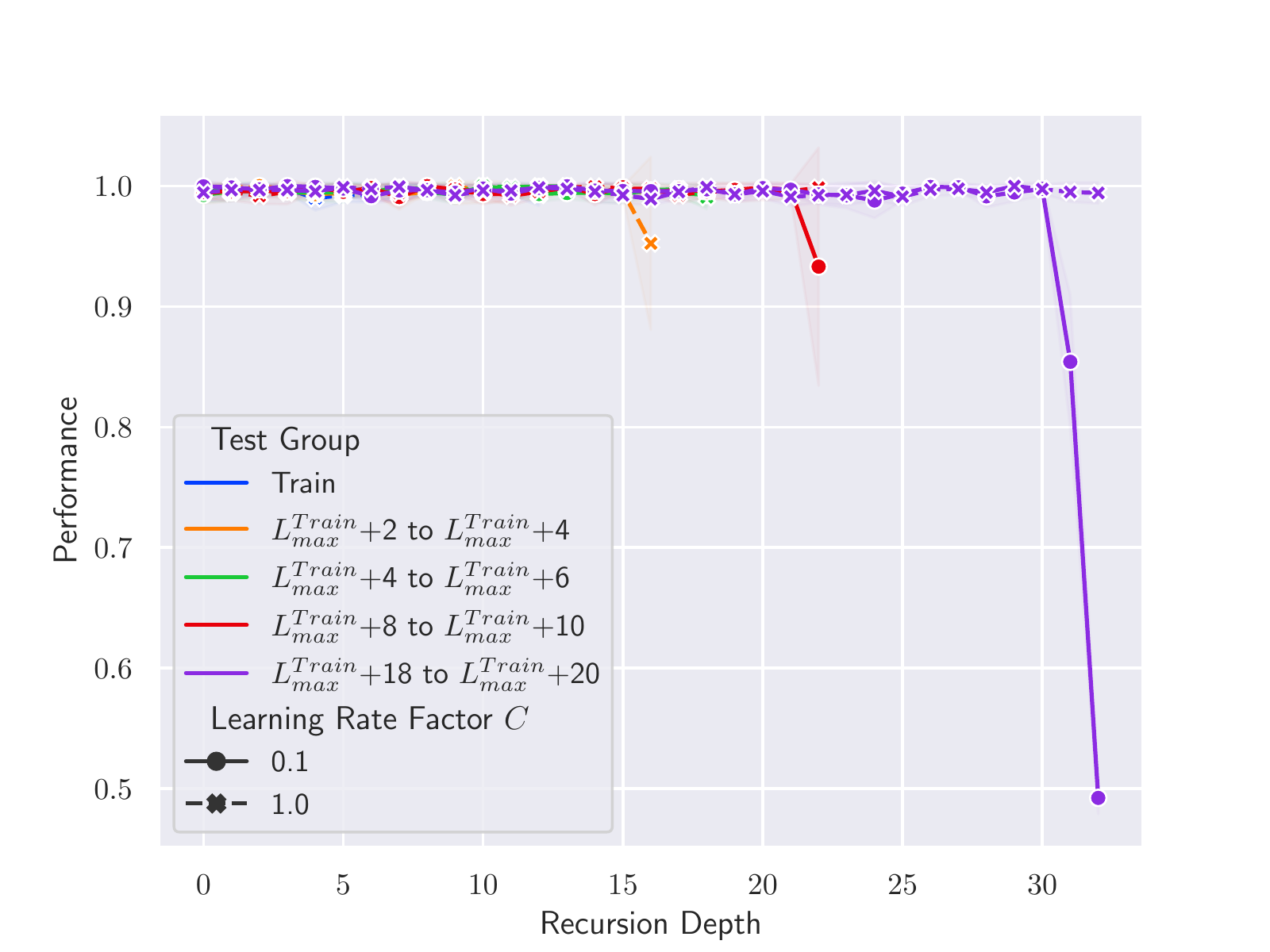}
         \caption{Trained on Binary Strings Up to 8192}
     \end{subfigure}
     \begin{subfigure}[b]{.48\columnwidth}
         \centering
         \includegraphics[width=\columnwidth]{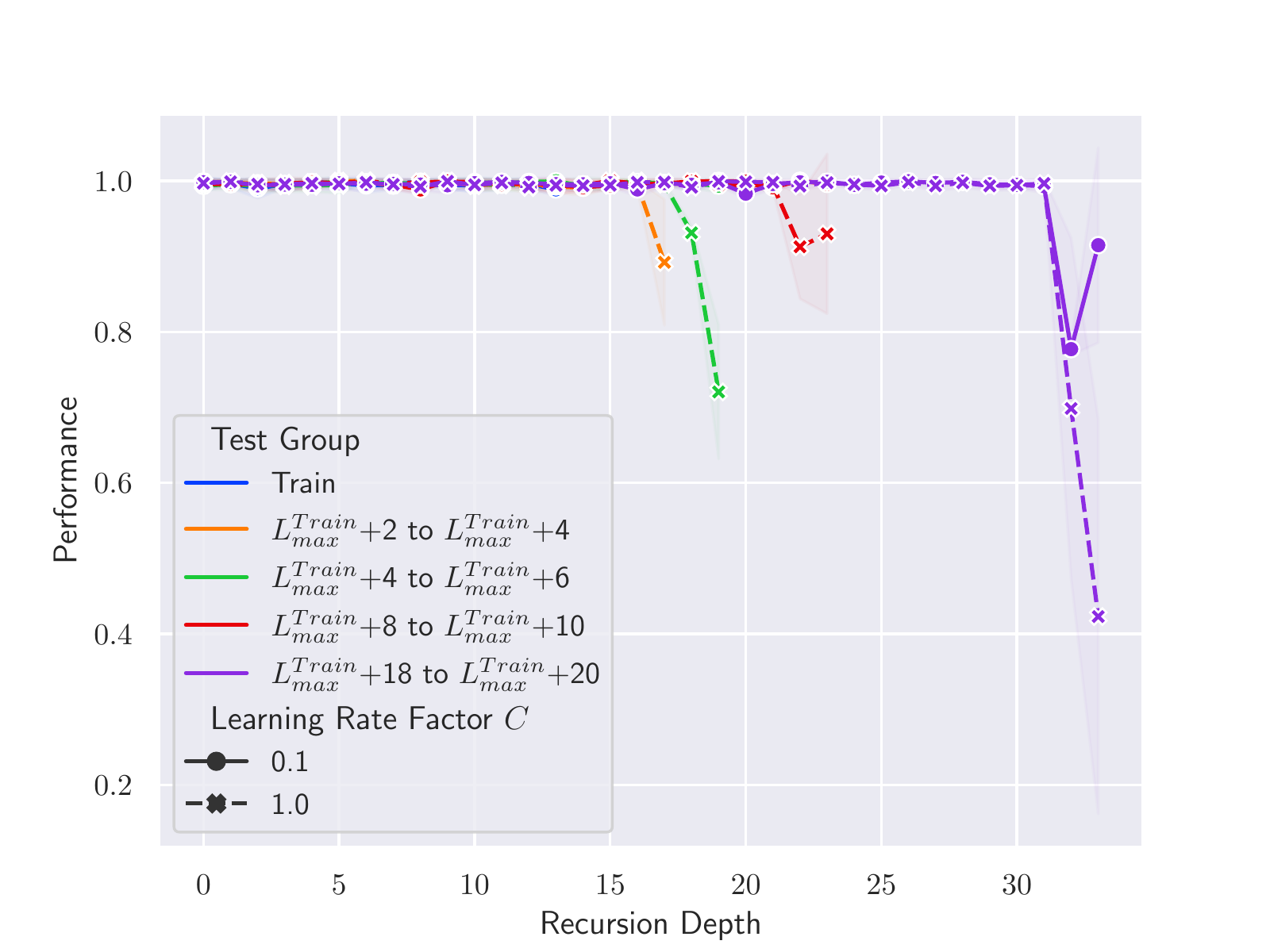}
         \caption{Trained on Binary Strings Up to 16384}
     \end{subfigure}
     \begin{subfigure}[b]{.48\columnwidth}
         \centering
         \includegraphics[width=\columnwidth]{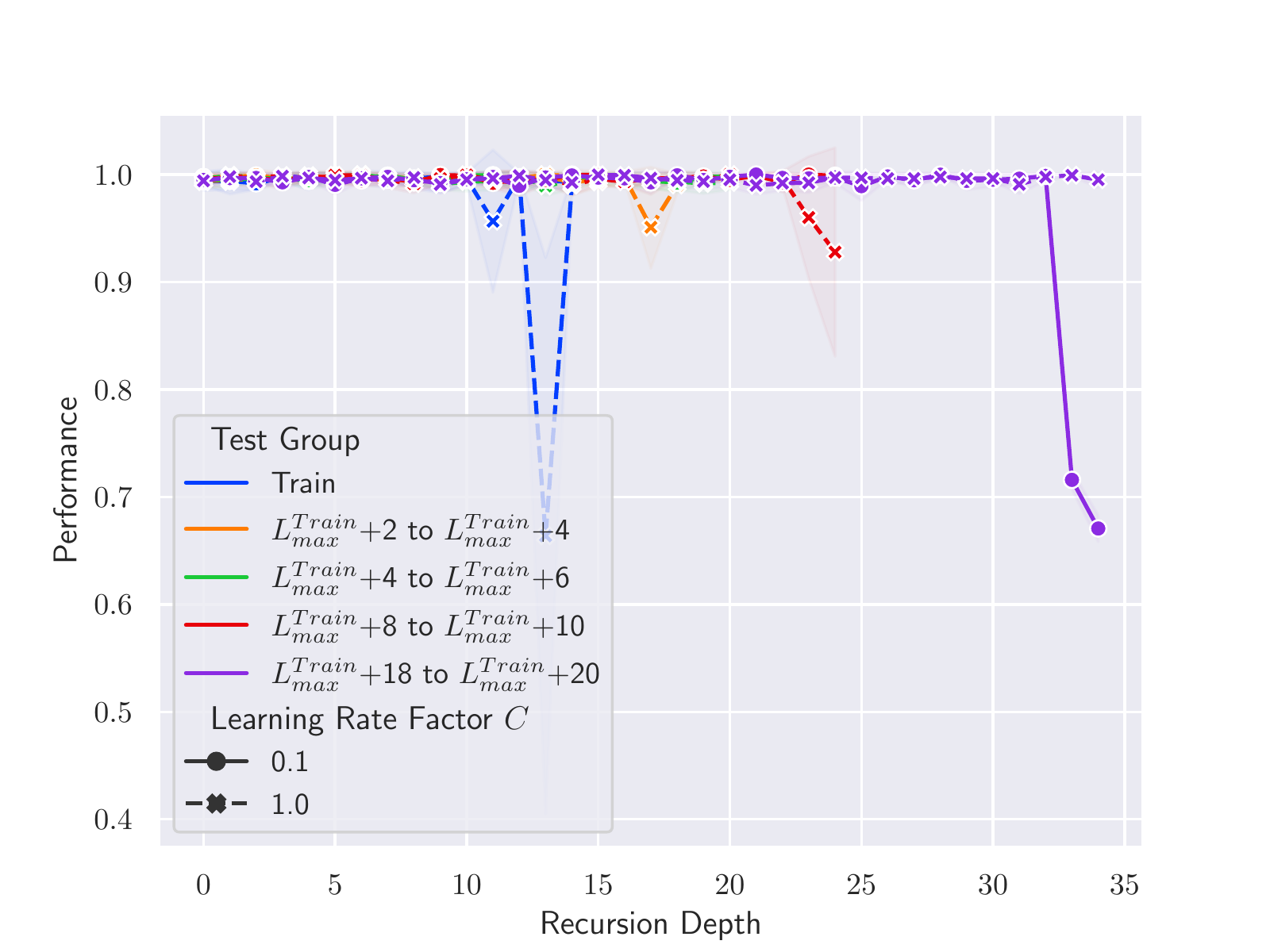}
         \caption{Trained on Binary Strings Up to 32768}
     \end{subfigure}
     \begin{subfigure}[b]{.48\columnwidth}
         \centering
         \includegraphics[width=\columnwidth]{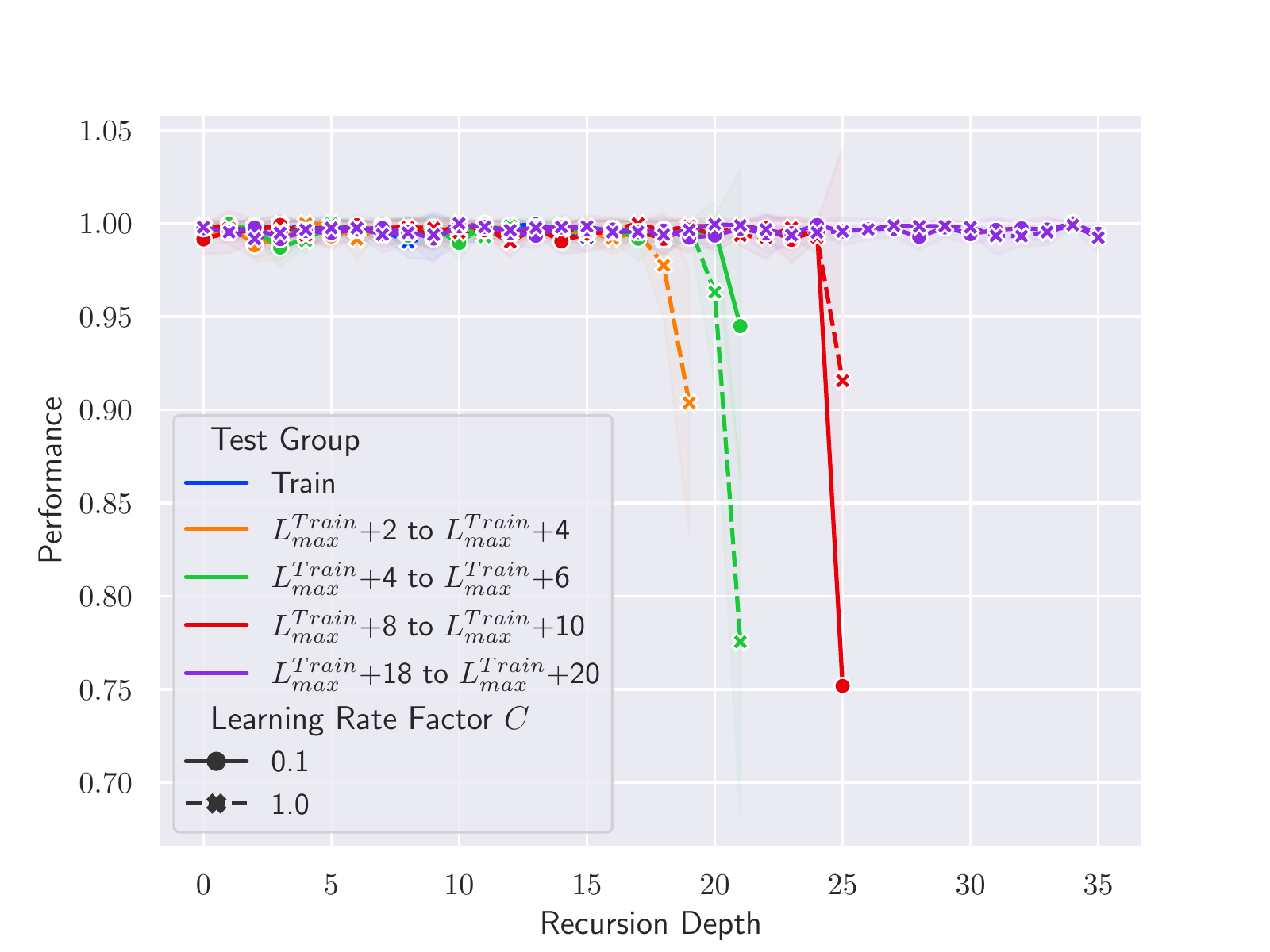}
         \caption{Trained on Binary Strings Up to 65536}
     \end{subfigure}
     \begin{subfigure}[b]{.48\columnwidth}
         \centering
         \includegraphics[width=\columnwidth]{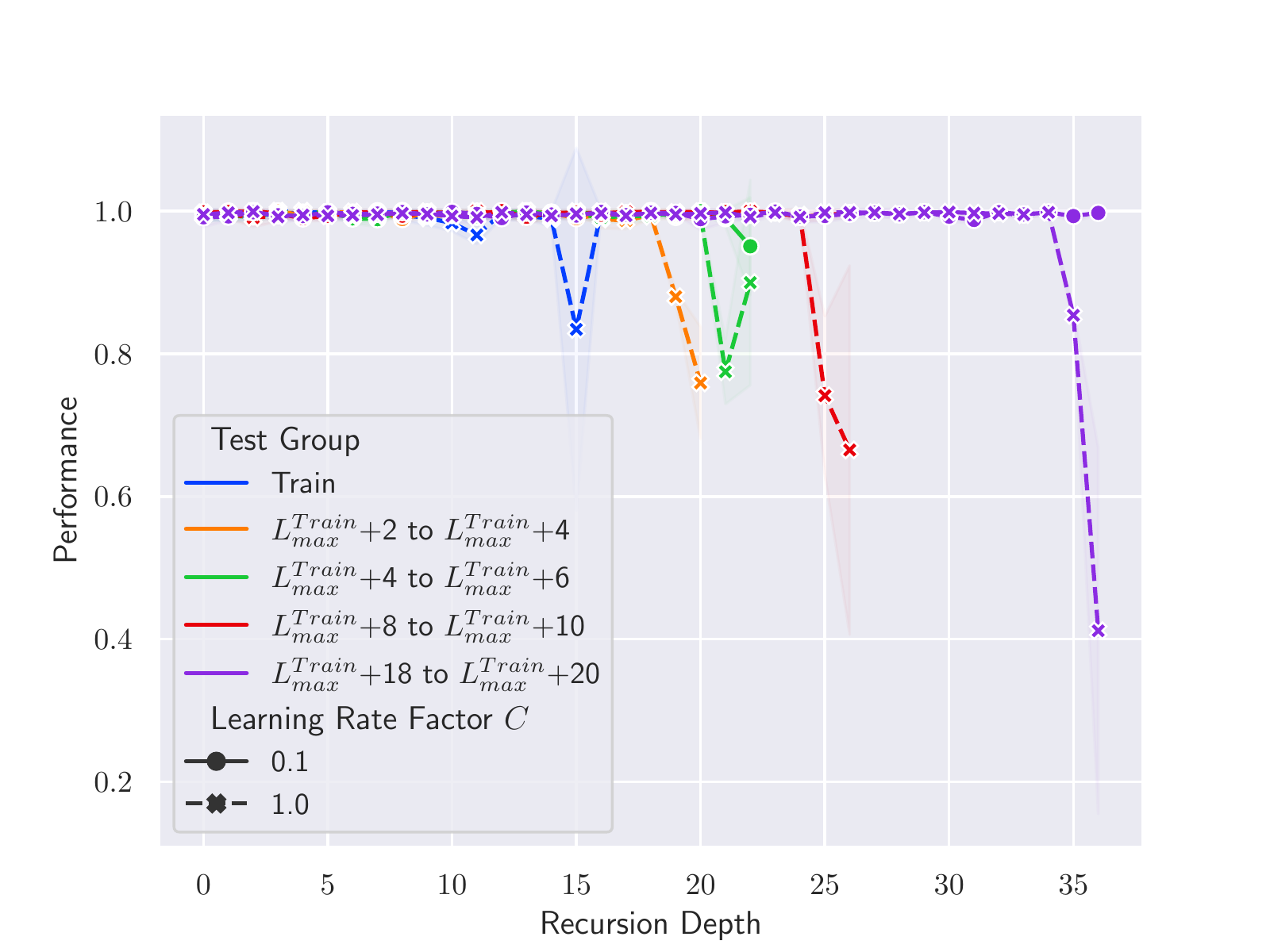}
         \caption{Trained on Binary Strings Up to 131072}
     \end{subfigure}
        % \vspace{-3mm}
    \caption{Accuracy versus recursion depth: reverse order.}
    \label{fig:rev_perf_depth2}
\end{figure*}

\begin{figure*}[!tb]
     \centering
     \begin{subfigure}[b]{.48\columnwidth}
         \centering
         \includegraphics[width=\columnwidth]{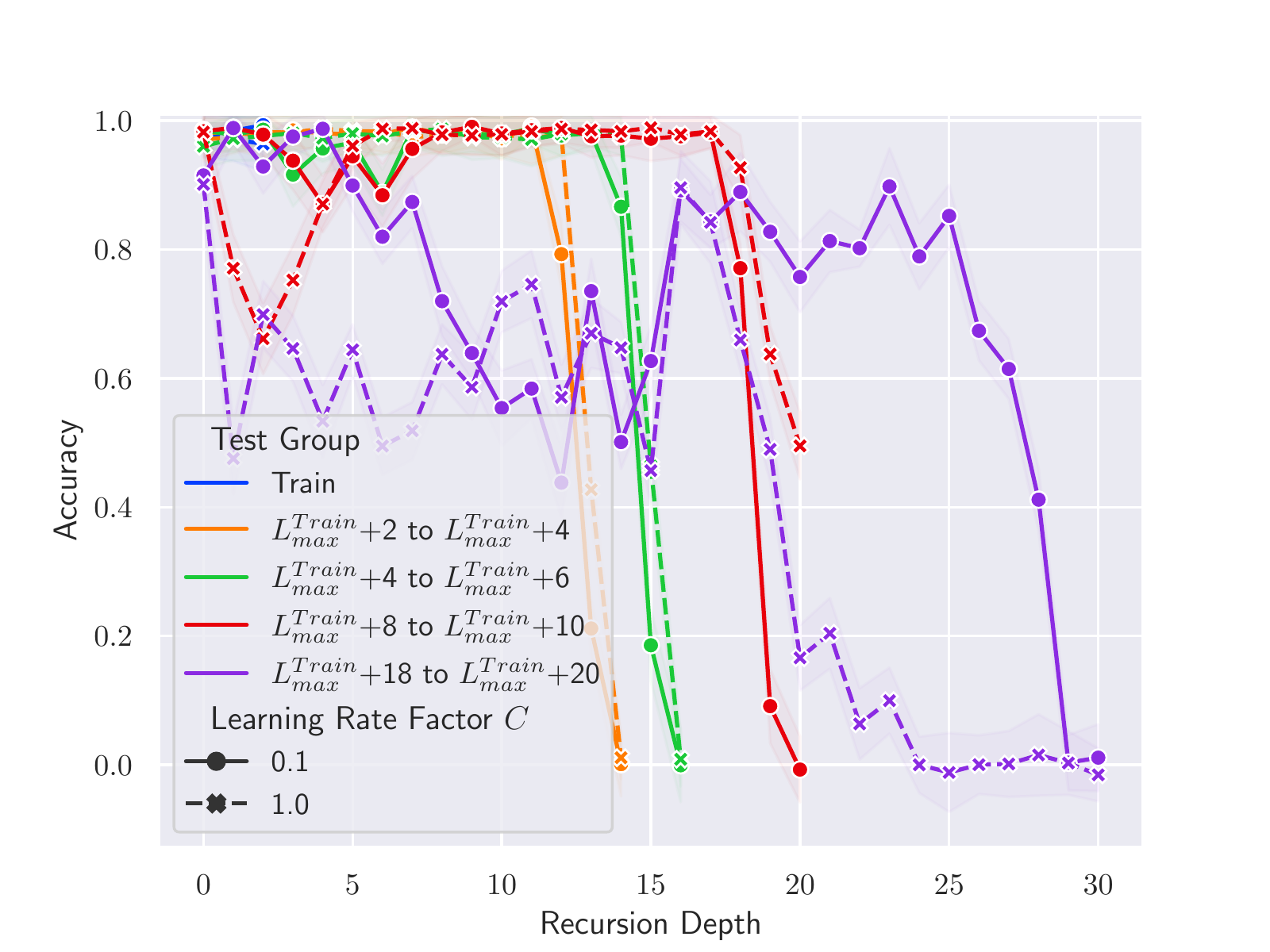}
         \caption{Trained on Binary Strings Up to 2048}
     \end{subfigure}
     \begin{subfigure}[b]{.48\columnwidth}
         \centering
         \includegraphics[width=\columnwidth]{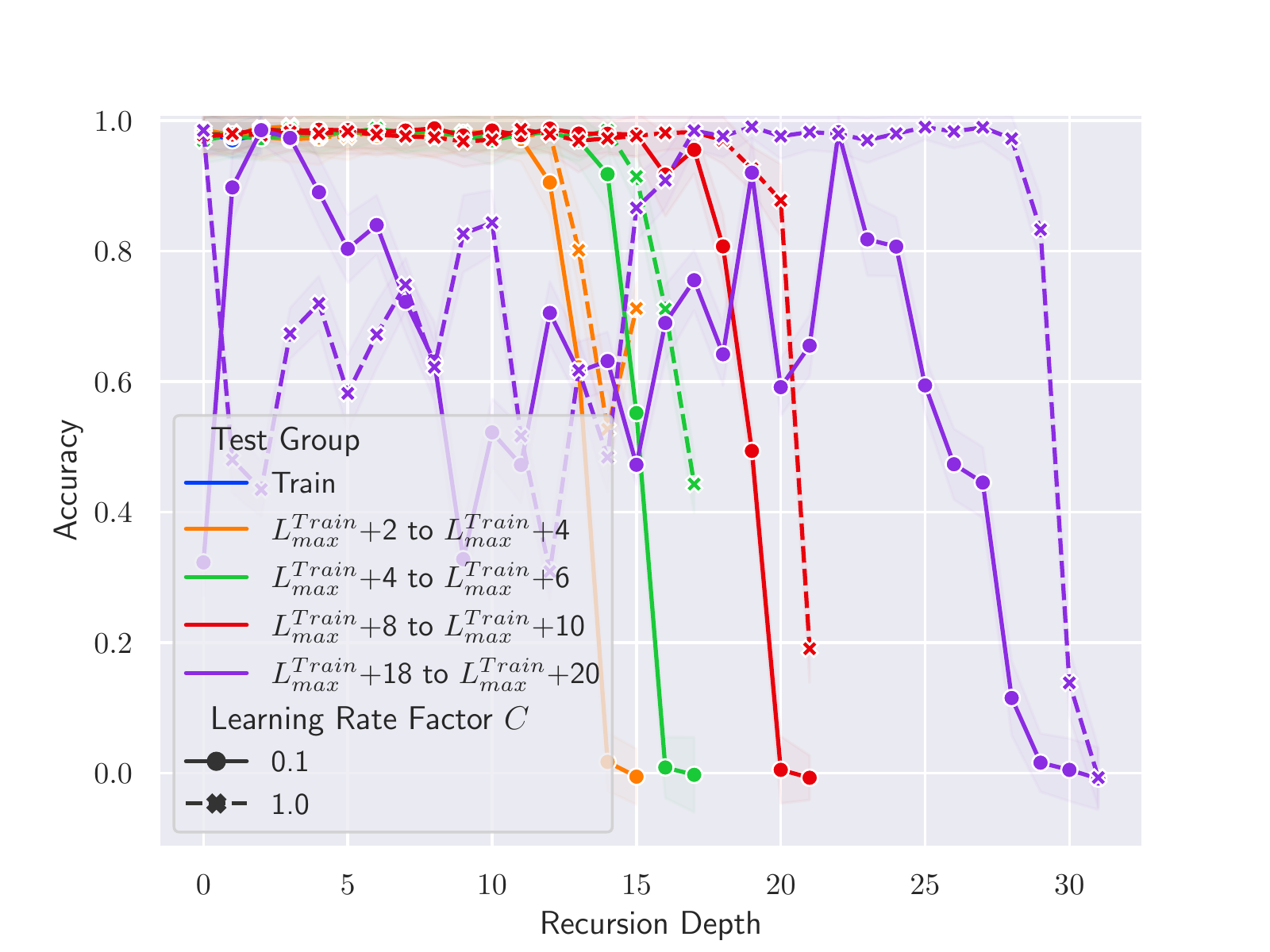}
         \caption{Trained on Binary Strings Up to 4096}
     \end{subfigure}
     \begin{subfigure}[b]{.48\columnwidth}
         \centering
         \includegraphics[width=\columnwidth]{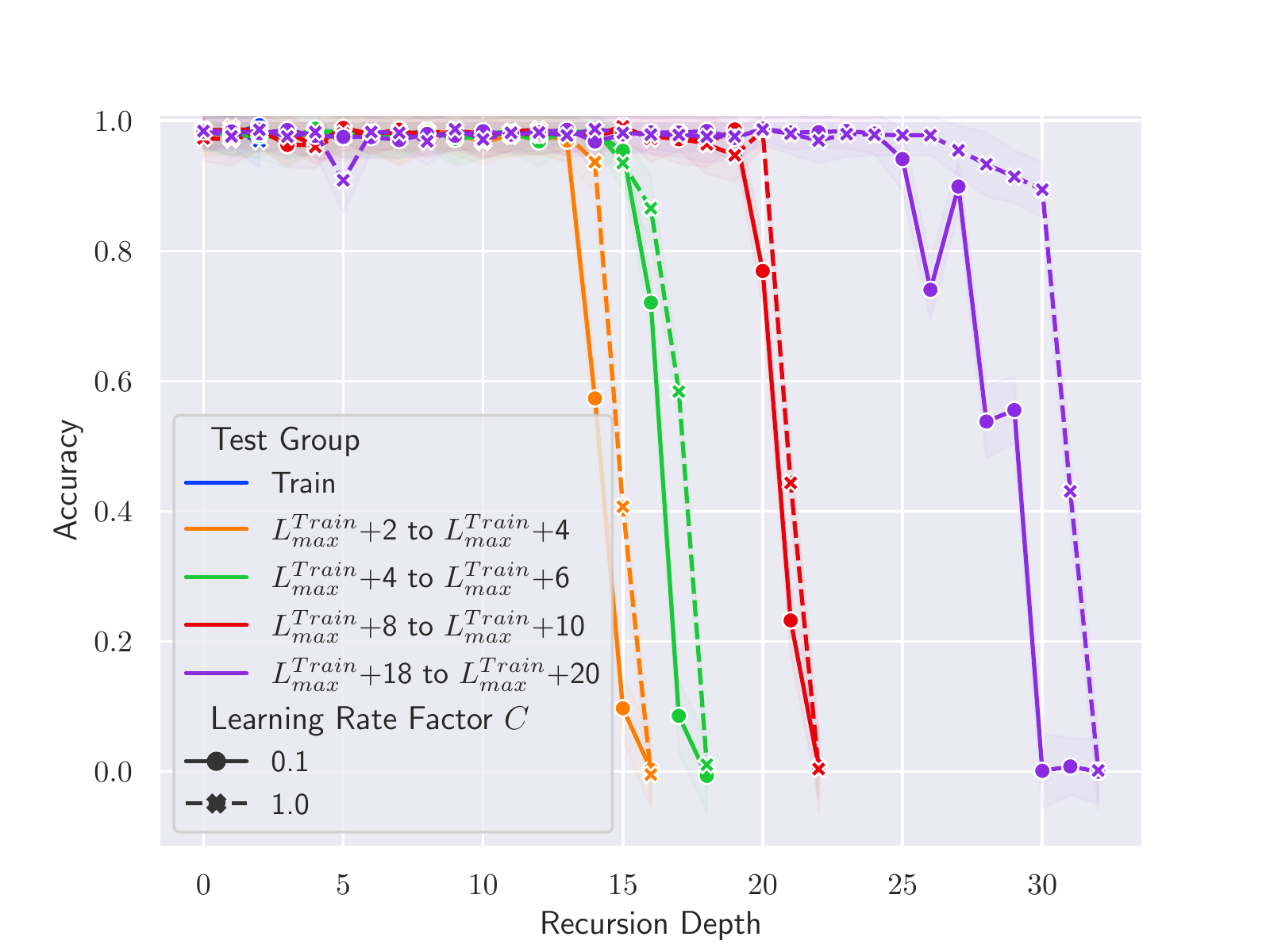}
         \caption{Trained on Binary Strings Up to 8192}
     \end{subfigure}
     \begin{subfigure}[b]{.48\columnwidth}
         \centering
         \includegraphics[width=\columnwidth]{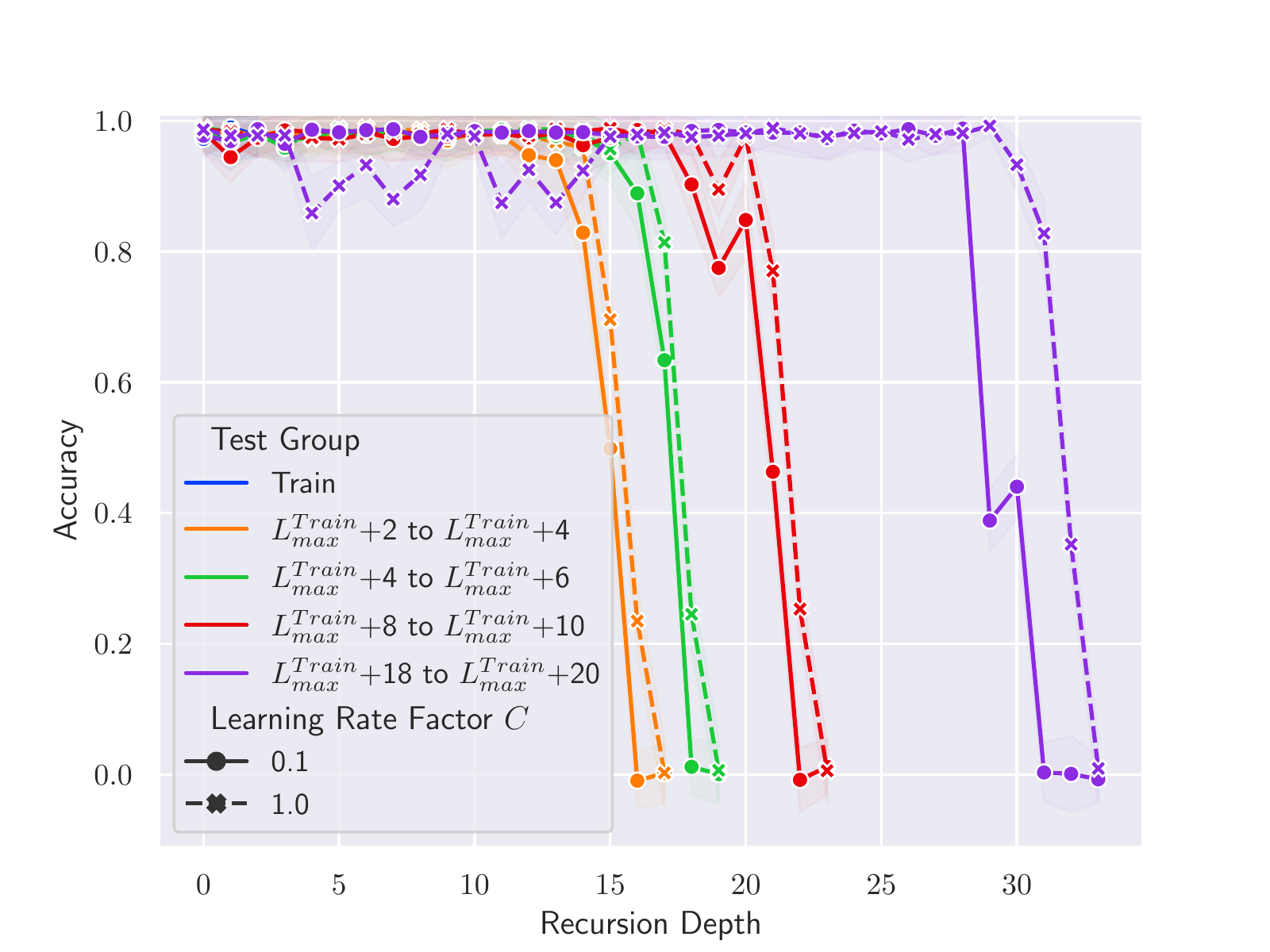}
         \caption{Trained on Binary Strings Up to 16384}
     \end{subfigure}
     \begin{subfigure}[b]{.48\columnwidth}
         \centering
         \includegraphics[width=\columnwidth]{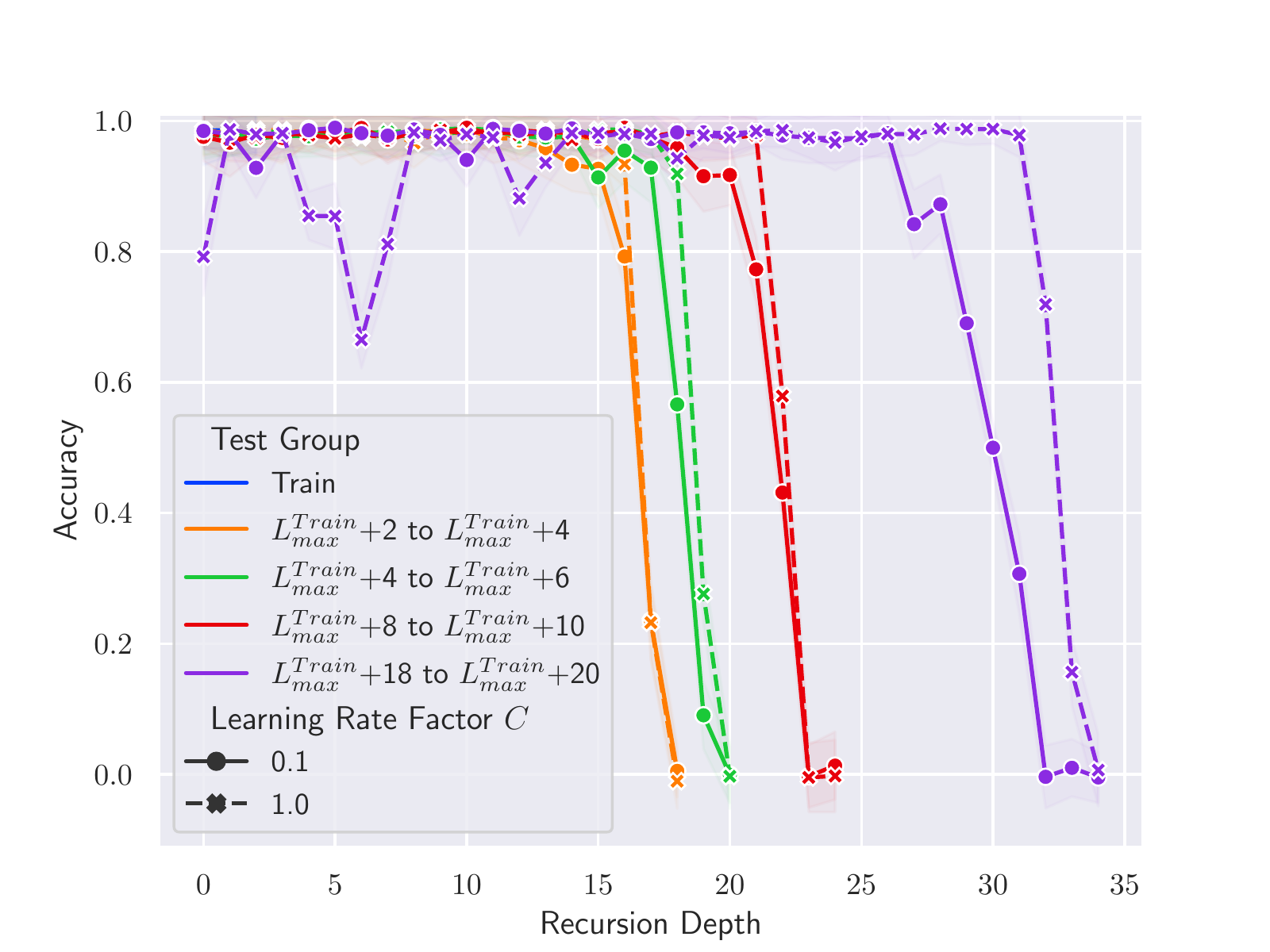}
         \caption{Trained on Binary Strings Up to 32768}
     \end{subfigure}
     \begin{subfigure}[b]{.48\columnwidth}
         \centering
         \includegraphics[width=\columnwidth]{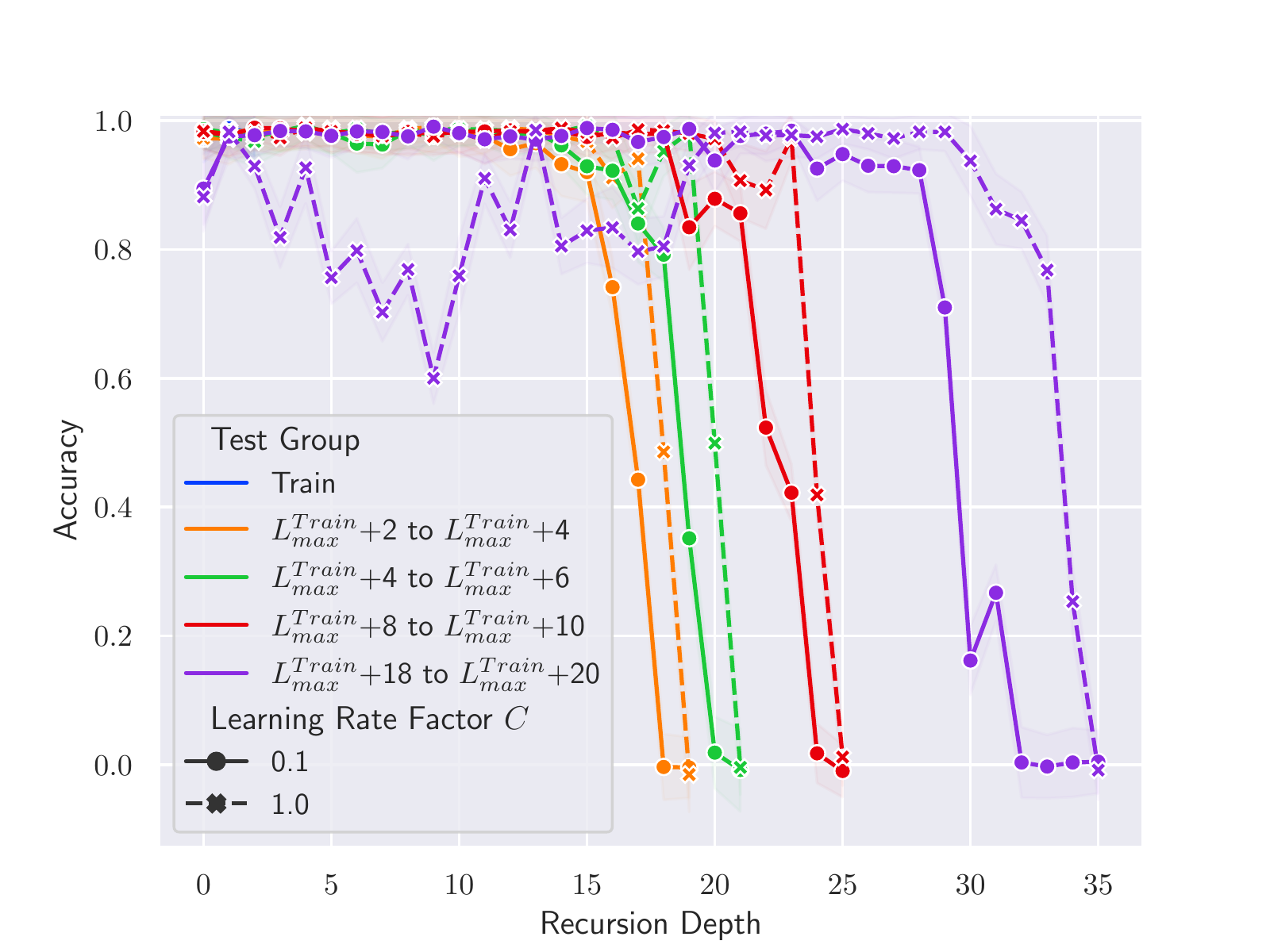}
         \caption{Trained on Binary Strings Up to 65536}
     \end{subfigure}
     \begin{subfigure}[b]{.48\columnwidth}
         \centering
         \includegraphics[width=\columnwidth]{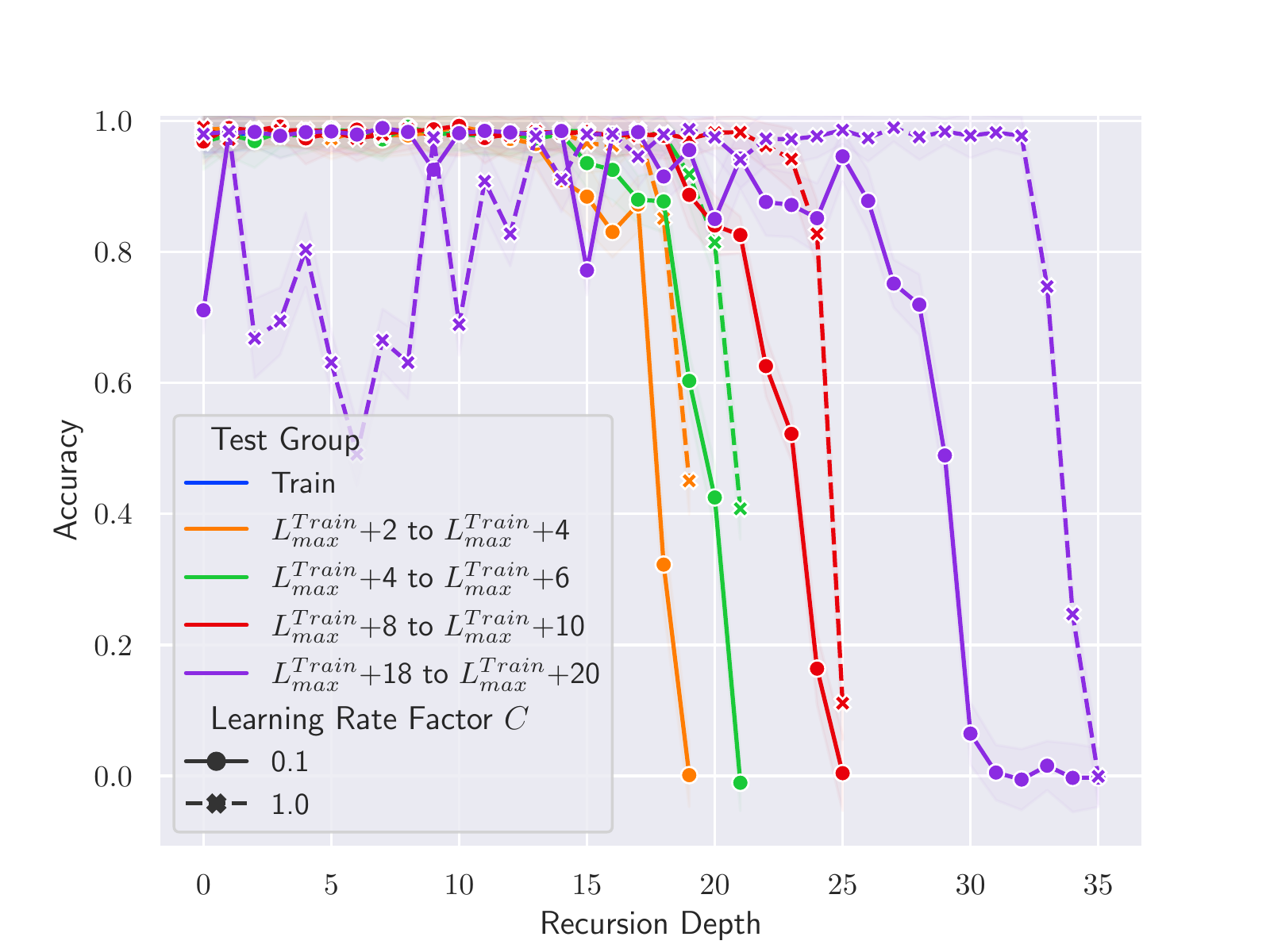}
         \caption{Trained on Binary Strings Up to 131072}
     \end{subfigure}
        % \vspace{-3mm}
    \caption{Accuracy versus recursion depth with maximum training depth constrained to 3: Natural order. }
    \label{fig:nat_perf_depth3}
\end{figure*}

\begin{figure*}[!tb]
     \centering
     \begin{subfigure}[b]{.48\columnwidth}
         \centering
         \includegraphics[width=\columnwidth]{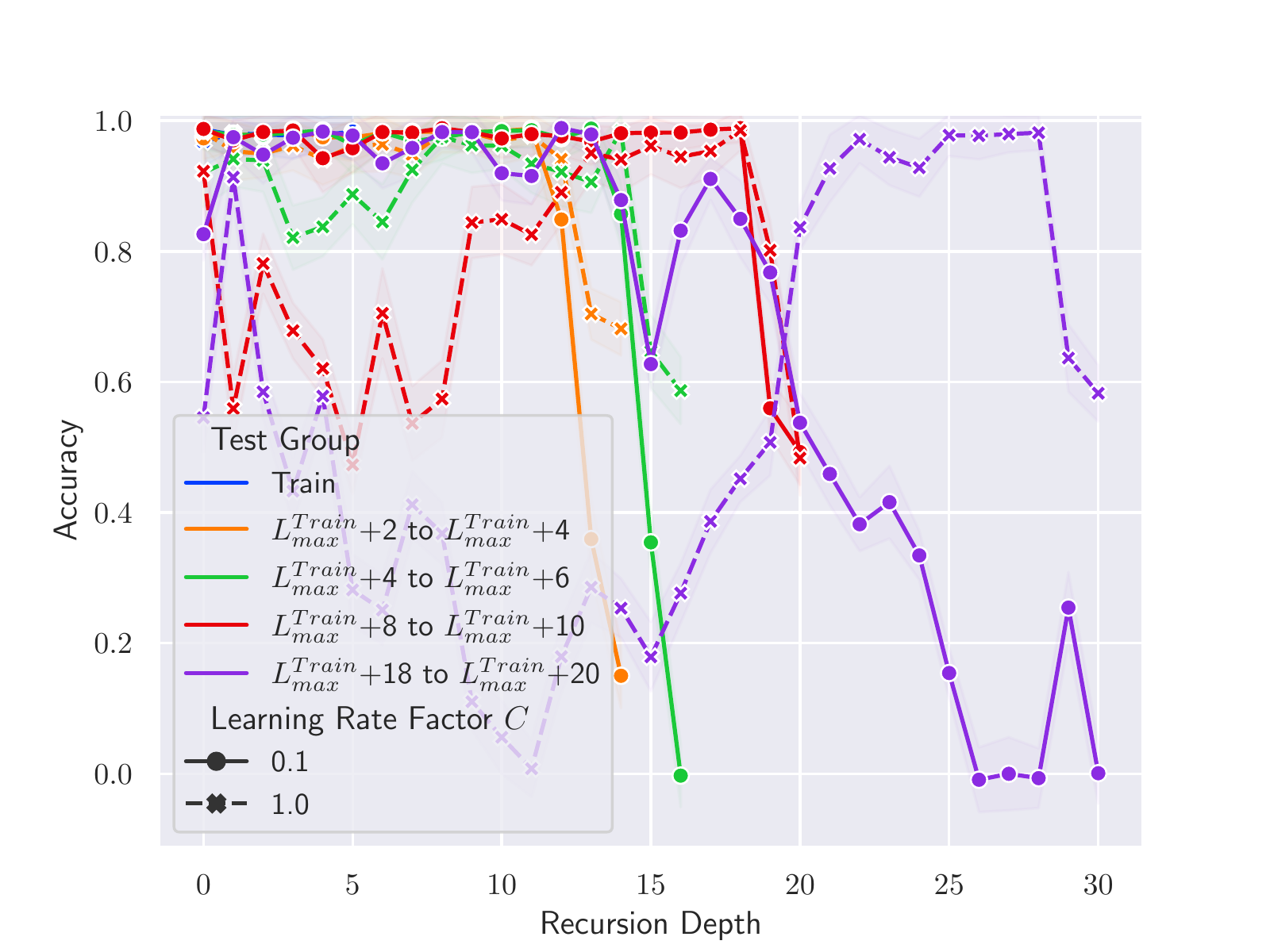}
         \caption{Trained on Binary Strings Up to 2048}
     \end{subfigure}
     \begin{subfigure}[b]{.48\columnwidth}
         \centering
         \includegraphics[width=\columnwidth]{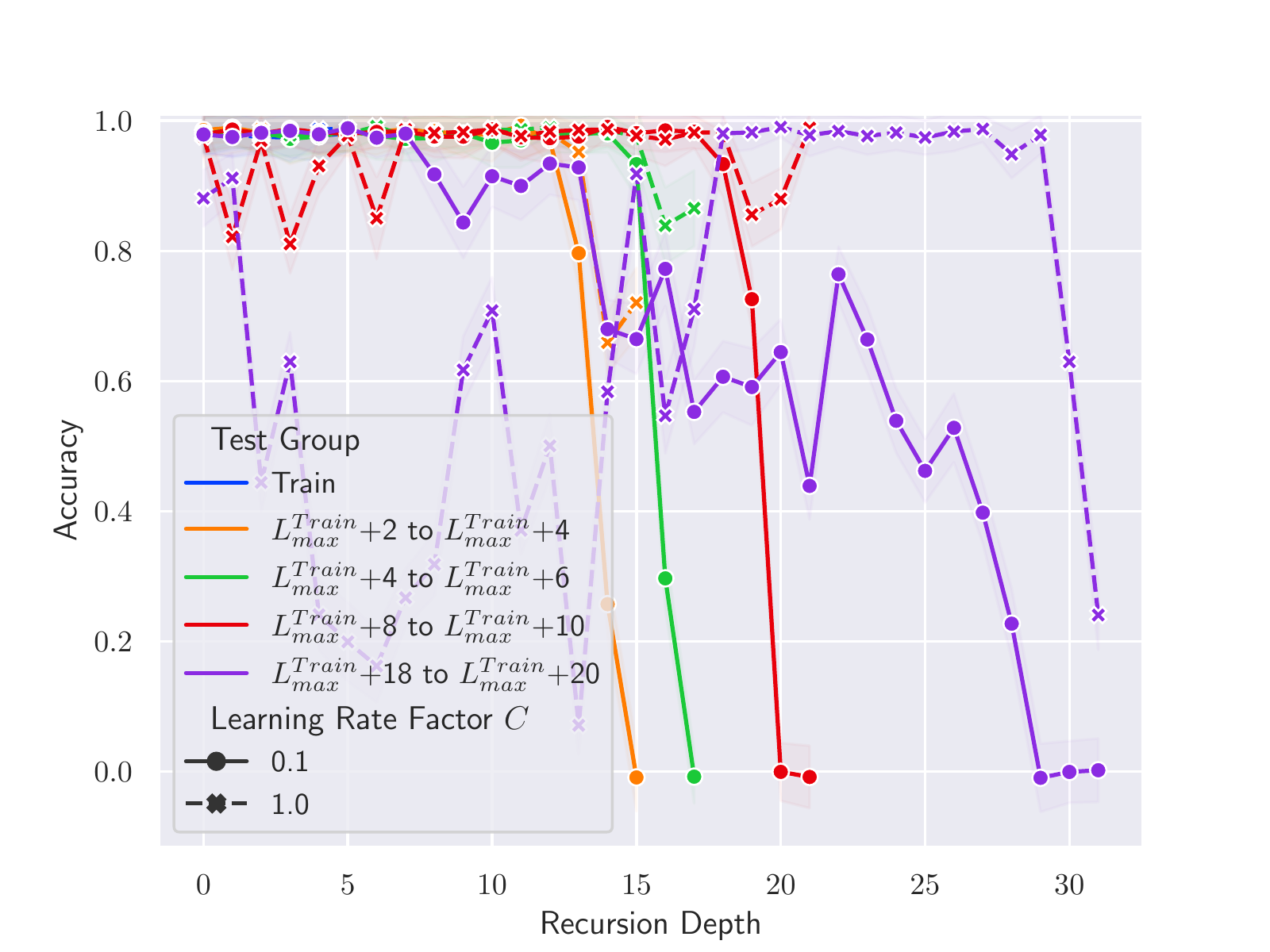}
         \caption{Trained on Binary Strings Up to 4096}
     \end{subfigure}
     \begin{subfigure}[b]{.48\columnwidth}
         \centering
         \includegraphics[width=\columnwidth]{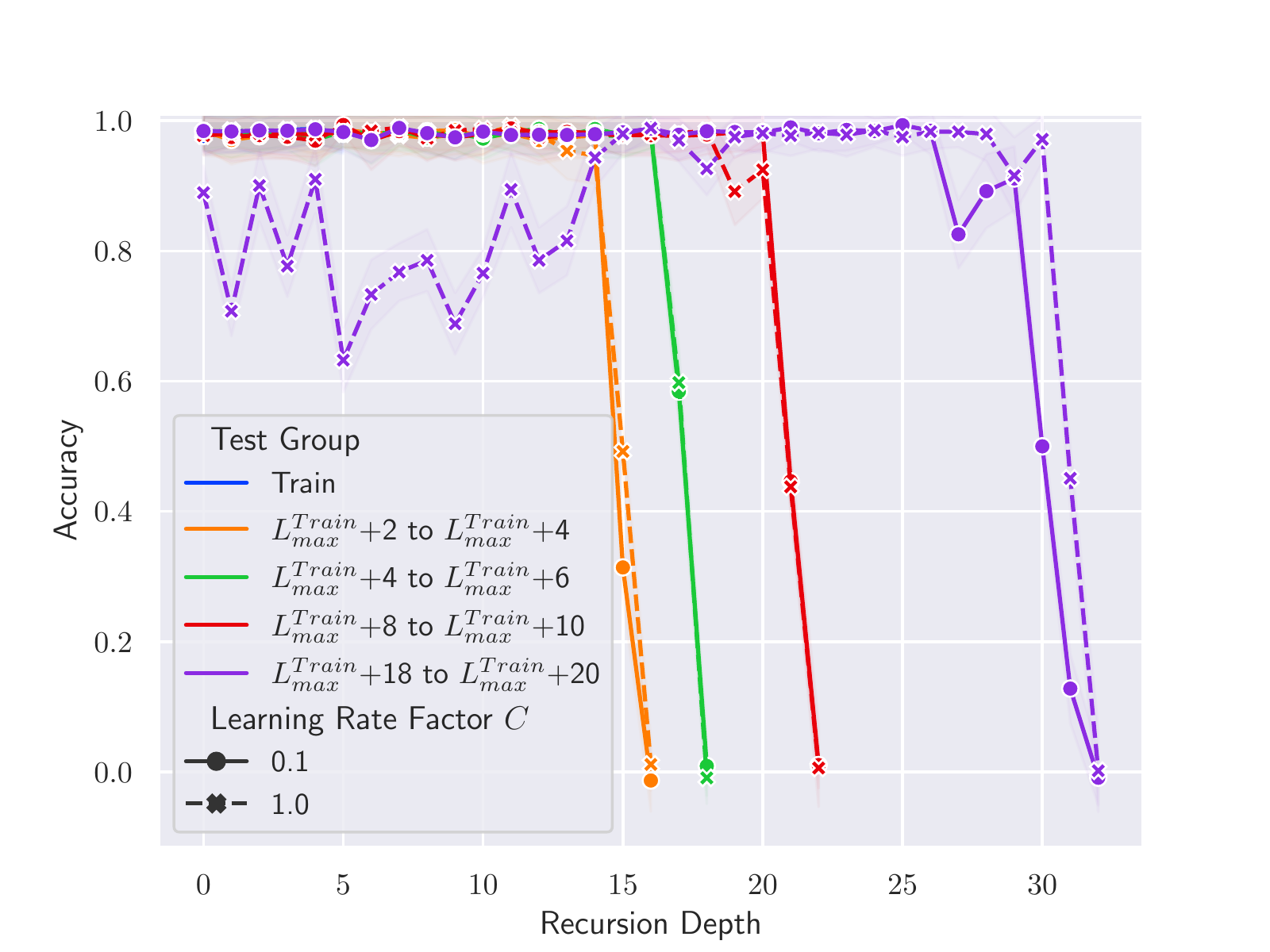}
         \caption{Trained on Binary Strings Up to 8192}
     \end{subfigure}
     \begin{subfigure}[b]{.48\columnwidth}
         \centering
         \includegraphics[width=\columnwidth]{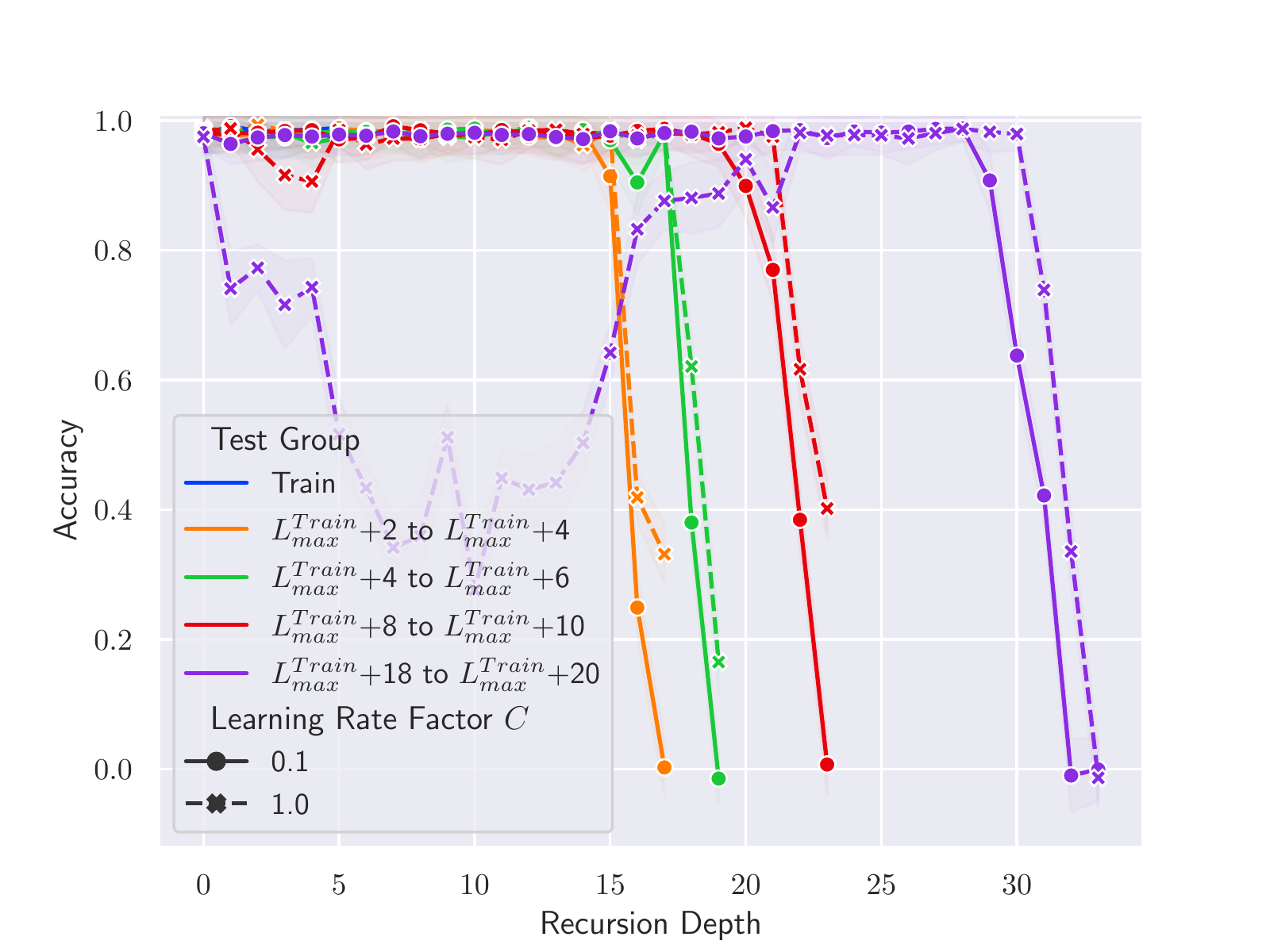}
         \caption{Trained on Binary Strings Up to 16384}
     \end{subfigure}
     \begin{subfigure}[b]{.48\columnwidth}
         \centering
         \includegraphics[width=\columnwidth]{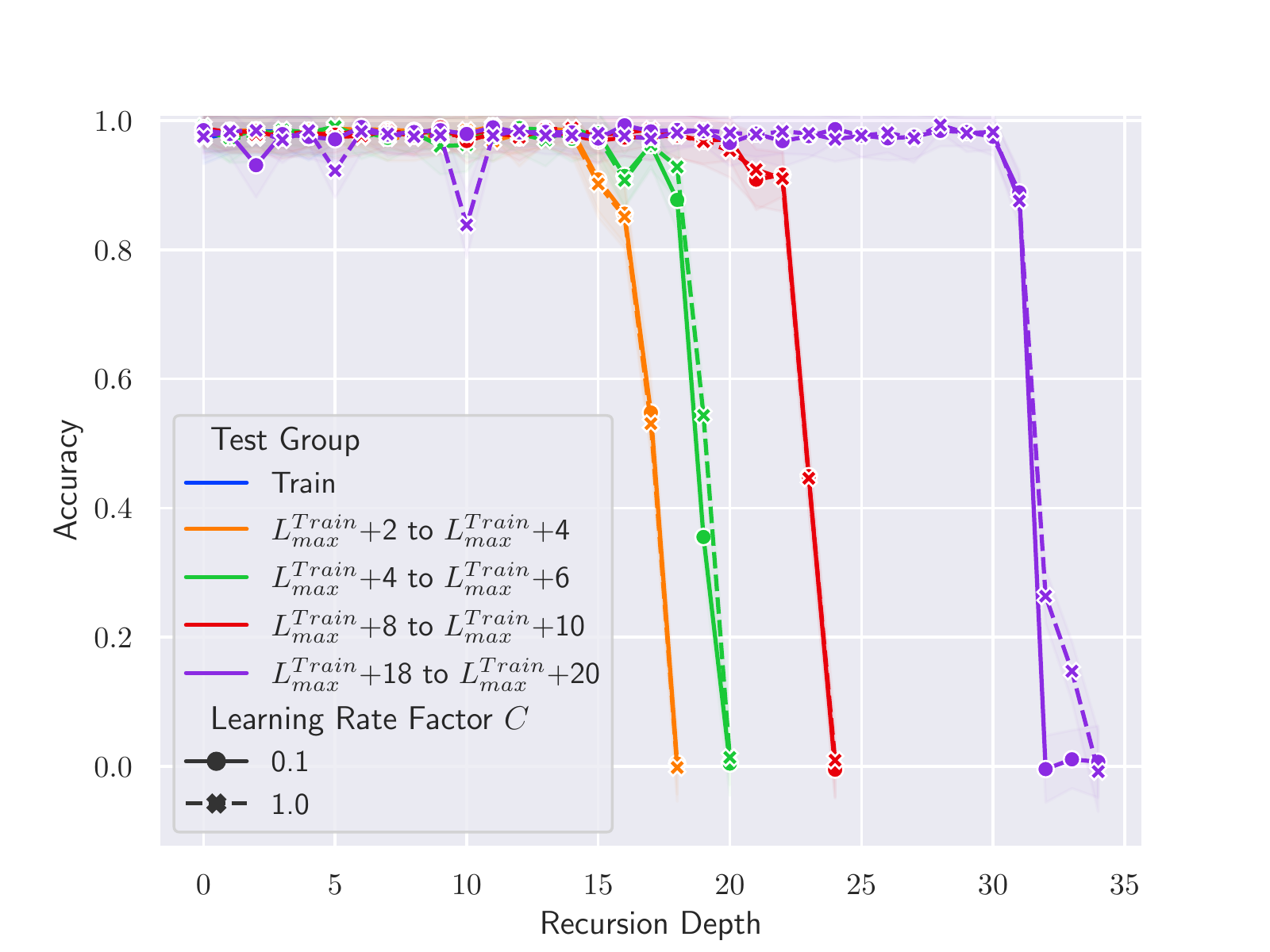}
         \caption{Trained on Binary Strings Up to 32768}
     \end{subfigure}
     \begin{subfigure}[b]{.48\columnwidth}
         \centering
         \includegraphics[width=\columnwidth]{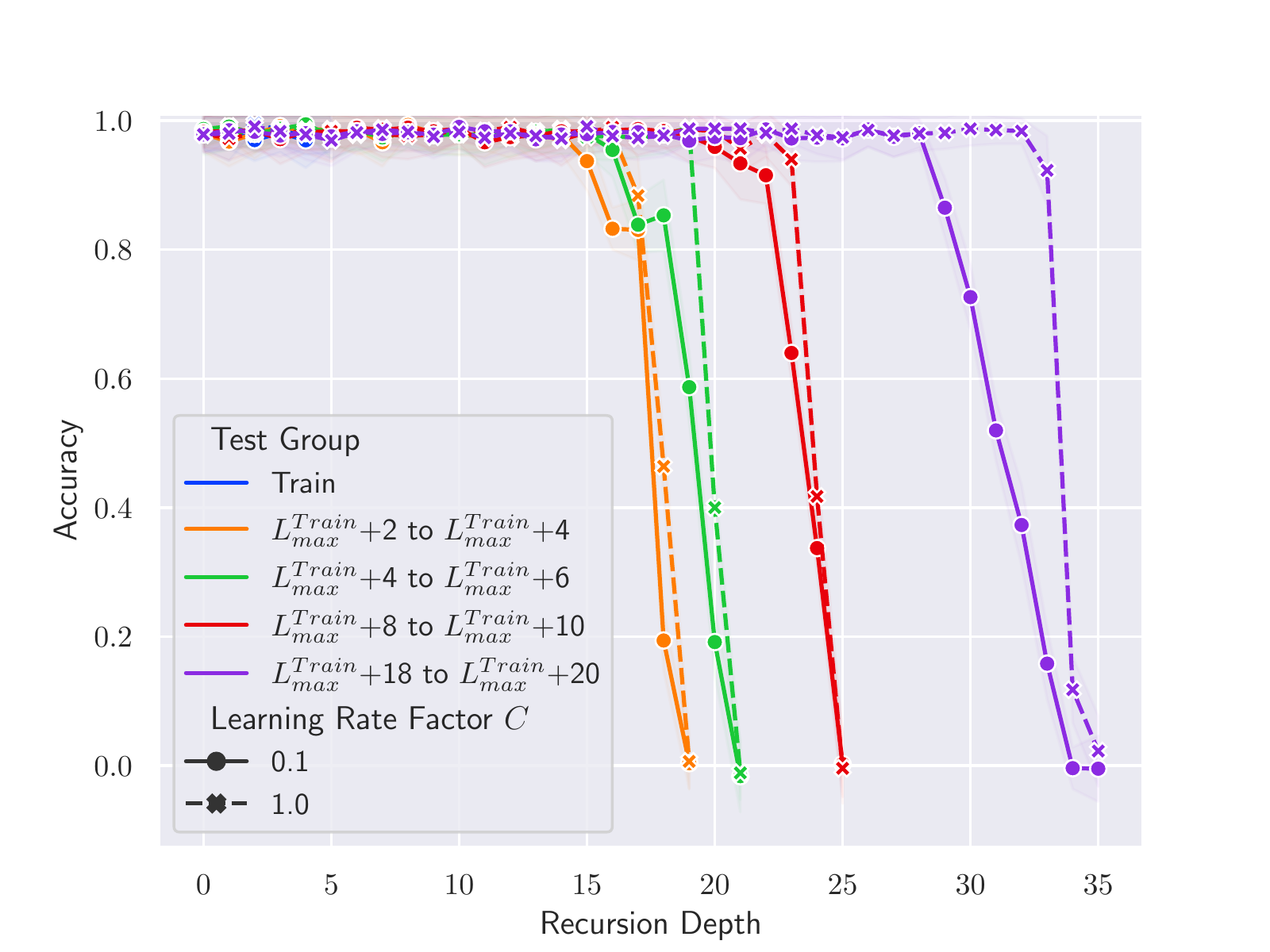}
         \caption{Trained on Binary Strings Up to 65536}
     \end{subfigure}
     \begin{subfigure}[b]{.48\columnwidth}
         \centering
         \includegraphics[width=\columnwidth]{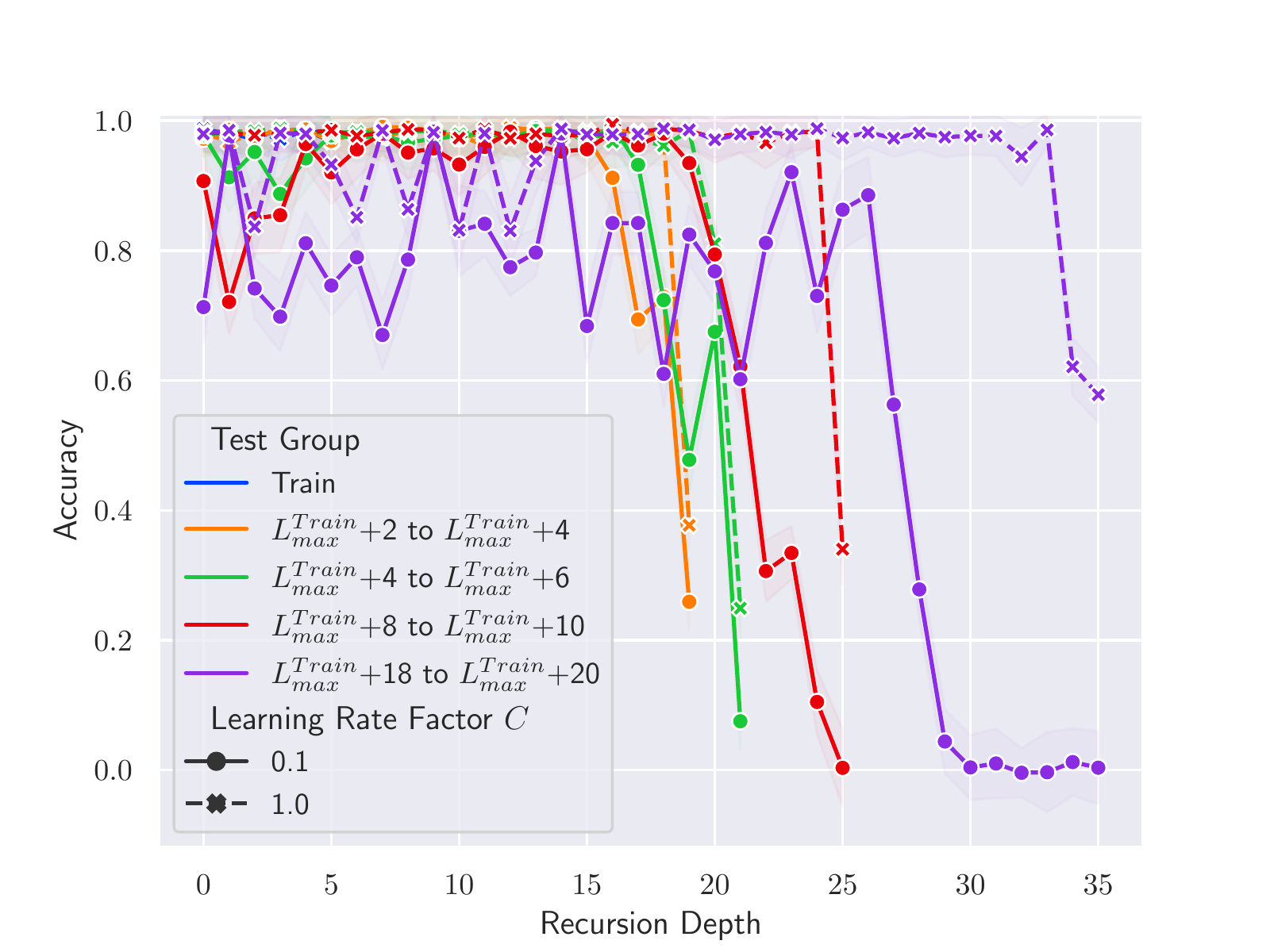}
         \caption{Trained on Binary Strings Up to 131072}
     \end{subfigure}
        % \vspace{-3mm}
    \caption{Accuracy versus recursion depth with maximum training depth constrained to 6: Natural order.}
    \label{fig:nat_perf_depth4}
\end{figure*}

\begin{figure*}[!tb]
     \centering
     \begin{subfigure}[b]{.48\columnwidth}
         \centering
         \includegraphics[width=\columnwidth]{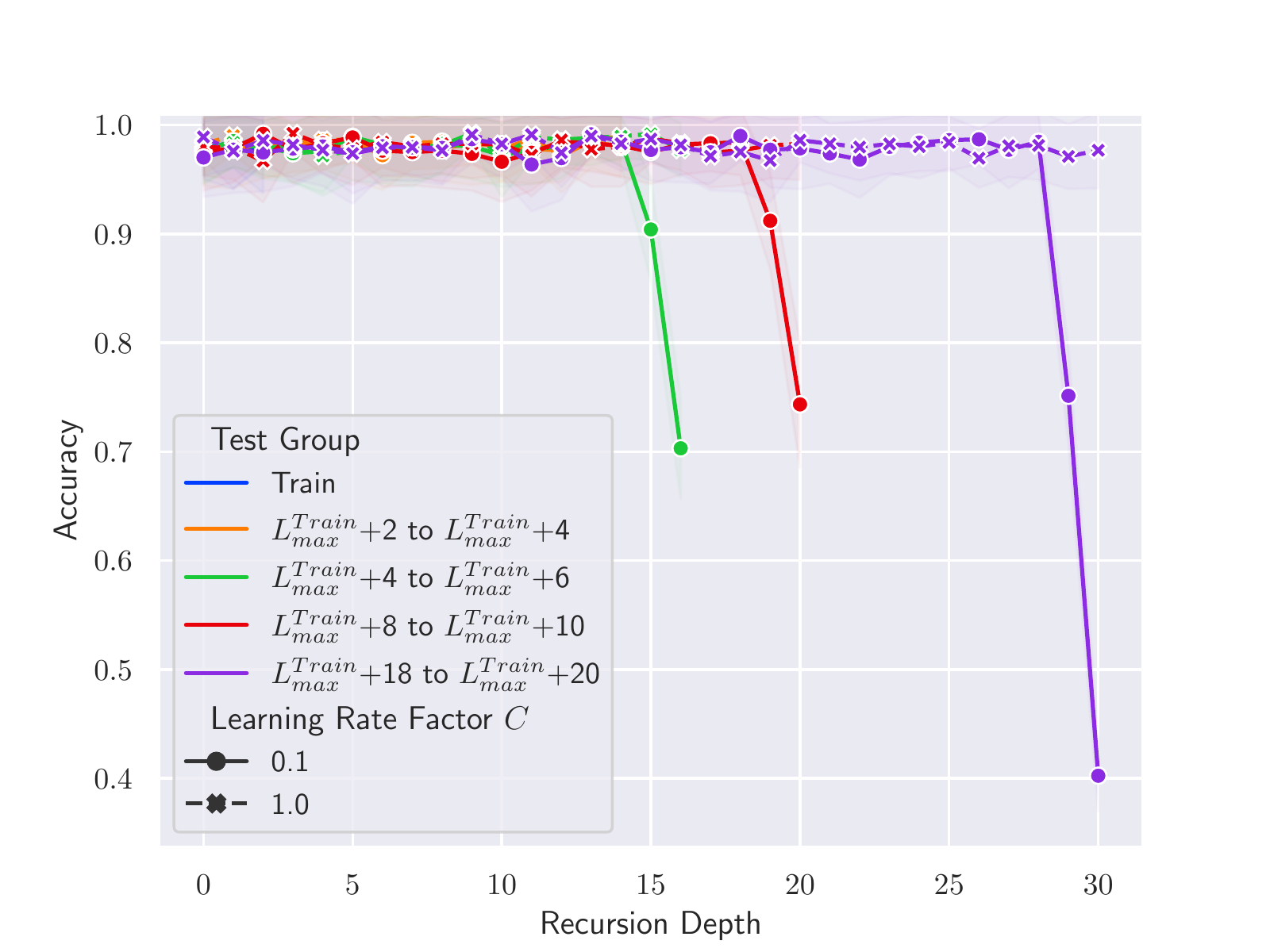}
         \caption{Trained on Binary Strings Up to 2048}
     \end{subfigure}
     \begin{subfigure}[b]{.48\columnwidth}
         \centering
         \includegraphics[width=\columnwidth]{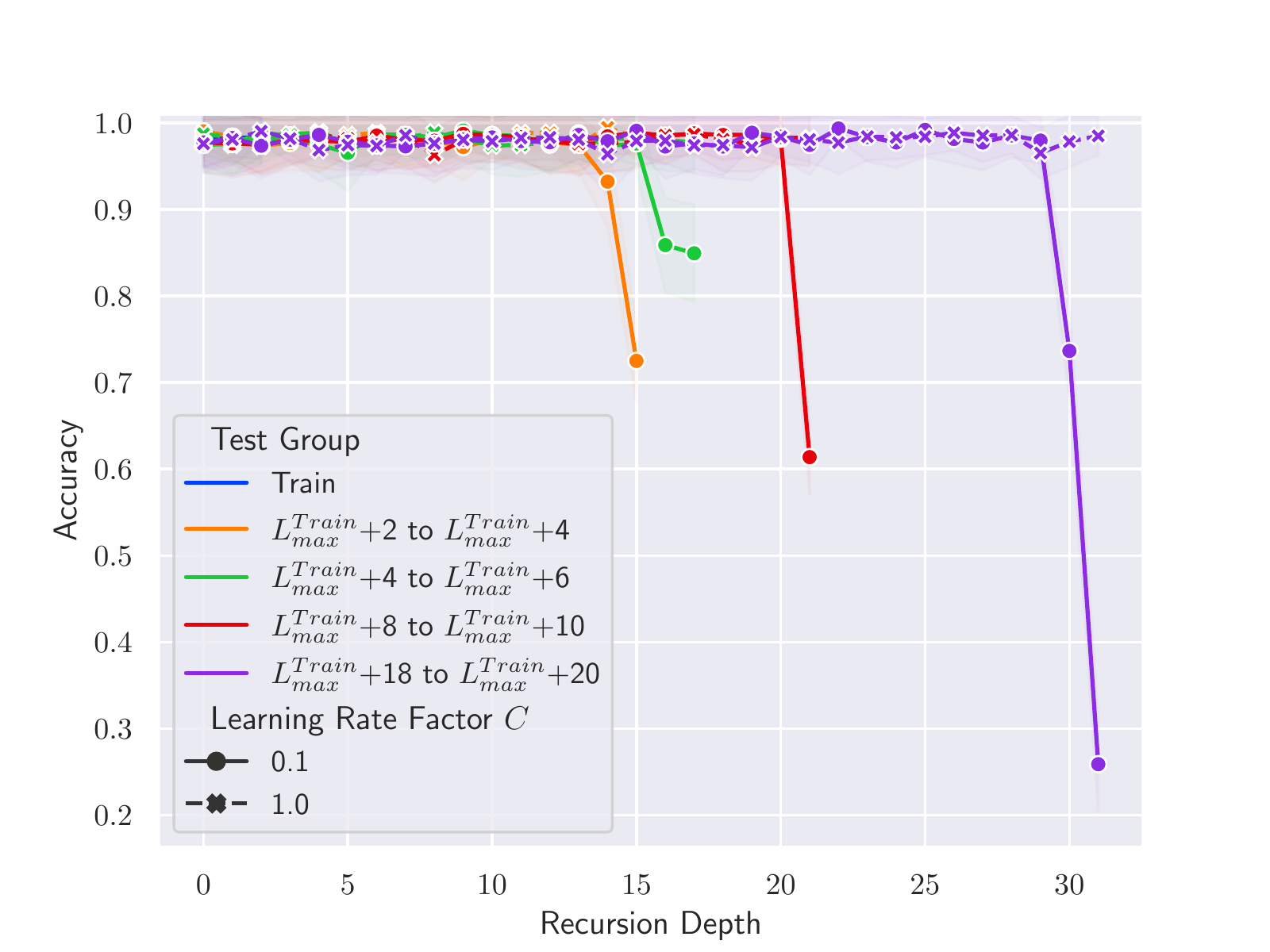}
         \caption{Trained on Binary Strings Up to 4096}
     \end{subfigure}
     \begin{subfigure}[b]{.48\columnwidth}
         \centering
         \includegraphics[width=\columnwidth]{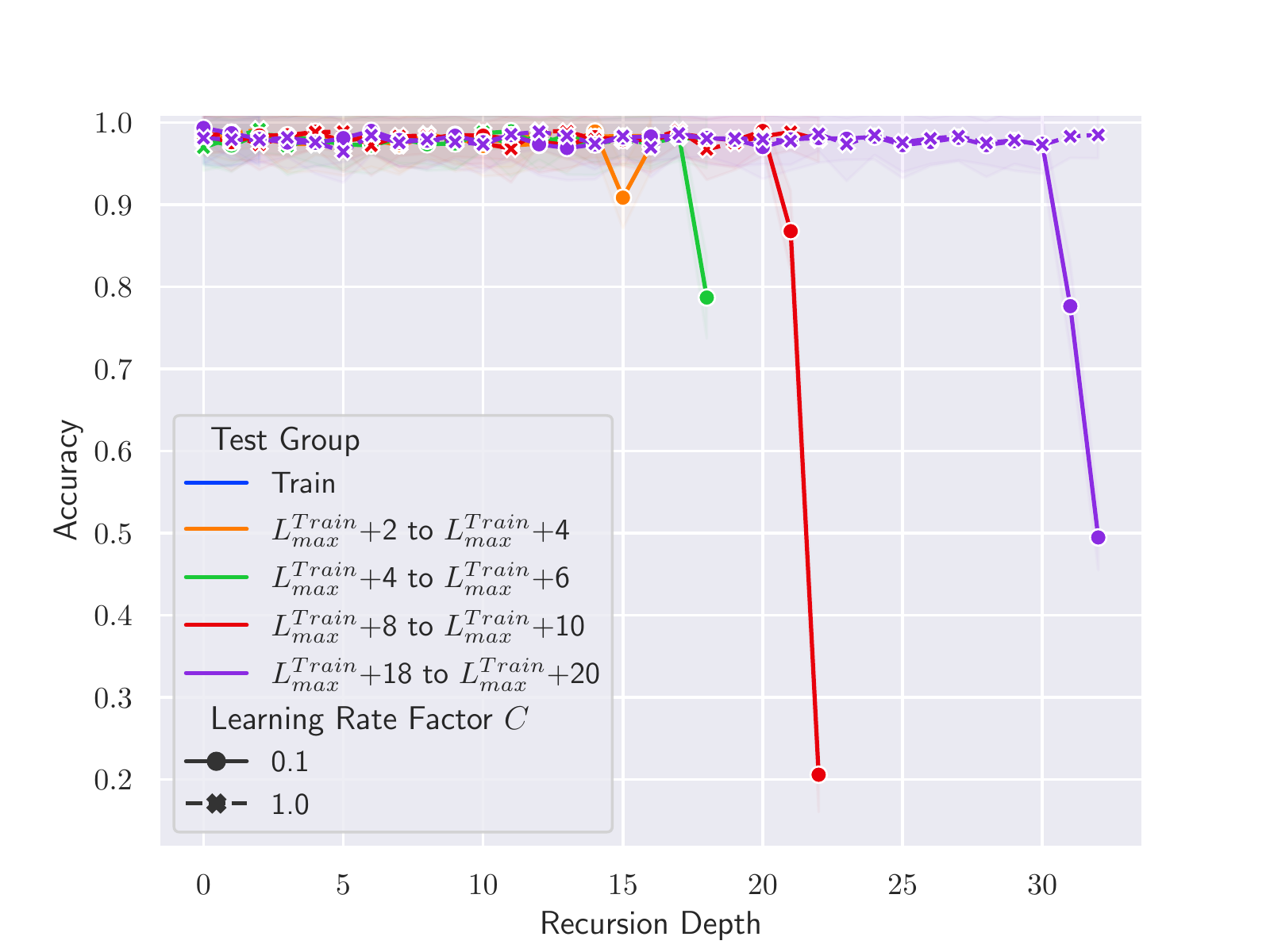}
         \caption{Trained on Binary Strings Up to 8192}
     \end{subfigure}
     \begin{subfigure}[b]{.48\columnwidth}
         \centering
         \includegraphics[width=\columnwidth]{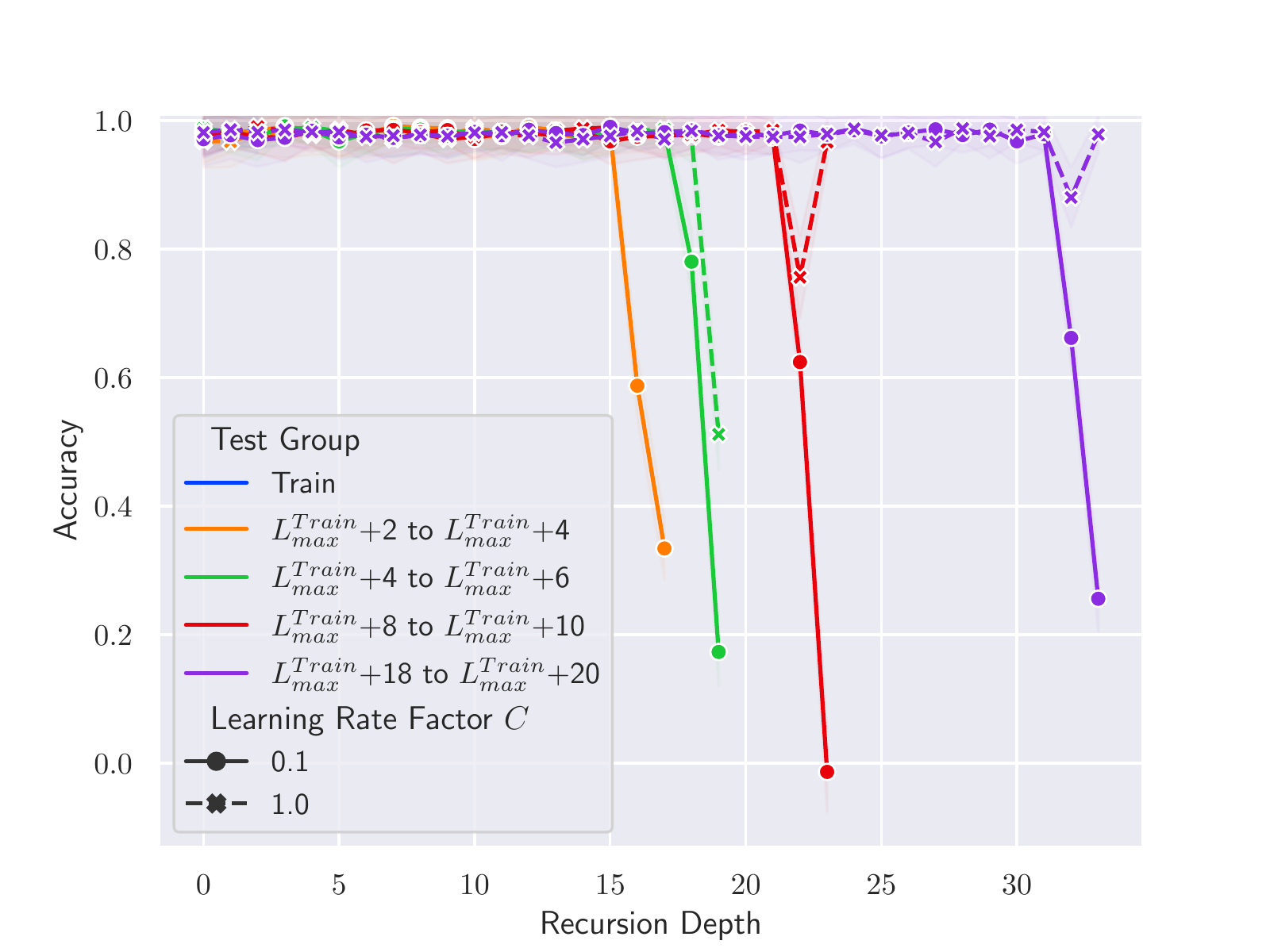}
         \caption{Trained on Binary Strings Up to 16384}
     \end{subfigure}
     \begin{subfigure}[b]{.48\columnwidth}
         \centering
         \includegraphics[width=\columnwidth]{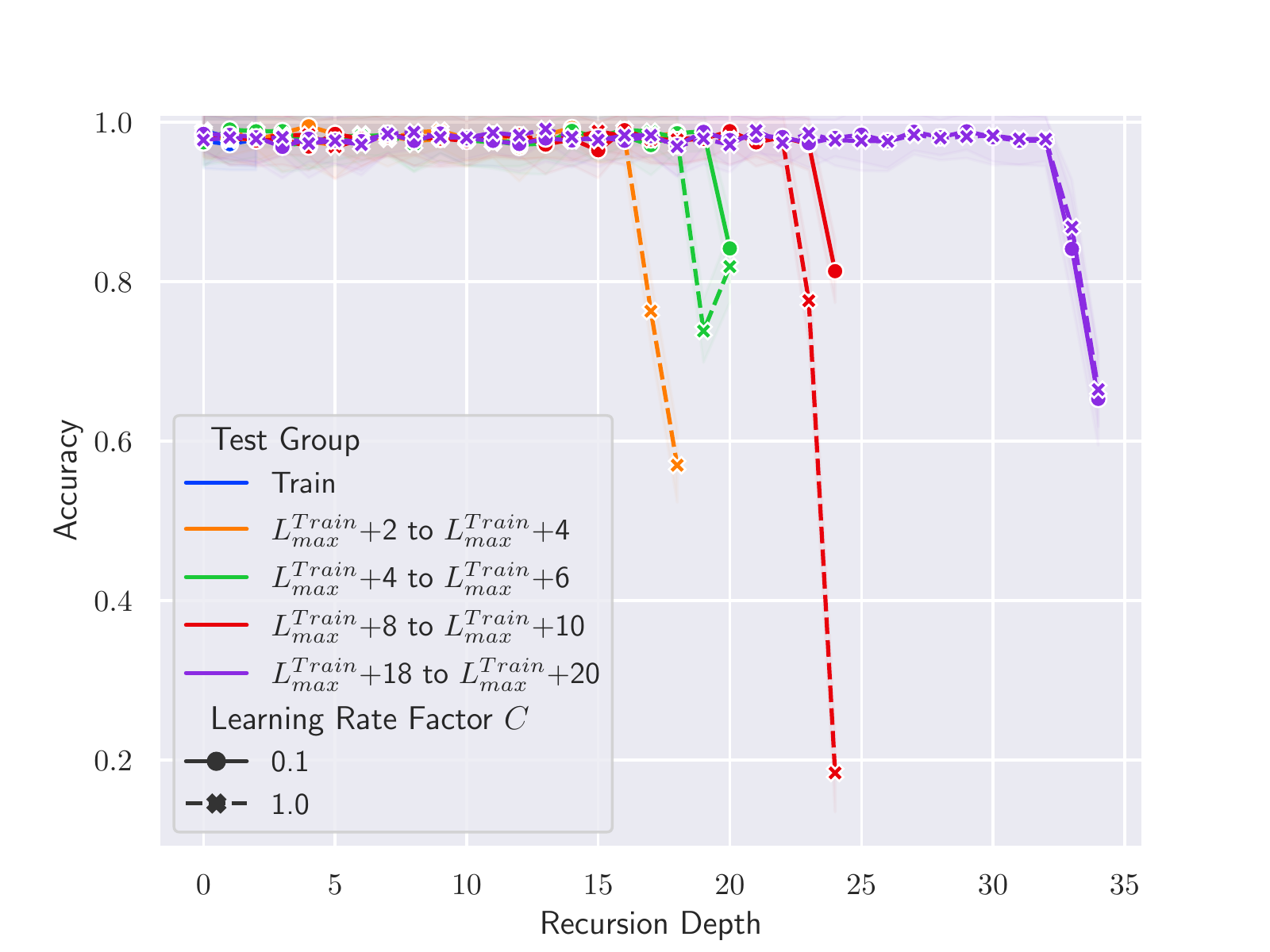}
         \caption{Trained on Binary Strings Up to 32768}
     \end{subfigure}
     \begin{subfigure}[b]{.48\columnwidth}
         \centering
         \includegraphics[width=\columnwidth]{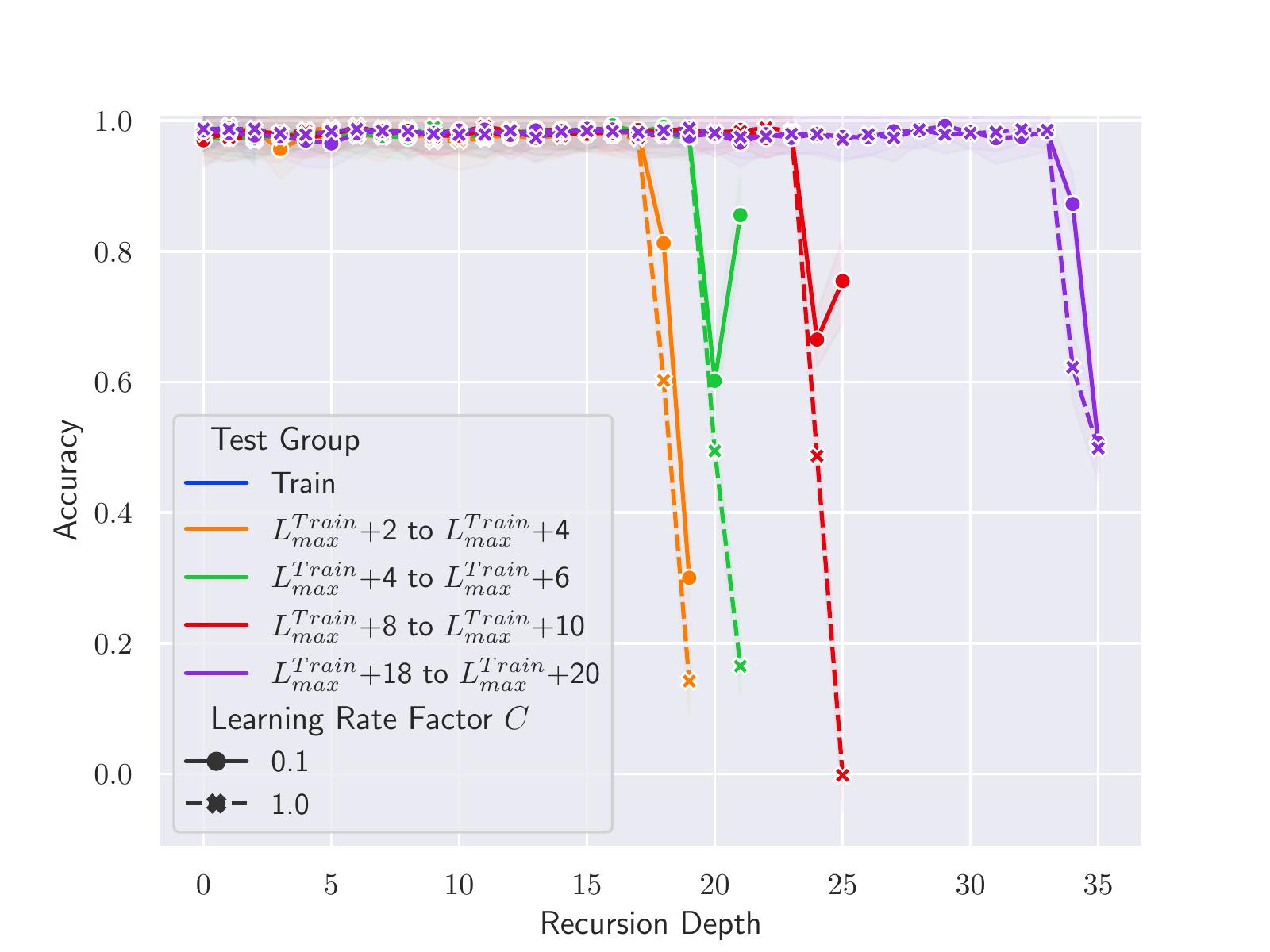}
         \caption{Trained on Binary Strings Up to 65536}
     \end{subfigure}
     \begin{subfigure}[b]{.48\columnwidth}
         \centering
         \includegraphics[width=\columnwidth]{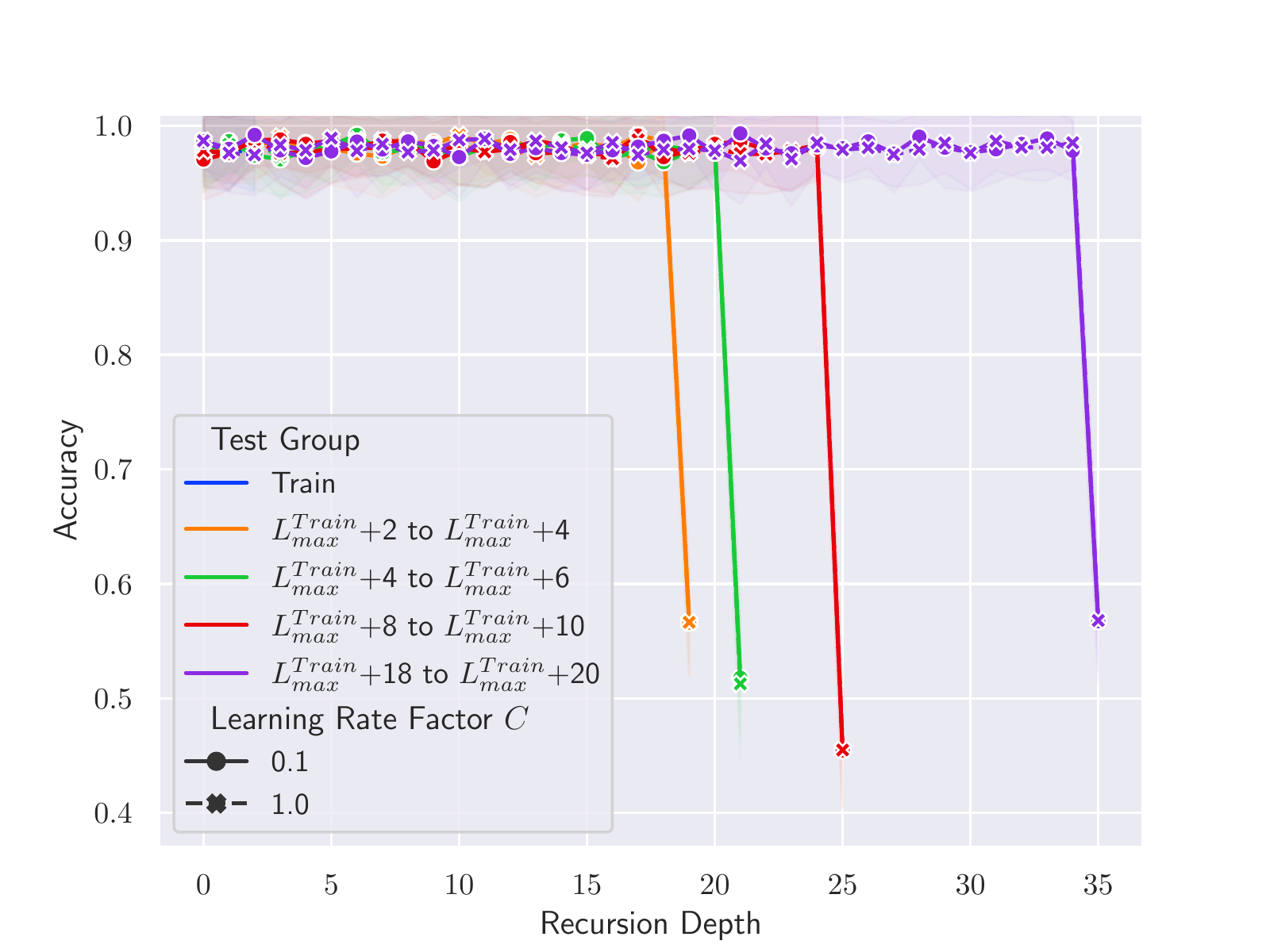}
         \caption{Trained on Binary Strings Up to 131072}
     \end{subfigure}
        % \vspace{-3mm}
    \caption{Accuracy versus recursion depth with maximum training depth constrained to 3: Reverse order. }
    \label{fig:nat_perf_depth5}
\end{figure*}

\begin{figure*}[!tb]
     \centering
     \begin{subfigure}[b]{.48\columnwidth}
         \centering
         \includegraphics[width=\columnwidth]{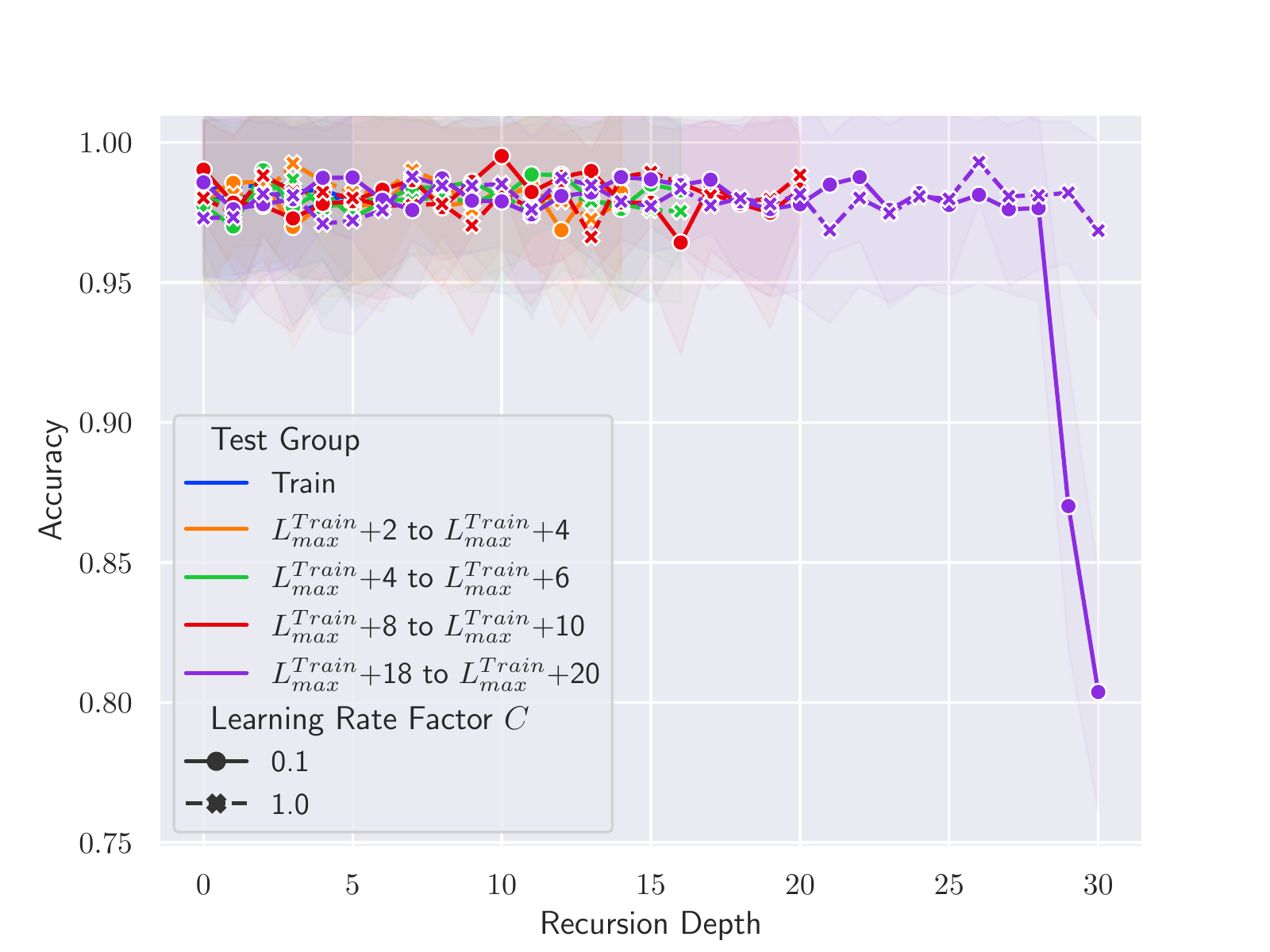}
         \caption{Trained on Binary Strings Up to 2048}
     \end{subfigure}
     \begin{subfigure}[b]{.48\columnwidth}
         \centering
         \includegraphics[width=\columnwidth]{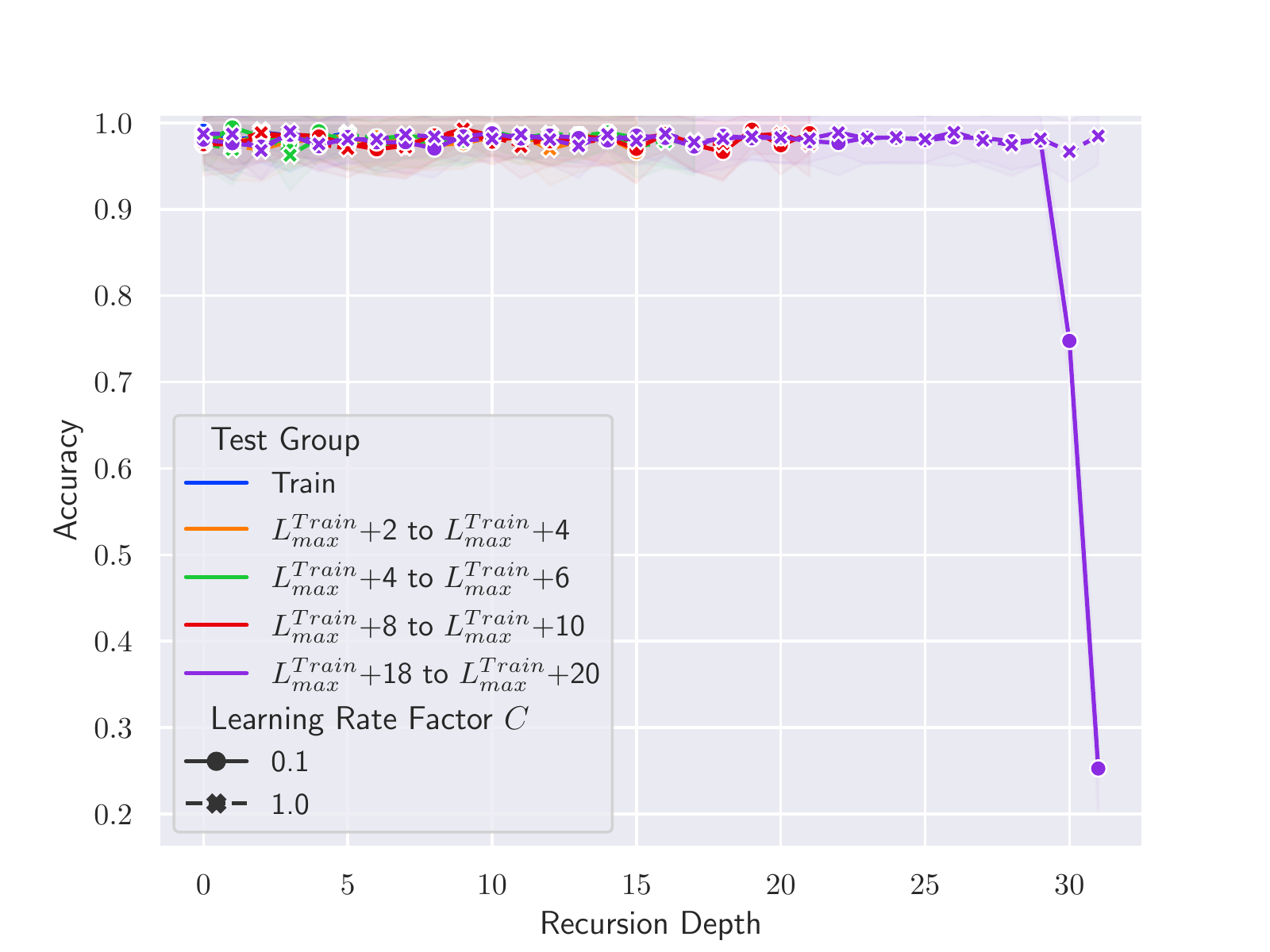}
         \caption{Trained on Binary Strings Up to 4096}
     \end{subfigure}
     \begin{subfigure}[b]{.48\columnwidth}
         \centering
         \includegraphics[width=\columnwidth]{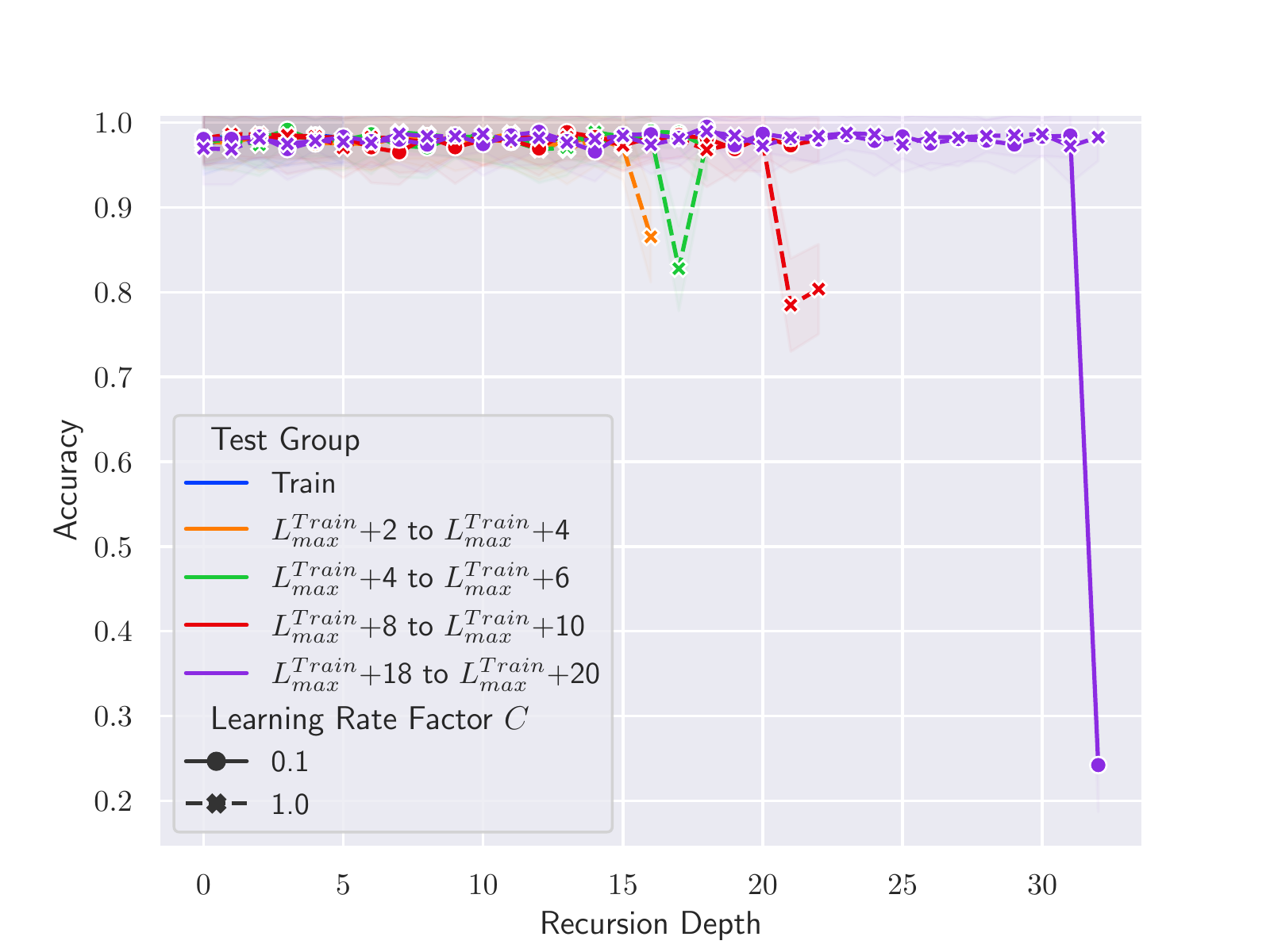}
         \caption{Trained on Binary Strings Up to 8192}
     \end{subfigure}
     \begin{subfigure}[b]{.48\columnwidth}
         \centering
         \includegraphics[width=\columnwidth]{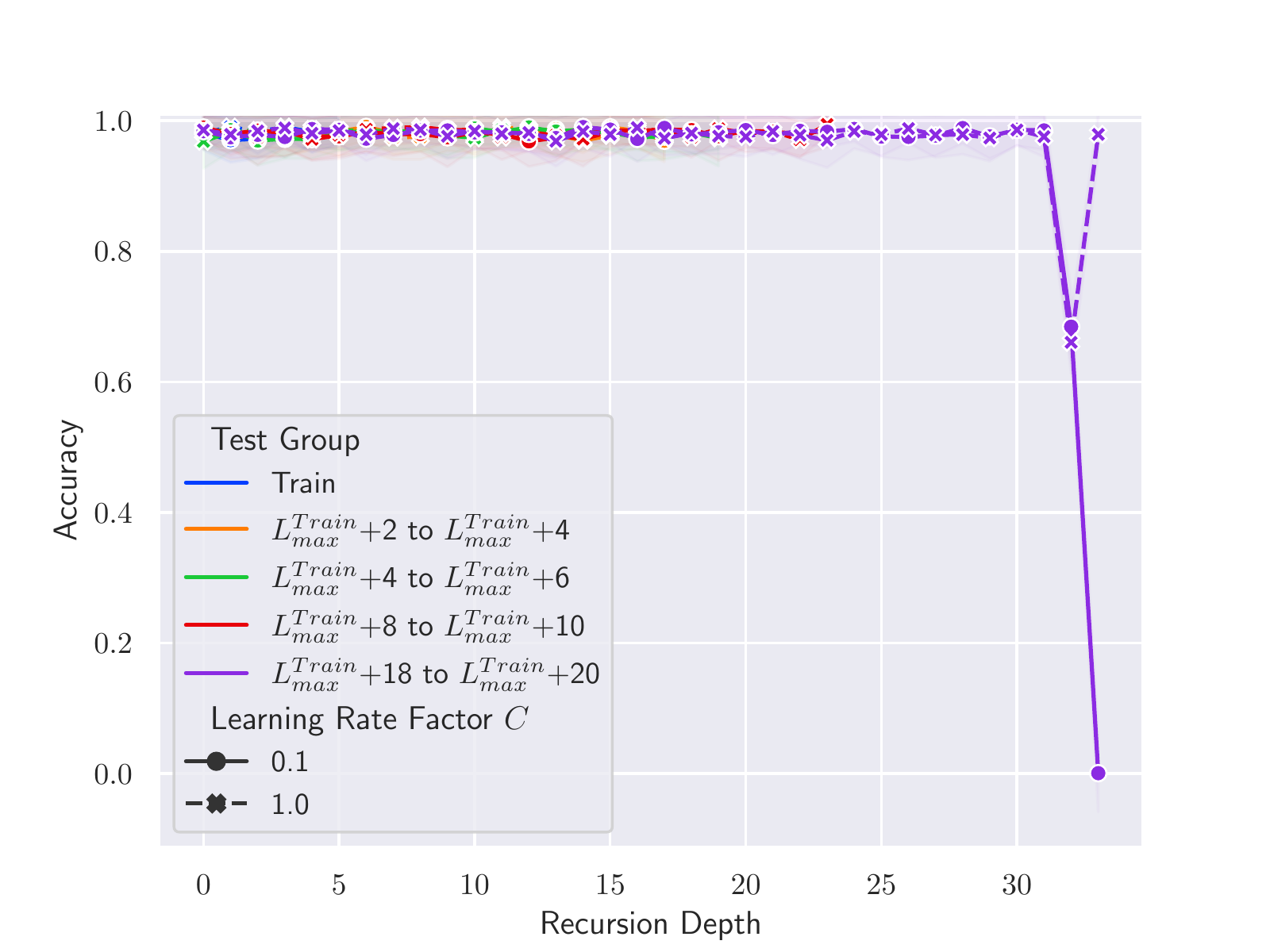}
         \caption{Trained on Binary Strings Up to 16384}
     \end{subfigure}
     \begin{subfigure}[b]{.48\columnwidth}
         \centering
         \includegraphics[width=\columnwidth]{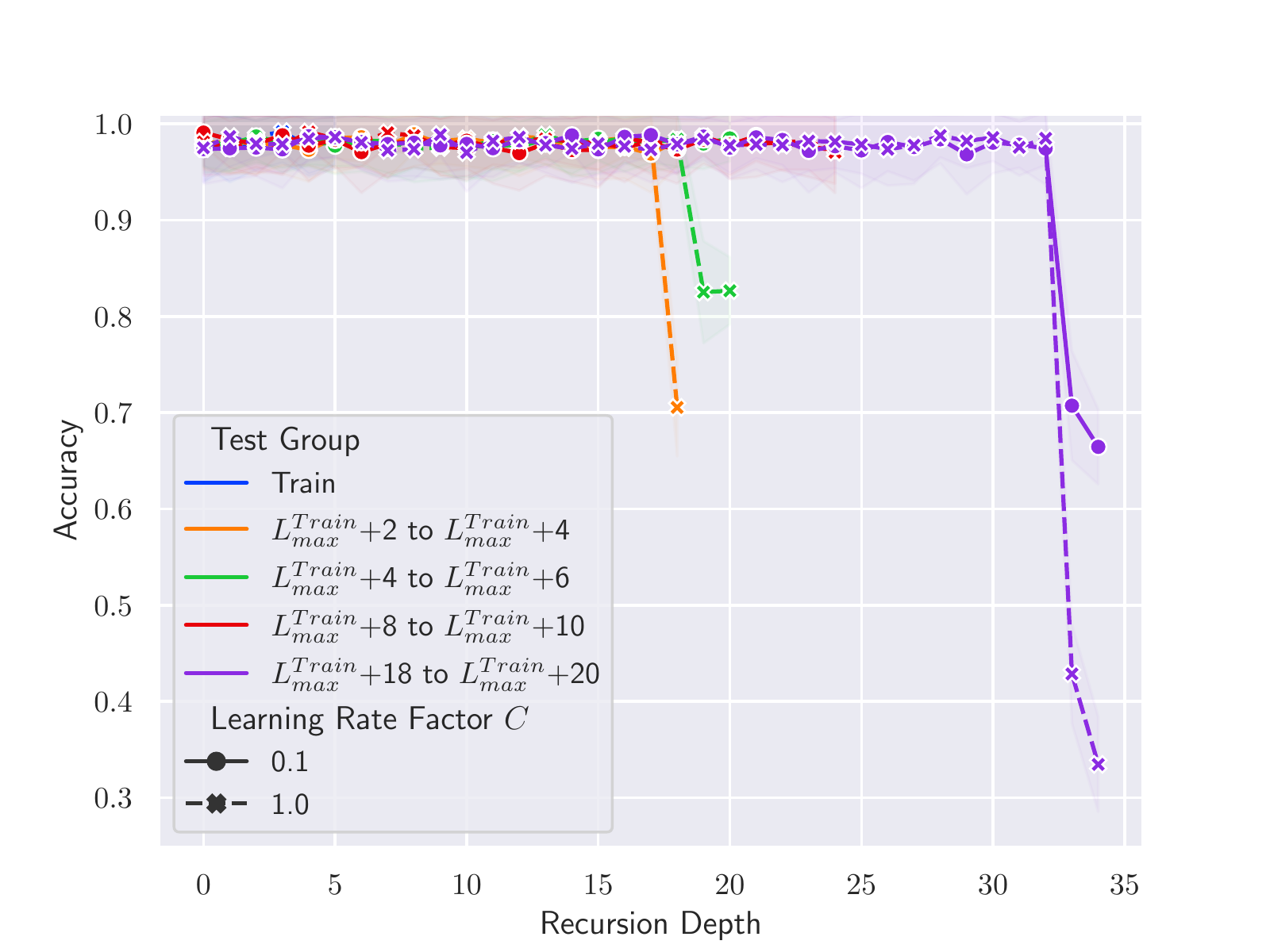}
         \caption{Trained on Binary Strings Up to 32768}
     \end{subfigure}
     \begin{subfigure}[b]{.48\columnwidth}
         \centering
         \includegraphics[width=\columnwidth]{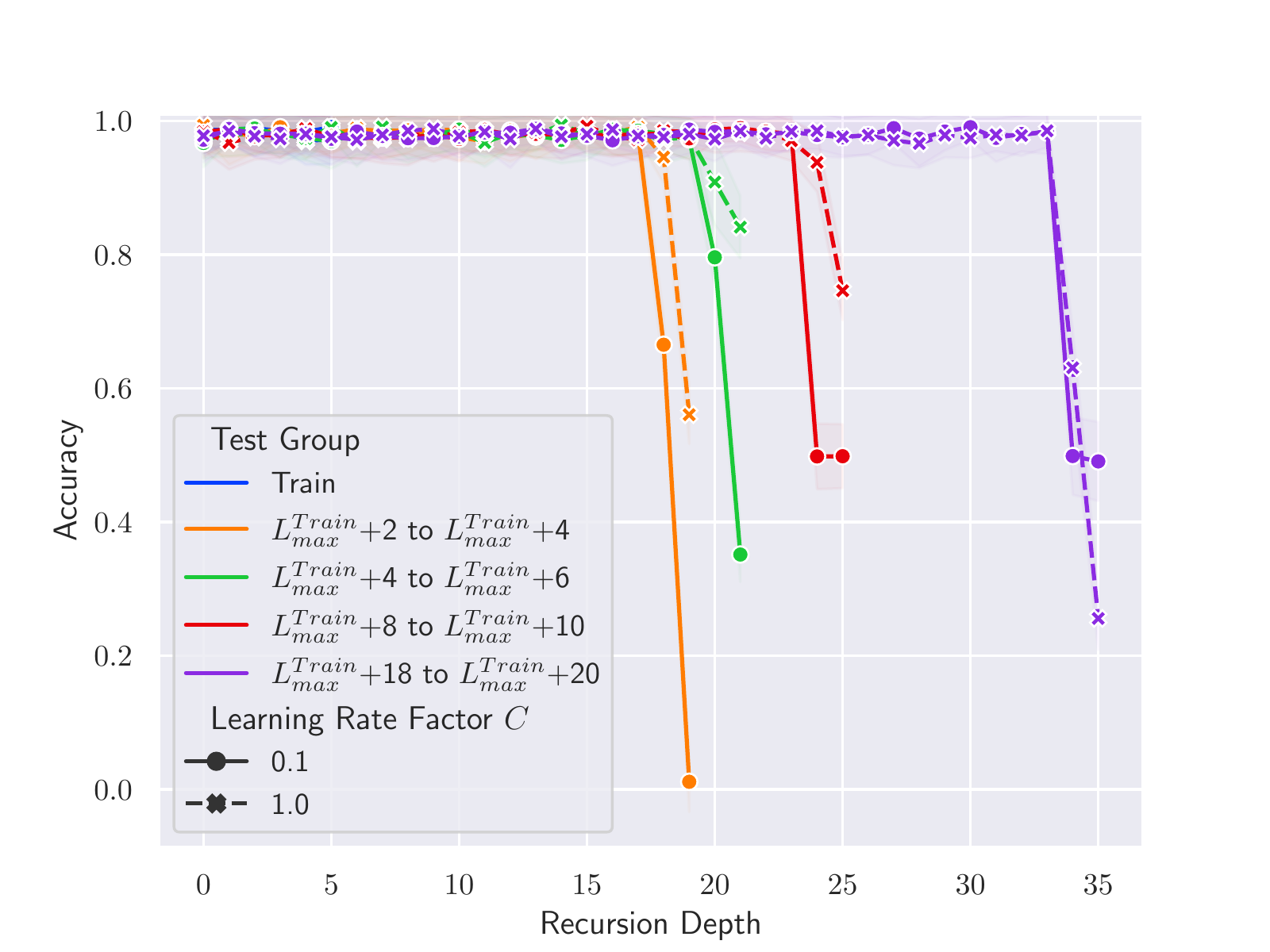}
         \caption{Trained on Binary Strings Up to 65536}
     \end{subfigure}
     \begin{subfigure}[b]{.48\columnwidth}
         \centering
         \includegraphics[width=\columnwidth]{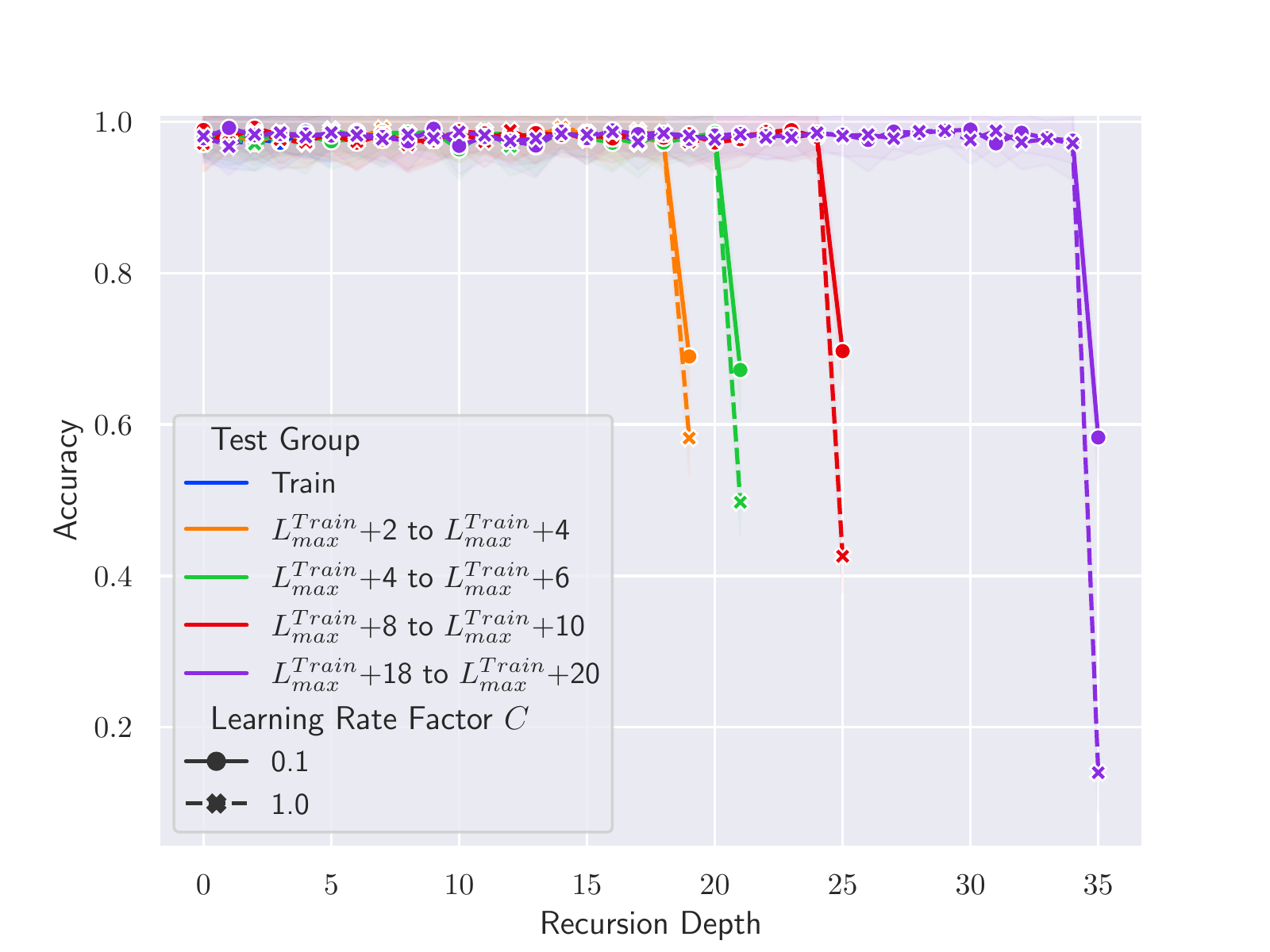}
         \caption{Trained on Binary Strings Up to 131072}
     \end{subfigure}
        % \vspace{-3mm}
    \caption{Accuracy versus recursion depth with maximum training depth constrained to 6: Reverse order.}
    \label{fig:nat_perf_depth6}
\end{figure*}

Figures~\ref{fig:nat_perf_depth} through~\ref{fig:nat_perf_depth6}
show detailed extrapolation performance for the binary successor task based on recursion depth. The error bars indicate the standard deviation
across three runs with different random seeds.

\section{Loss of Tree Traversal Reduction}
\begin{figure*}
    \centering
\includegraphics[width=.9\columnwidth]{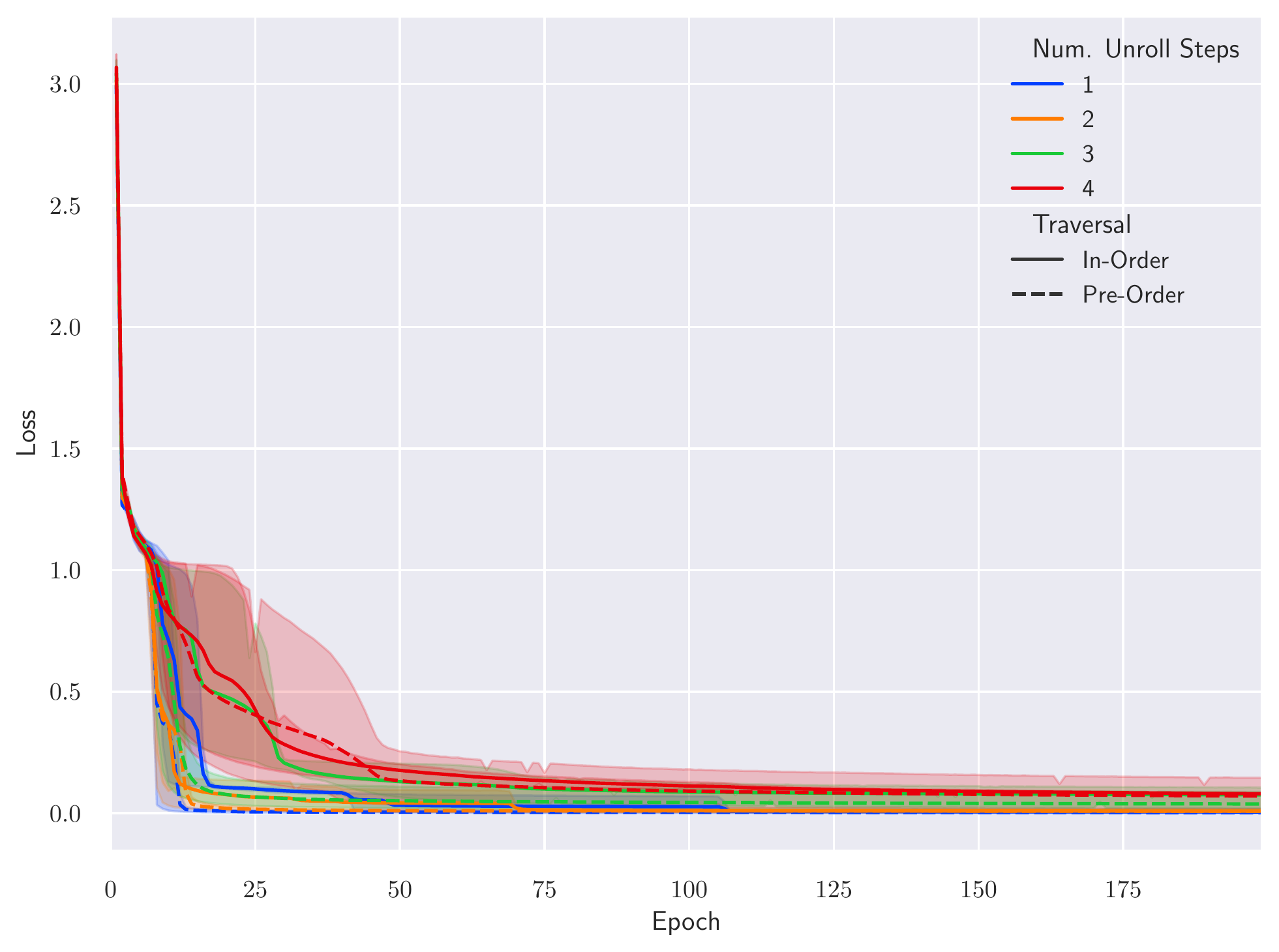}
    \caption{Tree Traversal Loss vs. Epochs}
    \label{fig:loss}
\end{figure*}

Figure ~\ref{fig:loss} shows the change of training loss during training. Models learning higher-order reductions would converge to a higher value compared with lower-order reductions. 
\newpage
\section{Binary Successor Performance By Depth}
\label{app:depth}
Figures~\ref{fig:nat_perf_depth} through~\ref{fig:nat_perf_depth6}
show detailed extrapolation performance for the binary successor task based on recursion depth. The error bars indicate the standard deviation
across three runs with different random seeds.

\end{document}

%% file: intro.tex
\section{Introduction}
\label{sec:intro}

A revolution in neural methods for programming languages tasks is underway.
Once confined to the realm of symbolic methods,
some of the most performant tools for 
synthesizing~\cite{chaudhuri2021neurosymbolic, chen2021codex, Li2020alphacode, palm}, % small sample ...
repairing~\cite{Gupta17, Saha17, xia2023automated}, % small sample ...
%understanding~{\color{purple} cite},
%debugging~{\color{purple} cite},
and even formally verifying~\cite{Agrawal23icse-demo, yang2019learning, First22icse, Sanchez-Stern22passport, proverbot9001, first2023baldur} % small sample ... 
programs now rest in part or in whole upon neural foundations.

But how sturdy are these foundations?
At the core of many of these tools are transformer-based
large language models~\cite{chen2021codex, palm, first2023baldur}.
It is an open question to what degree these models are simply 
repeating program syntax, and to what degree they have some model of 
program \emph{semantics}---how programs behave and what they mean.
State-of-the-art language models still rely on
tricks like chain of thought prompting~\cite{wei2022chain} and scratchpadding~\cite{nye2021show} to approximate program semantics. Even models trained on code often need to be finetuned to solve specific tasks instead of used in a multitask fashion~\cite{chen2021codex, allal2023santacoder, li2023starcoder}.

In this paper, we investigate the degree to which small transformer~\cite{allyouneed} models can learn to model the semantics of an important class of programs: \textit{structural recursion}. A program is an example of structural recursion if it is defined over some data structure (say, binary trees) by recursively calling itself
over smaller substructures (say, left and right subtrees).
Structural recursion is at the heart of 
important programming and theorem proving tasks for which neural methods
still lag behind symbolic methods, like inferring semantic relations between
datatypes~\cite{ringer2021pldi, ringer2021proof}.

Drawing on previous work on reverse engineering neural networks~\cite{wang2022interpretability, mechanic_intepret_grokking, reverse_engineering_group_op}, we train small transformer models to solve structural recursion problems and explore the extent to which the models are able to solve the problems. Our emphasis in particular is on understanding the ways in which the algorithms learned by transformers \textbf{fail to correctly solve the tasks for which they were trained}.

We make the following contributions: (1) We conduct a comprehensive \textbf{empirical study} on two representative types of recursive tasks for transformers (Section~\ref{sec:overview}): the binary successor function, which has a single recursive subcase, and a tree traversal, which has two recursive subcases. Our study investigates different aspects and granularities of the models' behaviors on these tasks. For example, we train a model on an atomic subtask of a tree traversal and find that it performs well on that subtask, even though a model trained with the same setup on the end-to-end traversal fails. (2) We describe a \textbf{conceptual framework} based on abstract-state machines (ASMs) that allows us to analyze programs, examining transformer behavior within a formal model of its \textit{practical} computation (Section~\ref{sec:model}).
% implemented either in a given symbolic programming language or using a learned transformer model, as sequential application of appropriately defined \textit{state updates} at the right level of abstraction (Section~\ref{sec:model}). This framework was originally introduced as a formal model of \textit{practical} computation, encompassing both low-level (processor instructions) and high-level (virtual machine) architectures. We believe that it provides a foundation for further research in the field.
(3) We reconstruct the \textbf{mechanisms} of transformers in learning the recursive tasks and identify their flaws (Section~\ref{sec:evaluation}). By reconstructing these mechanisms, we are able to \textit{correctly predict} specific classes of inputs on which the models fail---correctly predicting up to 91\% of failures! Our analysis also reveals evidence of differences in the learned algorithms and under different hyperparameter set-ups. 
% \end{enumerate}

% \begin{enumerate}
%     \item {\color{purple} something about empirical analysis, including any new methodologies and results (Sections blah)}
%     \item {\color{purple} something about conceptual analysis, including any new results (Sections blah)}
%     \item {\color{purple} something about reconstructed algorithms and what that tells us (Sections blah)}
% \end{enumerate}

%% file: overview.tex
\section{Representing Structural Recursion}
\label{sec:goals}

For this work, we are interested in how transformer models learn to approximate \textit{structural recursion}: a restricted but powerful class of recursive functions that are defined in a structurally decreasing manner, and so must
terminate. This is a popular representation of recursion in the program
and proof literature because it is very expressive, but easy to reason about.

Consider, for example, defining a recursive function that adds two positive natural numbers. We start by defining the datatype representing those numbers---a \textit{unary} encoding of Peano natural numbers---as in Figure~\ref{fig:sub:unary} (the syntax here comes from a proof tool called Coq)).

% \begin{lstlisting}
%   Inductive peano =
%   | I : peano (* base case *)
%   | S : $\forall$ (n : peano), peano. (* inductive case *)
% \end{lstlisting}
\begin{figure}[h]
\begin{minipage}[b]{.43\textwidth}
\small
\begin{lstlisting}
Inductive peano =
| I : peano (* base case *)
| S : $\forall$ (n : peano), peano. (* inductive case *)
\end{lstlisting}
\subcaption{\textsl{Unary encoding of Peano natural numbers.}\label{fig:sub:unary}}
% \textit{Unary encoding of Peano natural numbers.}

\end{minipage}
\begin{minipage}[b]{.55\textwidth}
\small
    \begin{lstlisting}
      Fixpoint add n m :=
        match n with (* break into cases *)
        | I => S m (* if n is one, return S m *)
        | S p => S (add p m) (* otherise, recurse *)
        end.
\end{lstlisting}
\subcaption{\textsl{Addition using structural pattern matching.}\label{fig:sub:addition_pattern}}
\end{minipage}
\caption{Representing structural recursion.}
\end{figure}
% This \textit{inductive} encoding of datatypes comes from the programming languages literature~\cite{inductive}.
% It is called inductive because it describes all ways
% one can construct instances of that datatype.
% For \lstinline{peano},
% there are exactly two ways of constructing them, corresponding
% to the two cases above: (1) In the \textbf{base case}, the number one (denoted \lstinline{I}) is a positive natural number. (2) In the \textbf{inductive case}, for any positive natural number,  we can get a new positive natural number by adding one (that is, taking its successor---denoted \lstinline{S n} for positive natural number \lstinline{n}).
% So, \lstinline{I} corresponds to one, \lstinline{S(I)} corresponds to two, \lstinline{S(S(I))} corresponds to three, and so on.

% Whenever we have a datatype defined like this, we can write
% recursive functions and proofs over it. Consider addition:\footnote{For those unfamiliar with structural pattern matching, one can view \lstinline{match} as a glorified \lstinline{if} statement that can split into cases based on substructures in a smart way. There are some helpful tutorials for Python online, for example: \url{https://peps.python.org/pep-0636/}.}
This describes the construction of instances of the datatype, with two cases: (1) the \textbf{base case} where one is a positive natural number denoted \lstinline{I}, and (2) the \textbf{inductive case} where adding one to any positive natural number \lstinline{n} gives a new positive natural number \lstinline{S n}

We can write recursive functions and proofs over this datatype. Consider addition using structural pattern matching and recursion, as in Figure~\ref{fig:sub:addition_pattern}.\footnote{For those unfamiliar with structural pattern matching, one can view \lstinline{match} as a glorified \lstinline{if} statement that can split into cases based on substructures in a smart way. There are some helpful tutorials for Python online, for example: \url{https://peps.python.org/pep-0636/}.}
In the base case, \lstinline{add 1 m}\footnote{Coq uses parentheses only for grouping, not for function calls; in Python this would be written \lstinline{add(1, m)}.} is its successor \lstinline{S m}. In the inductive case, we recurse: we compute \lstinline{add (S p)  m} by recursively computing \lstinline{add p m} and
then taking its successor. 
%This behaves like the addition function we
%know and love in real life.

% Note that this representation meets our first goal:
% (1) It starts from first principles, building datatypes from scratch
% by succinctly describing all ways of constructing
% them, and grounds their semantics by building functions on top of 
% them. Importantly, these semantics do not depend on the particular characters used to represent numbers, nor on any associations a model may have learned about numbers already.

The nice thing about this representation is that it constructs datatypes from scratch, by describing all ways of constructing those datatypes, and establishing their semantics by functions defined over them. Importantly, these semantics are independent of specific character representations and any preexisting associations (for example, to numbers) that a model may have learned; what we call these datatypes is irrelevant.
%---it just
%captures the essence of computation over recursive %structures.
% We could just as well rename \lstinline{peano}
% to \lstinline{dinosaur}, \lstinline{I} to
% \lstinline{shampoo}, and \lstinline{S} to 
% \lstinline{pencil}:

% \begin{lstlisting}
%   Inductive dinosaur :=
%   | shampoo : dinosaur (* base case *)
%   | pencil : $\forall$ (n : dinosaur), dinosaur. (* inductive case *)
% \end{lstlisting}
% and then we could \lstinline{add} dinosaurs.

% %which makes it interesting from the perspective of understanding model behavior.
Still, this simple representation corresponds to a broad class of datatypes that is well studied in programming languages, and that makes it simple for us to define important recursive tasks.

\section{Tasks}
\label{sec:overview}

\iffalse
{\color{blue} Feedback: ``One question I had reading this is, how do you define the distribution of examples you’ll sample from? You probably do this in some other section, but may want to consider including it here (since it’s common in ML to specify a task by specifying the distribution on samples).''}
{\color{red} Dylan's answer here: This is a good question. I think we will need to justify this. 
1. We are always training the model using the pairs from 1 to $n$ where $n$ is the maximum number for training. That is why we can test OOD performance. BUT, people may have questions like ``why don't you balance each case, i.e. data requiring different depths of recursion " ?  IDK how to justify that though :( 
}
{\color{purple} Dylan, can you please do a pass integrating that?}
\fi

We consider two tasks: the binary successor function (Section~\ref{sec:bin}) and
a tree traversal (Section~\ref{sec:tree}).
For each task, we choose (1) an inductive representation of a datatype (like \lstinline{peano}), (2) a recursive function over that datatype (like \lstinline{add}), and (3) variants of a learning task to approximate
that function. 
Since our interest is in whether the model can lean to emulate recursion, we train each model to approximate
the function's \emph{computation}, rather than to 
explicitly represent the function.

%Nonetheless, for many tasks, it will be important to also reify
%these functions explicitly as syntax (for example, by synthesizing programs); we leave this to future work.

\subsection{Binary Successor}
\label{sec:bin}

The binary successor function task is a simplification
of a common benchmark used for over two decades of symbolic 
proof automation~\cite{magaudbertot, ringer2021proof}.
It captures the essence of what it means to
adapt functions and proofs defined over the unary \lstinline{peano} natural numbers so that they instead are defined over binary natural numbers.
Its appeal is that it is simple and useful, yet still structurally interesting, in that its structure does not just
amount to counting.
%Moreover, it has broad implications for program and proof automation.

\paragraph{Inductive Representation}

A positive binary number can be defined inductively:

\begin{lstlisting}
  Inductive bin_pos :=
  | 01 : bin_pos (* base case *)       
  | XO : $\forall$ (b : bin_pos), bin_pos (* first inductive case: shift left *)
  | X1 : $\forall$ (b : bin_pos), bin_pos. (* second inductive case: shift right and increment *)
\end{lstlisting}
That is, a positive binary number is either (1) one (the base case, denoted \lstinline{01}), (2) any another positive binary number shifted to the left (the first inductive case, denoted \lstinline{XO b} for positive binary \lstinline{b}), or (3) any other positive binary number shifted to the left and then incremented by one (the second inductive case, denoted \lstinline{X1 b} for positive binary \lstinline{b}).
One can uniquely construct all positive binary numbers by making sequences of \lstinline{XO} and \lstinline{X1} calls on \lstinline{01}. For example, two can be written as \lstinline{XO 01}: shift \lstinline{01} to the left.
Three can be written as \lstinline{X1 01}: shift \lstinline{01} to the left and then increment.
%Four can be written as \lstinline{XO XO 1}: shift \lstinline{1} to the left twice.
And so on.

To recover the ``natural'' ordering we might write on paper, we
can just reverse the result and remove the \lstinline{X}s.
So, for example,
\lstinline{XO XO 01} becomes $100$ (four).
The ``reverse'' ordering makes it easy to define functions recursively, since it is structurally decreasing---functions on sequences can be defined in terms of functions on the subsequences that follow. We train separate models to learn from input/output examples defined using each of the ``natural'' and ``reverse'' orderings.

\paragraph{Recursive Computation}
%To write recursive functions over binary positive number written this way, we can simply pattern
%match over the first character, then recurse on the remaining characters when relevant.
%Borrowing from the symbolic proof automation literature,
We target the successor function: % which increments any positive binary number by \lstinline{1}:

\begin{minipage}{0.59\textwidth}
\centering
\begin{lstlisting}
  Fixpoint s n :=
    match n with (* break into cases *)
    | 01 => XO 01 (* if n is one, return two *)
    | XO b => X1 b (* if the LSB is zero, flip it *)
    | X1 b => XO (s b) (* otherwise, recurse and shift *)
    end.
\end{lstlisting}
\end{minipage}
\hfill
\begin{minipage}{0.28\textwidth}
        \centering
\includegraphics[width=\columnwidth]{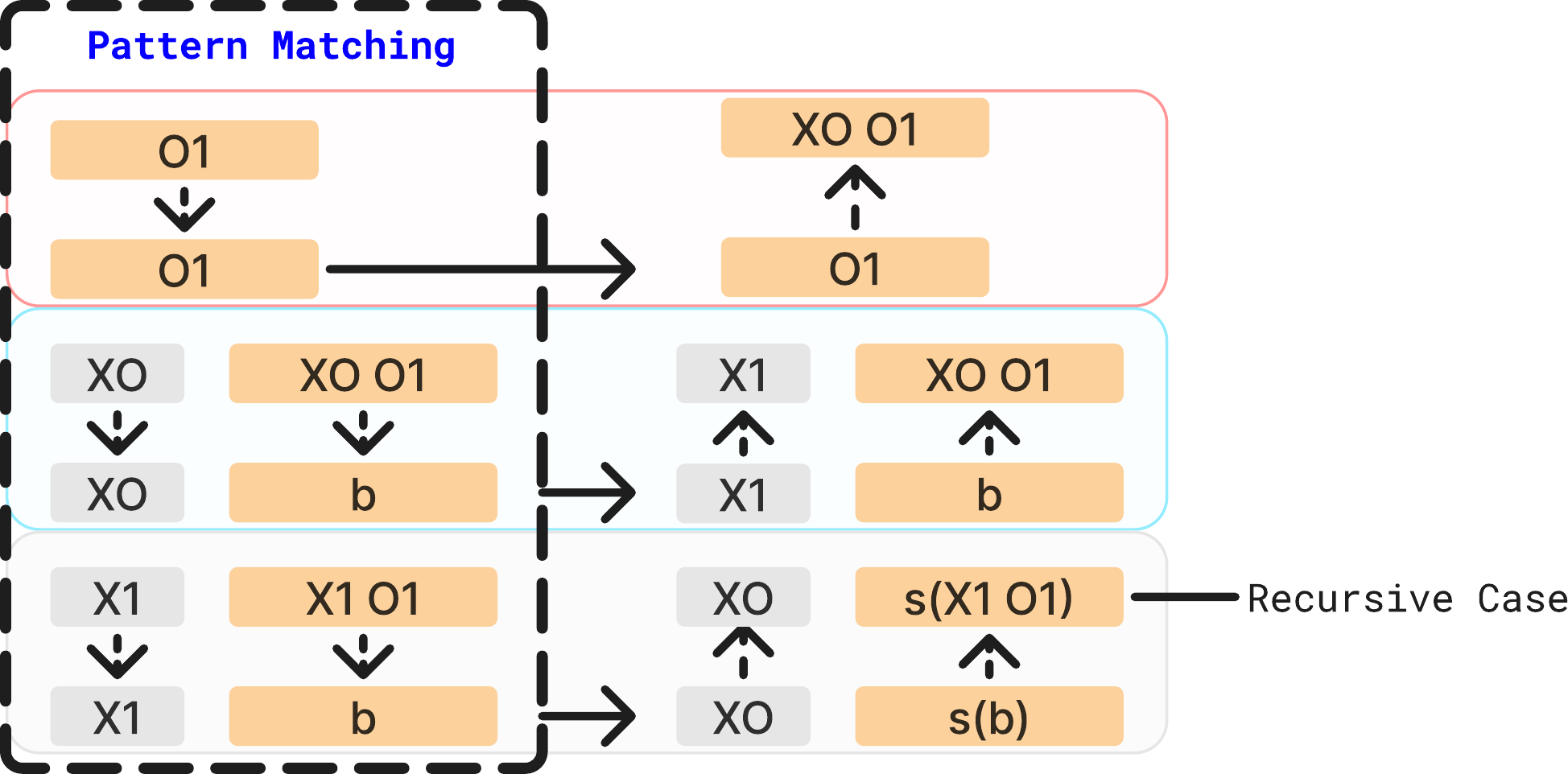}
    \label{fig:patternmatch}
\end{minipage}
\\
The function matches over the first character (the least significant bit or LSB) and increments. In the base case, it increments one (\lstinline{01}) to two (\lstinline{XO 01}). In the first inductive case, given some binary number \lstinline{b} shifted to the left (\lstinline{XO b}), it increments
by flipping the LSB from zero to one (\lstinline{X1 b}).
In the second inductive case, given some binary number \lstinline{b} shifted to the left and incremented by one (\lstinline{X1 b}), it increments by recursively computing the successor of \lstinline{b} (\lstinline{s b}), and then shifting the result to the left (\lstinline{XO (s b)}.
For example, the successor of eleven ($1011$, represented as \lstinline{X1 X1 XO 01}) is twelve ($1100$, represented as \lstinline{XO XO X1 01}), since:

\begin{lstlisting}
  s (X1 X1 XO 01) = XO (s (X1 XO 01)) = XO XO (s (XO 01) = XO XO X1 01
\end{lstlisting}
Details on the importance of this particular function can be found in Appendix~\ref{app:bin}.

\paragraph{Recursion by Example}

% https://twitter.com/niemasd/status/1650534767060987904?t=lrigLaT336AMYN32IDnKzw&s=19

We train our binary successor models on input-output pairs representing computation of the successor function. % with some
%variation in representation to measure impacts on both extensional and %intensional behavior.
For inspiration and understanding, it is worth thinking through how to learn the binary successor function from examples as a human. We can learn this function from a small number of input/output examples, without invoking knowledge of numbers:

\begin{minipage}{0.48\textwidth}
\begin{lstlisting}
  s 01 = XO 01
  s (XO 01) = X1 01
  s (X1 01) = XO XO 01
\end{lstlisting}
\end{minipage}
\hfill
\begin{minipage}{0.48\textwidth}
\begin{lstlisting}
  s (XO XO 01) = X1 XO 01
  s (X1 XO 01) = XO X1 01
  s (XO X1 01) = X1 X1 01
\end{lstlisting}
\end{minipage}\\
% These six examples fully represent the three cases, since we know that the only way to compose an instance of this datatype is by composing \lstinline{XO} and \lstinline{X1} with the base case \lstinline{01}, which can only ever appear last. Of course, both heuristics and prior knowledge come into play here, since (1) we understand the concepts of pattern matching and recursion, and (2) we tend to favor simple functions over very complex ones (so we do not expect that there will be unexpected, new behavior after, say, recursion of depth fifteen).
% It is from these priors that we can infer a template for the function, which matches over the input and breaks into cases; then, from our input-output examples, we can fill the ``holes'' in the template in a way that satisfies our cases and our priors. See Appendix~\ref{app:tree} for more details.
These examples fully represent the three cases. We compose instances using only three symbols: \lstinline{XO}, \lstinline{X1}, and the base case \lstinline{01}. Heuristics and prior knowledge guide us, considering pattern matching and recursion, and preferring simple functions. We can infer a template, match cases, and fill in the template based on input-output examples. More details can be found in Appendix~\ref{app:bin}.

This approach to learning \lstinline{s} is \emph{roughly} how symbolic automation for program synthesis works. This can get considerably more complicated---and require more examples---if we assume that it is possible for our synthesized function to call other functions in its computation, or to do unbounded recursion~\cite{lee2023popl}. But this is still the essence of synthesizing recursive functions symbolically---and it needs no knowledge of the fact that \lstinline{bin_pos} represents numbers, let alone binary positive numbers.

Our goal for this task is to see how much of this semantic information a transformer model can learn \emph{without} the priors we just described---plus how it represents the information it learns, and where the learned algorithms fall short.
%We look in particular at transformer models because of their ubiquity, and because of disagreement over their theoretical and practical abilities.

\subsection{Tree Traversal}
\label{sec:tree}

For a second and more challenging task, we consider tree traversals.
How transformer models approximate the behavior of
tree traversals is informative for many important
symbolic domains, since tree traversals are at the heart of the symbolic search procedures for programs, games, and proofs.
%Tree traversals are also at the heart of important
%algorithms in other domains like bioinformatics~{\color{purple} cite}.
If transformers can approximate tree traversals, this may mean better performance of neural tools for these procedures without the need for a symbolic search procedure to guide the way.

\paragraph{Inductive Representation}
We study the traversal of binary trees with character values. The code for this is in
Appendix~\ref{app:tree}; it includes an empty \lstinline{Leaf} base case,
and a \lstinline{Branch} inductive case that stores a character value, a left subtree, and a right subtree.
For example, we can represent a tree with \lstinline{'a'} at the root, \lstinline{'c'} in a node to its left, and \lstinline{'t'} in a node to its right as follows:

\begin{lstlisting}
  Branch 'a' (Branch 'c' Leaf Leaf) (Branch 't' Leaf Leaf)
\end{lstlisting}

% \paragraph{Recursive Computation}

% We consider preorder and inorder traversals.
% The code for an inorder traversal can be found in Appendix.%~\ref{app:tree}.
% Its base case returns the empty list \lstinline{[]}; its inductive case
% recurses on the left and right subtrees, and composes those with the new value in the appropriate order.
% For example, \lstinline{inorder} of the tree from above is
% \lstinline{['c', 'a', 't']}. 
% \iffalse
% (while \lstinline{preorder} is
% \lstinline{['a', 'c', 't']}).
% \fi

\paragraph{Recursion by Example}

We consider preorder and inorder traversals. Details are deferred to Appendix~\ref{app:tree}.
As with the previous example, we consider the problem of learning
these recursive functions from input-output examples. Since the tree
datastructure is not sequential in its recursive structure
(that is, each pass of recursion visits both the left and right subtrees), we also decompose these traversals into atomic computations
and see if those are easier to learn. 
The machine learning inspiration for these atomic computations comes
from chain of thought reasoning, while the exact atomic computations we
choose come from programming languages research. These atomic computations decompose recursion into one \textit{reduction} step at a time. For example, to compute:

\begin{lstlisting}
  inorder (Branch 'a' (Branch 'c' Leaf Leaf) (Branch 't' Leaf Leaf))
\end{lstlisting}
we can reduce one step by selecting the \lstinline{Branch} case and substituting in the right values:

\begin{lstlisting}
  (inorder (Branch 'c' Leaf Leaf)) ++ ['a'] ++ (inorder (Branch 't' Leaf Leaf))
\end{lstlisting}
%where \lstinline{++} appends two lists (like \lstinline{+} in Python).
Two more reductions (see Appendix~\ref{app:tree}) get us to
the final result, \lstinline{['c'; 'a'; 't']}.
Each reduction step here has a formal meaning in programming languages,
in terms of ``reduction rules'' named after various Greek letters. Computation amounts to composition of these
reduction rules until it is no longer possible to reduce further; the order of reduction does not matter.
Using this as inspiration, in addition to training models to learn the traversal all at once, we also train models to reduce the traversal just one, two, or three times at once, to see at what point performance degrades.

%% file: asm.tex
    \section{Computation Model: Abstract State Machines}
\label{sec:model}
% Now that we have introduced the two prototypical programming tasks whose solution involves the use of structural recursion, we need to settle on a model of computation that can encompass both the original recursive implementation and its approximate simulation by a learned transformer network. A natural impulse would be to situate the original recursive implementation at an appropriate level in the usual hierarchy of formal computational models and then investigate the ability of transformers to simulate the models at that level (see Related Work for a discussion of the extensive literature dedicated to this line of research). However, this approach does not capture the relevant aspects of the problem at the right level of abstraction. Taking the binary successor function as an example, one could first construct an iterative implementation of the recursion using a stack and then instantiate this iterative implementation using a Turing machine. The inevitable (and unfortunate) consequence is that, since the Turing machine implementation operates at a much lower level of abstraction, the elegance, parsimony, and interpretability of the original recursive definition are lost. In fact, one can appreciate this loss of interpretability by simply noting that, even though both the recursive and the Turing machine implementations of binary successor admit finite descriptions, it is much harder to infer program semantics from the latter than from the former. 

We now need a computation model that can encompass both the original \textit{recursive implementation} and its \textit{approximate simulation} by a learned transformer network. 
% Instead of situating the recursive implementation within the hierarchy of formal computational models and investigating the transformers' ability to simulate them, or using , we should focus on capturing the relevant aspects at the appropriate level of abstraction. 
Intuitively, transformers do not implement stacks to trace recursion, yet instead are sequence models by construction. On the other hand, computation models like Turing machines operate on low levels of abstraction, making them hard to interpret. Abstract State Machines (ASMs)~\cite{gurevich1995evolving}
were introduced with the explicit goal of capturing the semantics of programs at various levels of abstraction, and so provide us with a flexible yet powerful framework to analyze transformers.  

The states of ASMs are first-order structures of the form $(U,f_1,\dots,f_k)$, where $U$, called the algorithm's \textit{universe}, is a set that includes the constants {\tt{true, false, undef}} and where the collection of functions $f_i : U^{n_i} \to U$ includes Boolean functions {\tt{not}}$ : U \to \{{\tt true,false,undef}\}$ and {\tt{and}}$ : U^2 \to \{{\tt true,false,undef}\}$.
%(All Boolean functions return \lstinline{undef} when at least one of their arguments is not an element of $\{{\tt true, false\}}$.)
The Boolean constants and functions are needed to implement conditionals and other flow control structures and guards, such as {\tt{if ... then ... else}}. The set $U$ can be infinite and has minimal restrictions on the datatypes it can take, e.g., real matrices or tensors, streams, and so on.
%- real numbers, matrices, vectors, lists, streams, etc. 
Likewise, the functions $f_i$ can include operations like matrix multiplication, list or stream operations, and so on. Closer to our purposes, $U$ may include token- and positional-embeddings, while the functions $f_i$ may include the MLPs, self- and cross-attention matrices, and all other building blocks of transformer networks \cite{phuong2022formal}. Each decoding step of an encoder-decoder transformer is naturally a sequential ASM update. 

Notably, the two key features of ASMs are a finite \textit{instruction set} and a possibly infinite state. Starting from this perspective, we can analyze the algorithms implemented by learned transformers by attempting to search for its \textbf{pattern classifiers} and the {\tt{if-else}} structure following each pattern, as well as the functions applied in each case at each time step. A simple example of such procedure would be ``{\tt{if}} \{the sequence to be generated is complete\} {\tt{then}} generate \textbf{[EOS]} token.''

The challenge arises when using learned transformer networks to approximate or simulate recursive algorithms. First, the ASM it simulates lacks recursive structure by nature. Also, training samples %, in the form of input-output pairs or program traces, 
provide a partial \textit{extensional} description of the unknown recursive function, while the pattern classifier determining recursive calls is an \textit{intensional} description not encoded in the training objective or data. To this end, understanding whether the class of \textit{sequential} ASMs implemented by transformer networks can effectively approximate \textit{recursive} ASMs comes down to \textbf{reconstructing} the program by identifying the conditionals and functions it implements, \textbf{examining} the correctness of its approximated program, and \textbf{evaluating} its capability of correctly executing that program. We discuss experimentation toward this end in Appendix~\ref{app:recursive}.

% In essence, the question at hand is whether the class of \textit{sequential} ASMs implemented by transformer networks can effectively approximate recursive ASMs.

%% file: evaluation.tex
\section{Empirical Analysis}
\label{sec:evaluation}

We trained transformer models to tackle the binary successor (Section~\ref{sec:evalbinsuc}) and tree traversal (Section~\ref{sec:evaltree})
tasks from Section~\ref{sec:overview}. We focused on encoder-decoder models since
encoder-only and decoder-only architectures performed worse under our set-up (Appendix~\ref{app:extrap}). We framed both tasks as
sequence-to-sequence problems. We summarize the results here; we defer training details to Appendix~\ref{app:training}.

\subsection{Binary Successor Task}
\label{sec:evalbinsuc}

For the binary successor task, we found the following:

\begin{enumerate}
    \item \textbf{The model's attention maps exhibit clear recursion-capturing patterns} (Section~\ref{sec:recursionheads}). These attention maps unveil what we call \textit{recursion heads}.
    \item \textbf{A perturbation analysis provides a granular explanation of the algorithm} (Section~\ref{sec:perturb}). We are able to \textit{reverse engineer} the learned algorithm by perturbing tokens on the fly.
    \item \textbf{A majority of failures are foreseeable from the reconstructed algorithm} (Section~\ref{sec:failures}). We can predict \textit{91\% of failure cases} from the reverse engineered algorithm.
    \item \textbf{Learning rates impact learned algorithms and generalization abilities} (Section~\ref{sec:lr}). The model appears to learn \textit{different algorithms} for different learning rates.
\end{enumerate}
A detailed reconstruction of the learned algorithms we reverse engineered for this task is in Appendix~\ref{app:reconstruct}.

\subsubsection{The Model's Attention Maps Exhibit Clear Recursion-Capturing Patterns}
\label{sec:recursionheads}

Our first step to understanding the algorithm implemented by the model was to visualize the attention maps of the model. For an encoder-decoder transformer, three types of attention can be analyzed: decoder self-attention, encoder-decoder cross attention, and encoder self-attention.
For this task, we found that cross-attention was not interesting, but decoder and encoder self-attentions exhibited visibly interesting behaviors.

\begin{figure}
\centering
     \begin{subfigure}[t]{.2\columnwidth}
         \centering
         \includegraphics[width=\columnwidth]{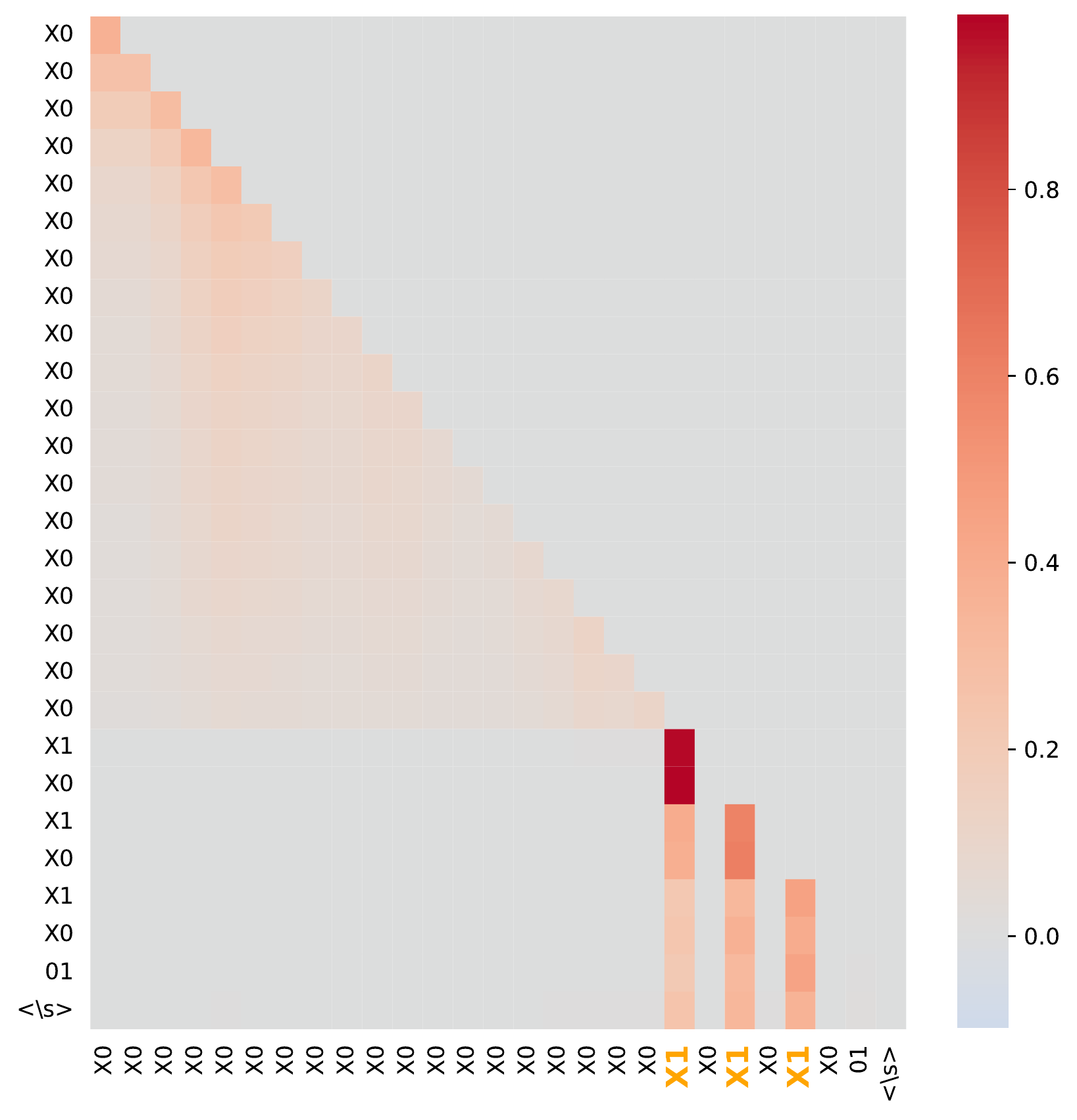}
         \caption{Decoder self-attention. Reverse Order.}
         \label{fig:rev_dec}
     \end{subfigure}
     \hspace{5mm}
     \begin{subfigure}[t]{.2\columnwidth}
         \centering
         \includegraphics[width=\columnwidth]{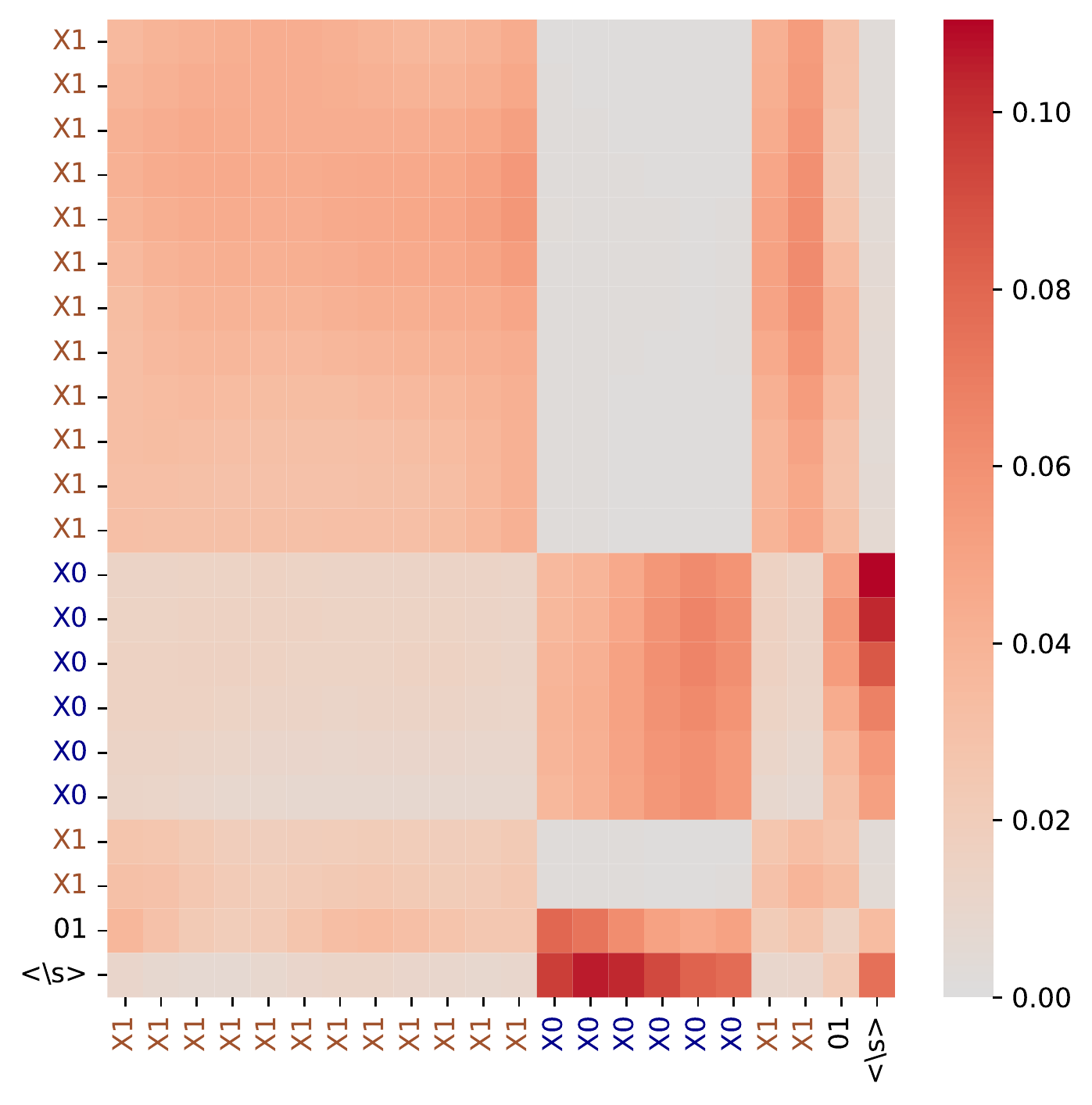}
         \caption{Encoder self-attention. Reverse Order}
         \label{fig:enc_rev}
     \end{subfigure}
     \hspace{5mm}
     \begin{subfigure}[t]{.2\columnwidth}
         \centering
         \includegraphics[width=\columnwidth]{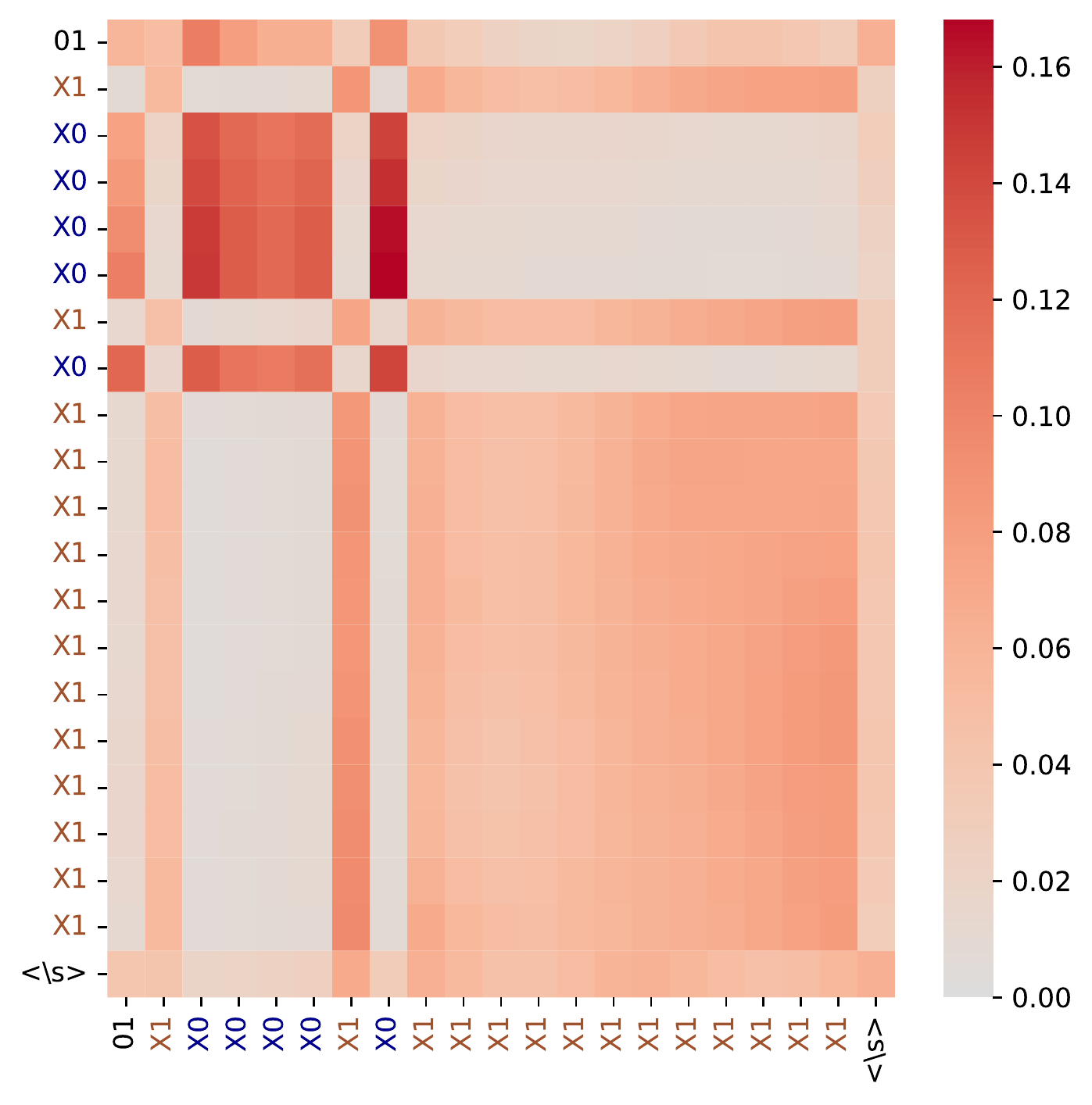}
         \caption{Encoder self-attention. Natural order}
         \label{fig:enc_nat}
     \end{subfigure}
        % \vspace{-3mm}
        \caption{Attention Maps.}
\end{figure}

Our visualization of decoder self-attention revealed a noteworthy phenomenon in the final layer---something we call a \textit{recursion head}
that specializes to recursive cases.
This was present in both the natural and reverse orders, though it served different purposes in each order, since each order implemented a different algorithm:
\begin{itemize}
\item \textbf{In the natural order} (Figure~\ref{fig:dec_nat_1}), the model commences attending to the last bit prior to flipping from \lstinline{X1} to \lstinline{XO}, and continues to do so until the end of the sequence. Thereafter, the recursion head predominantly allocates its attention towards the token we have described.
\item \textbf{In the reverse order} (Figure~\ref{fig:rev_dec}), the recursion head directs its attention towards the \lstinline{X1} tokens that have been generated earlier. This attention mechanism distinguishes between recursive segments, which necessitate modifications, and non-recursive segments, which do not require any rewrites. The first occurrence of an \lstinline{X1} token encountered by the recursion head serves as a boundary that separates these distinct segments.
\end{itemize}

% \begin{itemize}
% \item In the natural order (Figure~\ref{fig:dec_nat_1}), the model commences attending to the last bit prior to flipping from \lstinline{X1} to \lstinline{XO}, and continues to do so until the end of the sequence. Thereafter, the recursion head predominantly allocates its attention towards the token we have described.
% \item In the reverse order (Figure~\ref{fig:rev_dec}), the recursion head directs its attention towards the \lstinline{X1} tokens that have been generated earlier. This attention mechanism distinguishes between recursive segments, which necessitate modifications, and non-recursive segments, which do not require any rewrites. The first occurrence of an \lstinline{X1} token encountered by the recursion head serves as a boundary that separates these distinct segments.
% \end{itemize}

The encoder self-attention maps suggest that encoders also play a part in modeling semantics by helping differentiate between cases (Figure~\ref{fig:enc_rev}, and Figure~\ref{fig:enc_nat}). %and~\ref{fig:enc_nat}).
%in the process of rewriting the binary sequence into its successor.
%In the absence of semantic information provided to the tokens, the foremost task for the model is to ascertain their semantics. 
In particular, the model employs its low-level attention to identify and differentiate symbols by attending to tokens of the same kind. 

\subsubsection{A Perturbation Analysis Provides a Granular Explanation of the Algorithm}
\label{sec:perturb}
\begin{figure}[tb]
    \centering
    % \hspace{-5mm}
    \begin{minipage}{.57\columnwidth}
     \begin{subfigure}[tb]{\linewidth}
         % \centering
         \includegraphics[width=\columnwidth]{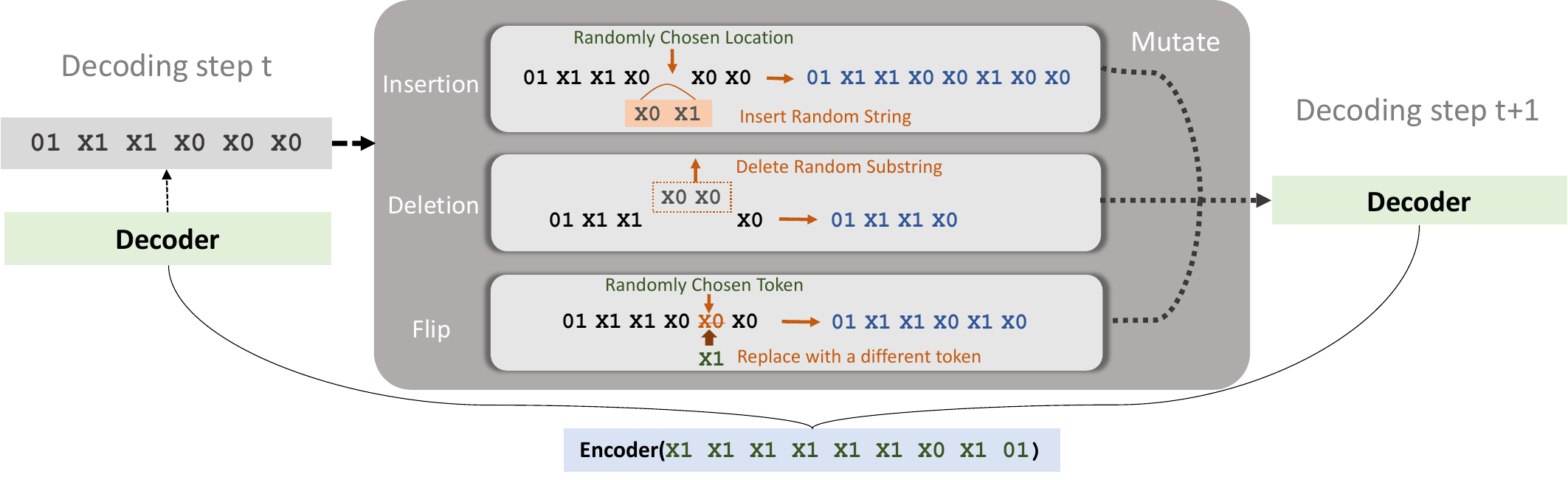}
         \caption{String manipulation workflow for natural order.}
         \label{fig:flip_nat}
     \end{subfigure}

     \begin{subfigure}[tb]{\linewidth}
         % \centering
         \includegraphics[width=\columnwidth]{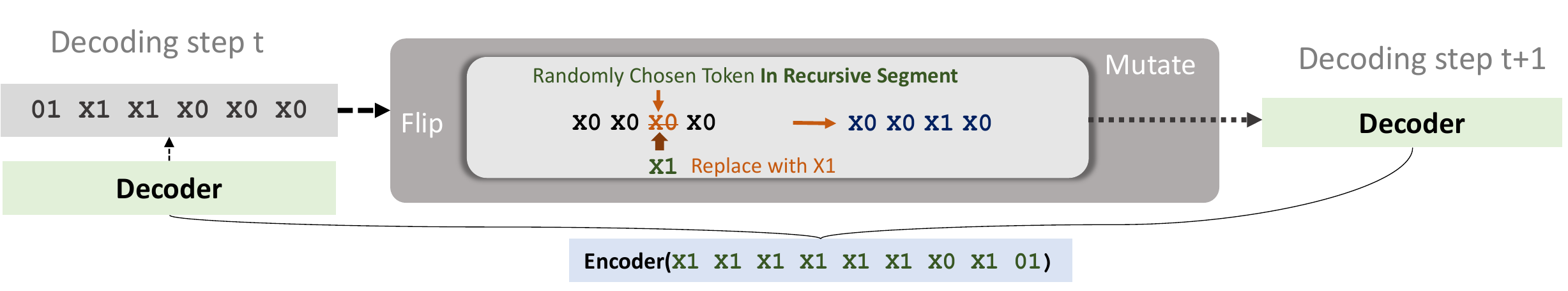}
         \caption{String manipulation workflow for reversed order.}
         \label{fig:flip_rev}
     \end{subfigure}
     \end{minipage}
     \hfill
     \begin{minipage}{.2\columnwidth}
     \begin{subfigure}[tb]{\columnwidth}
         \centering
        \includegraphics[width=\columnwidth]{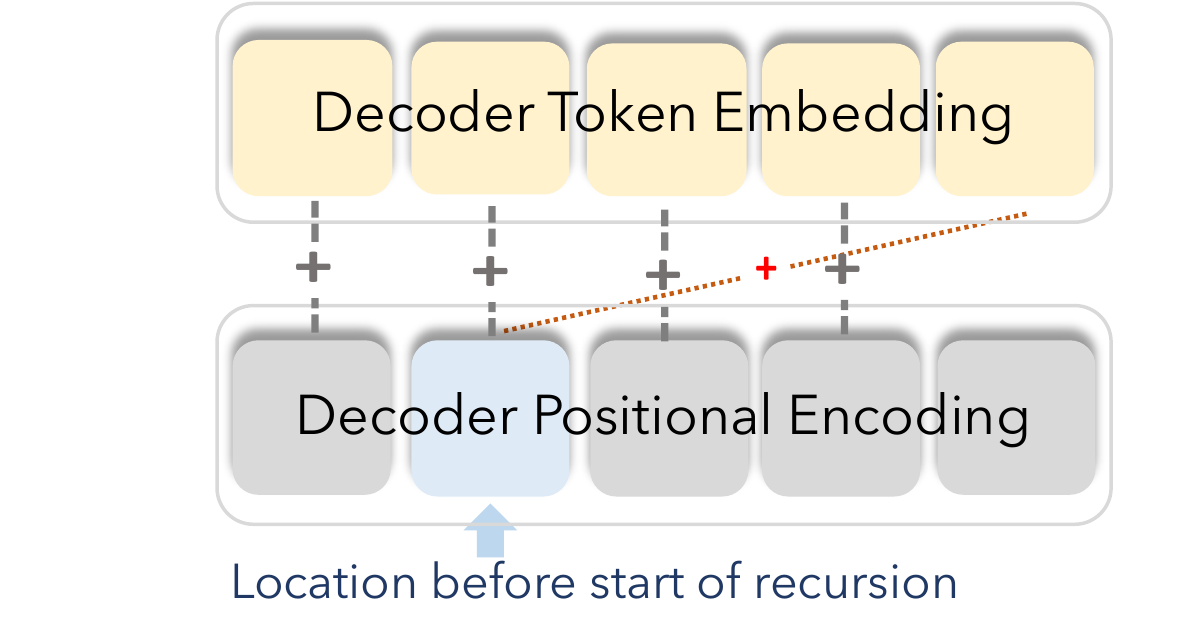}
         \caption{\textbf{SOR} Perturb.}
         \label{fig:posemb}
     \end{subfigure}
     \hfill
         \begin{subfigure}[tb]{\columnwidth}
         \centering
         \includegraphics[width=\columnwidth]{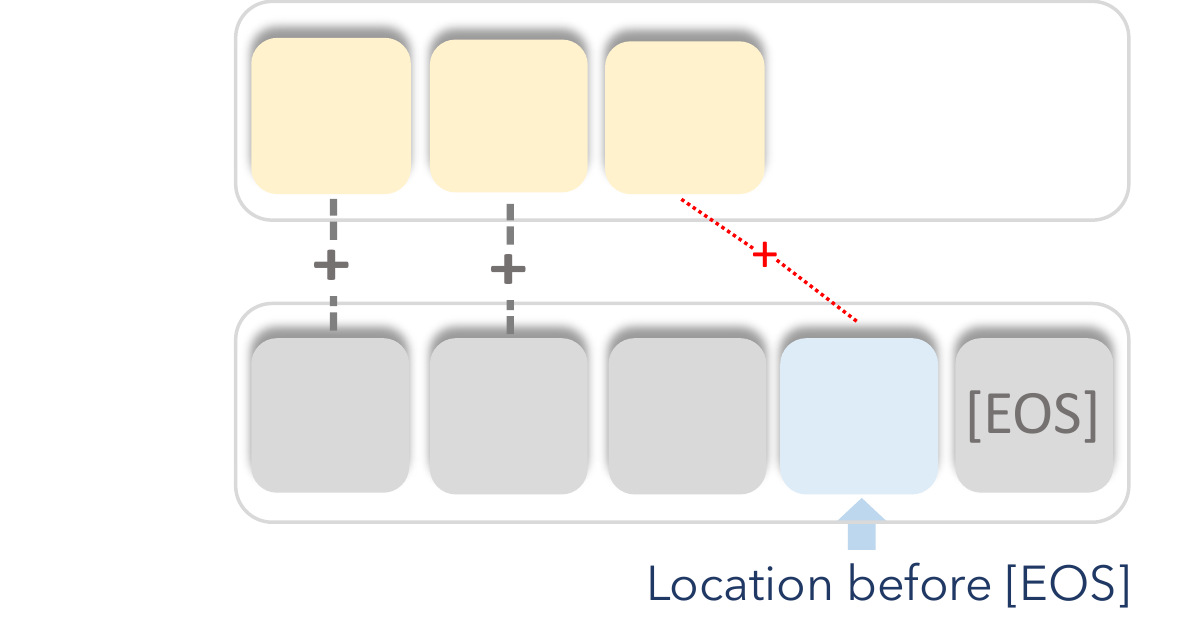}
         \caption{\textbf{EOS} Perturb.}
         \label{fig:posemb_eos}
     \end{subfigure}

 \end{minipage}
 \hfill
 \begin{minipage}{.2\textwidth}
     \begin{subfigure}[t]{\columnwidth}
     \centering
\includegraphics[width=\columnwidth]{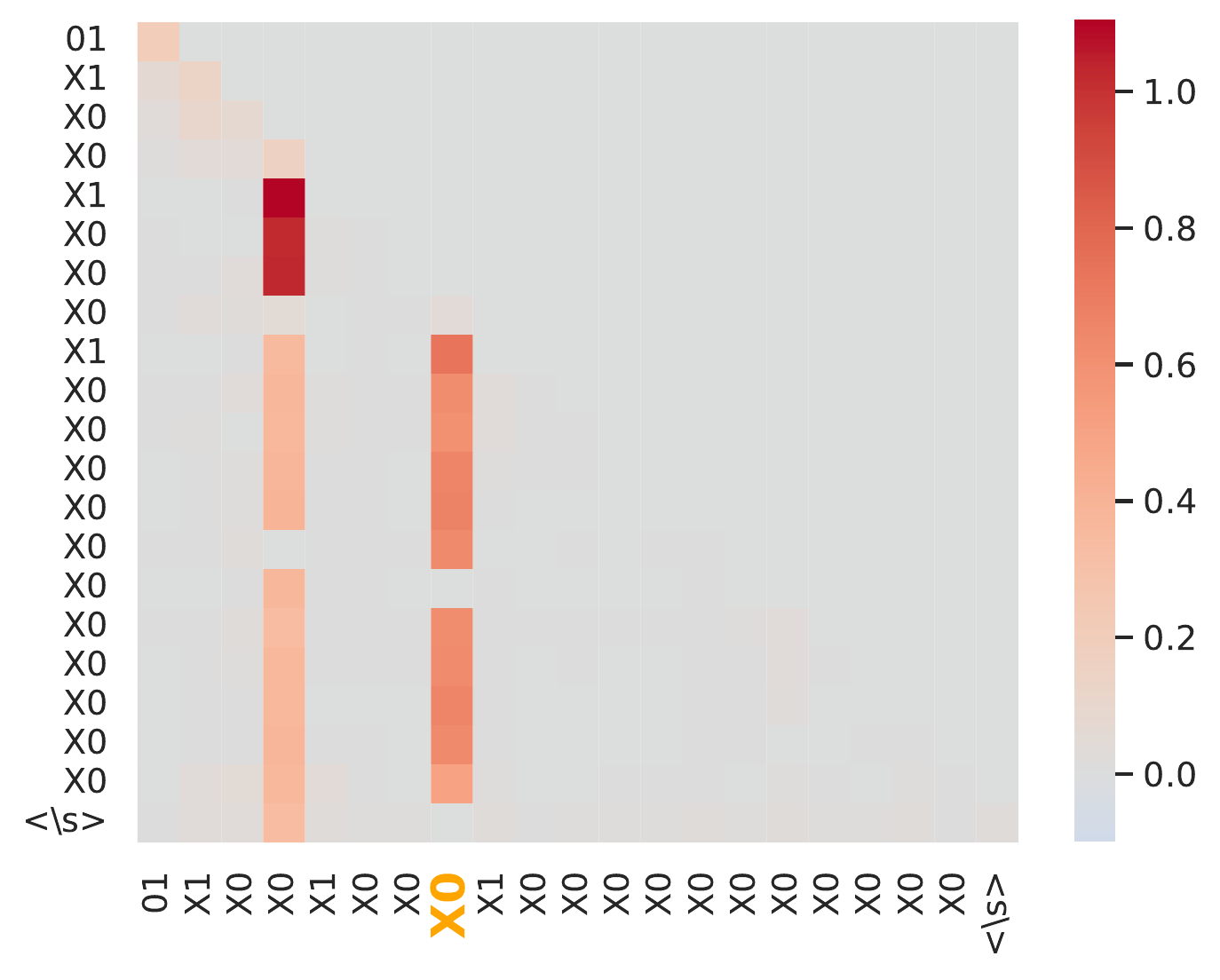}
     \caption{Nat. \textbf{SOR}.}
     \label{fig:perturb_nat_perturbed}
     \end{subfigure}
     % \hfill
     \begin{subfigure}[t]{\columnwidth}
     \centering
\includegraphics[width=\columnwidth]{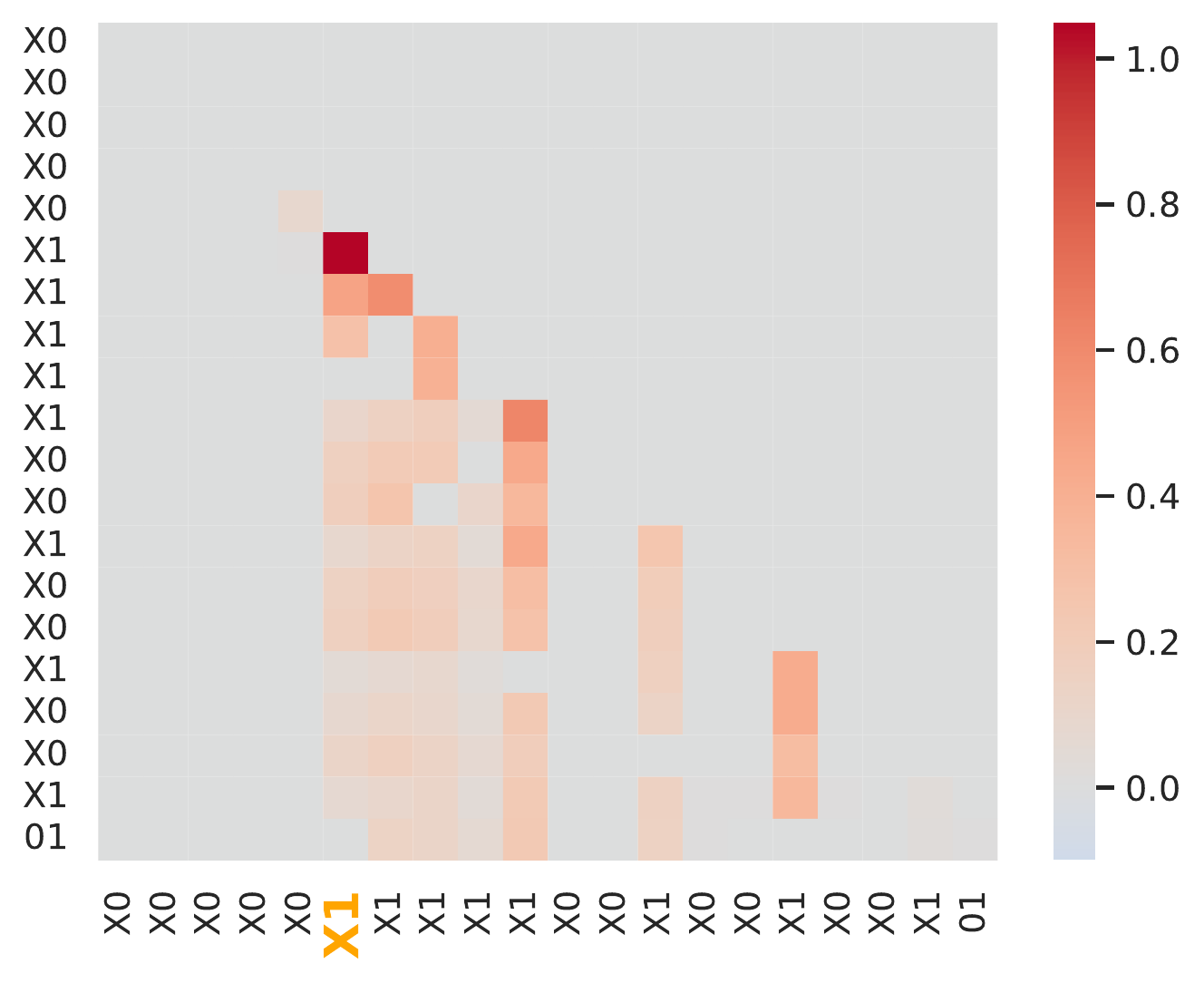}
     \caption{Rev. Flip Token.}
     \label{fig:perturb_rev_perturbed}
    \end{subfigure}
 \hfill
\end{minipage}
\hfill
    \caption{Perturbation analysis for the binary successor task. Figures~\ref{fig:posemb} and~\ref{fig:posemb_eos} are positional-encoding perturbations. \textbf{SOR} stands for `start-of-recusion' and \textbf{EOS} stands for `end-of-sentence'. Figure~\ref{fig:perturb_nat_perturbed} is the attention map of the recursion head on the natural order (Nat.) after adding the positional encoding of the token before the start of recursion to the bit after recursion starts. Figure~\ref{fig:perturb_rev_perturbed} is the attention map of the recursion head on the reversed order (Rev.) after randomly flipping a token in the recursive segment to \lstinline{X1}.}
    \label{fig:perturb_nat}
\end{figure}

Attention maps are useful for forming hypotheses
about model behavior, but they reveal only correlations, without any causal information.
To gain causal insights into the model's behavior, we conducted perturbation analyses---mutating tokens (i.e. randomly inserting, removing or flipping, as illustrated in Figures~\ref{fig:flip_nat} and ~\ref{fig:flip_rev}) and swapping positional encodings (see Figures ~\ref{fig:posemb} and ~\ref{fig:posemb_eos}) on the fly on the decoder side to see how this impacted the model's behavior. 

From this analysis, we were able to reconstruct the algorithm the model learns for the natural order:
(1) It computes the total number of bits required based on the input, and identifies the position at which the concluding bit of the subsequence eligible for direct copying from the input sequence is located. (2) It copies this particular segment, followed by a single \lstinline{X1}, 
followed by \lstinline{XO} tokens until it reaches the designated halting position.

%Interestingly, for natural order,
We determined this algorithm by perturbing both token content and positional encodings.
Interestingly, we found that the decoder exhibits a stronger reliance on positional information rather than the content associated with each position.
When we corrupted partial output using the token mutation process (Figure~\ref{fig:flip_nat}), the model could still recover the remaining sequence.
But when we changed the positional encoding
of the bit before the recursive segment to a random location (Figure~\ref{fig:posemb}), the model started ``recursing'' at the next time step by generating an \lstinline{X1} followed by \lstinline{XO}s. Furthermore, if we replaced the positional encoding just before \textbf{[EOS]} with a non-terminal token, the model immediately stops generation by producing \textbf{[EOS]}. 
% In fact, we found out that the decoder cares more about positional information instead of the actual content of that position when generating the output sequence. First, the model calculates the number of bits it needs to generate and figures out the sequence ending position. By randomly replacing  the positional embedding of any position in of the last bit of the sequence to any bit in the sequence will cause the model to halt generation immediately. Besides that, by replacing the positional embedding of the bit before recursion happens, the model always generates an \lstinline{X1} after this position and continues with \lstinline{XO}. 

In the reverse order, the model behaves differently. For the most part, it behaves as follows: (1) Based on the input sequence, the it determines the appropriate position for generating the first \lstinline{X1} token. (2) The decoder, while generating subsequent tokens, simultaneously examines the tokens it has previously generated to determine if an \lstinline{X1} token has already been produced. The presence of an \lstinline{X1} token serves as a signal for the model to switch from generating \lstinline{XO} tokens to copying the remaining portion of the sequence.
% This two-step process reveals a distinct mechanism employed by the model for handling reversed order sequences, further highlighting the complexity and adaptive nature of the algorithm developed by the model.

We determined this by systematically replacing each \lstinline{XO} token within the recursive segment (excluding the last token) with an \lstinline{X1} token. The purpose was to observe whether the model would indeed initiate the process of copying the remaining tokens. Intriguingly, our results indicate that in approximately 93.15\% of the cases, the model successfully copied the remaining tokens with complete accuracy. However, in the remaining cases, the model initially began generating \lstinline{X1} tokens, but exhibited confusion after a few tokens, deviating from the expected behavior. These findings provide empirical support for our hypothesis and echo the model behavior reflected from the attention maps.

% The attention map  By replacing an \lstinline{XO} in the recursive segement by \lstinline{X1}, the model will start to generate 

% For reversed order of
% \begin{figure}
%     \centering
%      \begin{subfigure}[tb]{.35\columnwidth}
%          \centering
%          \includegraphics[width=\columnwidth]{recursive_func_transformer_neurips/plots/perturbation/dec_self_head0_original.pdf}
%          \caption{Original attention map.}
%          \label{fig:perturb_nat_orig}
%      \end{subfigure}
%      % \hspace{30mm}
%      \begin{subfigure}[tb]{.35\columnwidth}
%          \centering
%          \includegraphics[width=\columnwidth]{recursive_func_transformer_neurips/plots/perturbation/dec_self_head0_perturbed.pdf}
%          \caption{Attention map after replacement.}
%          \label{fig:perturb_nat_perturbed}
%      \end{subfigure}
%     \caption{Decoder self-attention before and after replacing the positional encoding of a random position after recursion starts with the position of the last bit before recursion.}
%     \label{fig:replace}
% \end{figure}
% Both large LR and small LR models will 

% Interestingly, our current finding is that the model seem to be able to do well in the recursive cases but fails zero-recursion case. The suspicion is that the model was unable to copy a sequence of length beyond its training length. {\color{red}TODO: To verify this assumption, I will train models with 'zero-depth only', and examine the works on 'copying' capabilities of transformers.}

\begin{figure}[t]
\centering
    \begin{subfigure}[t]{.35\columnwidth}
     \centering
\includegraphics[width=\columnwidth]{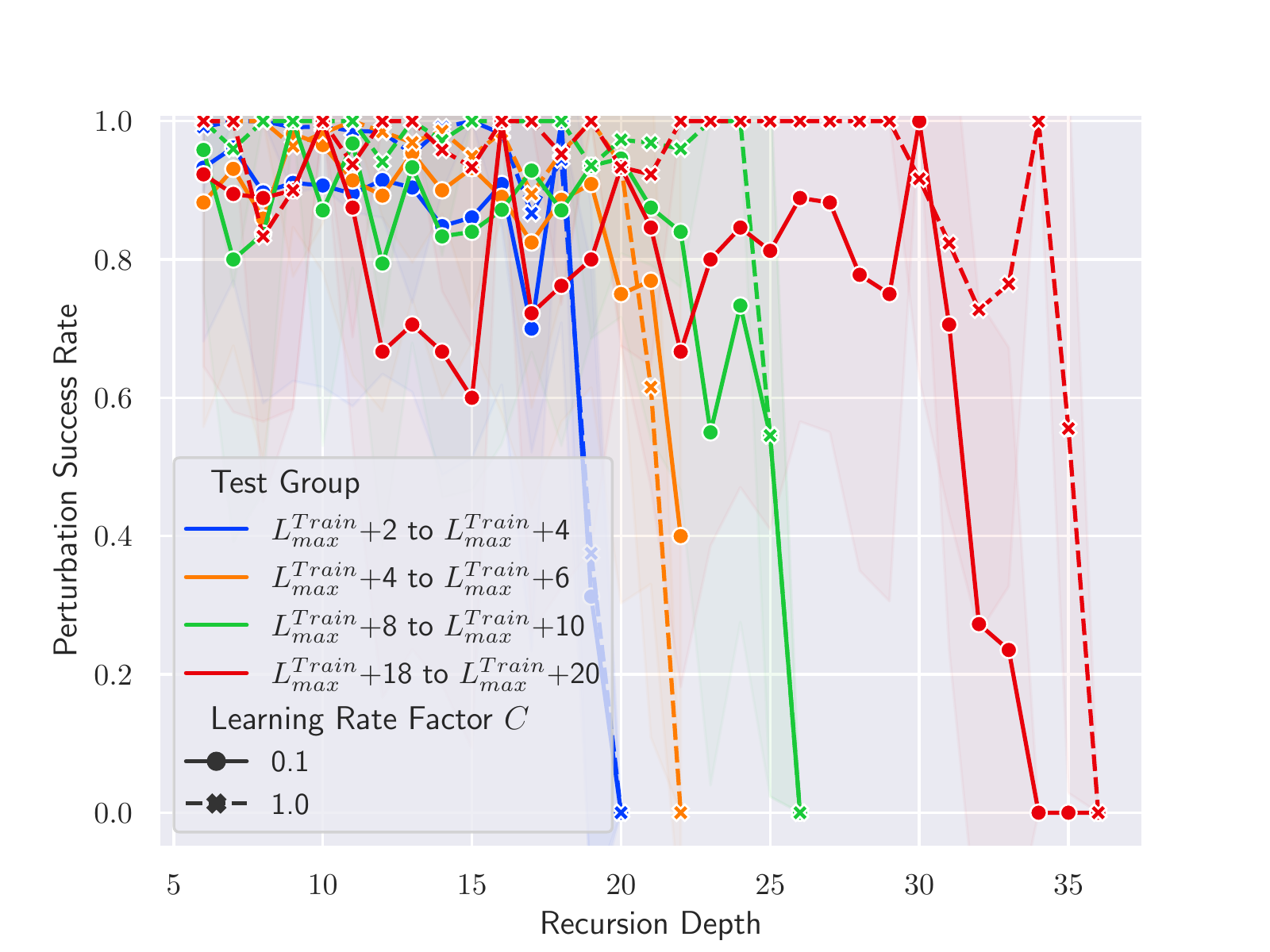}
     \caption{\textbf{SOR} Perturbation.}
     \label{fig:perf_perturb_sor}
 \end{subfigure}
     \begin{subfigure}[t]{.35\columnwidth}
     \centering
\includegraphics[width=\columnwidth]{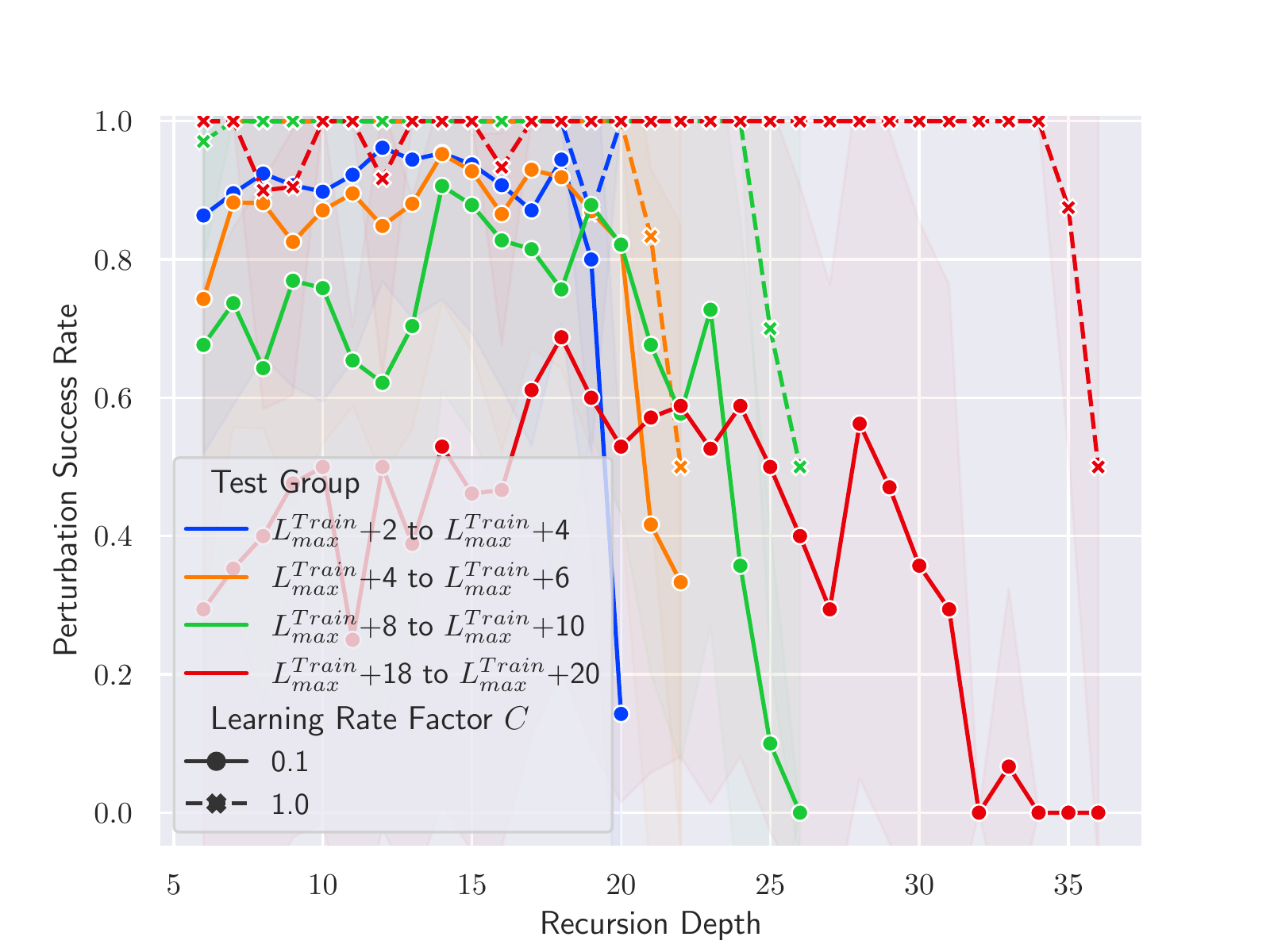}
     \caption{\textbf{EOS} Perturbation.}
     \label{fig:perf_perturb_eos}
     \end{subfigure}
     \caption{Results of positional embedding swap for natural order models. Perturbation success rate is the percentage of cases following the behavior described in Section~\ref{sec:perturb}}
\end{figure}
\subsubsection{A Majority of Failures are Foreseeable from the Reconstructed Algorithm}
\label{sec:failures}

The models fail in interesting ways. As shown in Figures~\ref{fig:nat_perf_depth} through~\ref{fig:nat_perf_depth6} in Appendix~\ref{app:depth}, the model is prone to failing in the maximum possible recursion depths for each test group on both directions. Interestingly, our perturbation analysis lets us \emph{correctly predict} that the model will fail on these cases---and gives us an understanding of \emph{why} that is true, too.

The specific failure cases are constructed by applying consecutive \lstinline{X1} operators immediately after the \lstinline{01} case.
The algorithm that the model learns in the natural order falls short for these cases: it 
identifies the location before recursion starts and generates an \lstinline{X1} followed by \lstinline{XO}s, when the correct answer
should be applying \lstinline{XO}s immediately after \lstinline{01}. In these cases, the shortcut is not applicable. 

In line with our understanding of the learned algorithm, the model fails on these cases $\mathbf{100\%}$ of the time for the natural order task! Among all failure cases (for $C=1$), ~$\mathbf{91\%}$ are due to one less \lstinline{XO} token generated, which is a consequence of the flaw of the model's learned algorithm. From observation, we saw that the model indeed attempts to play the same trick by finding the position right before recursion starts.
However, that position is no longer within the actual sequence, but rather in the ``pre-padding'' location. It encounters confusion between generating an \lstinline{X1} or a \lstinline{01} to start. It settles on \lstinline{01}, but this leads it to prematurely terminate, generating a sequence that is one token too short.

% For natural order, we discovered that the large and small LR models fail in distinct ways. 
\subsubsection{Learning Rates Impact Learned Algorithms and Generalization Abilities}
\label{sec:lr}

\begin{figure*}[!tb]
     \centering
     \begin{subfigure}[t]{.32\columnwidth}
         \centering
         \includegraphics[width=\columnwidth]{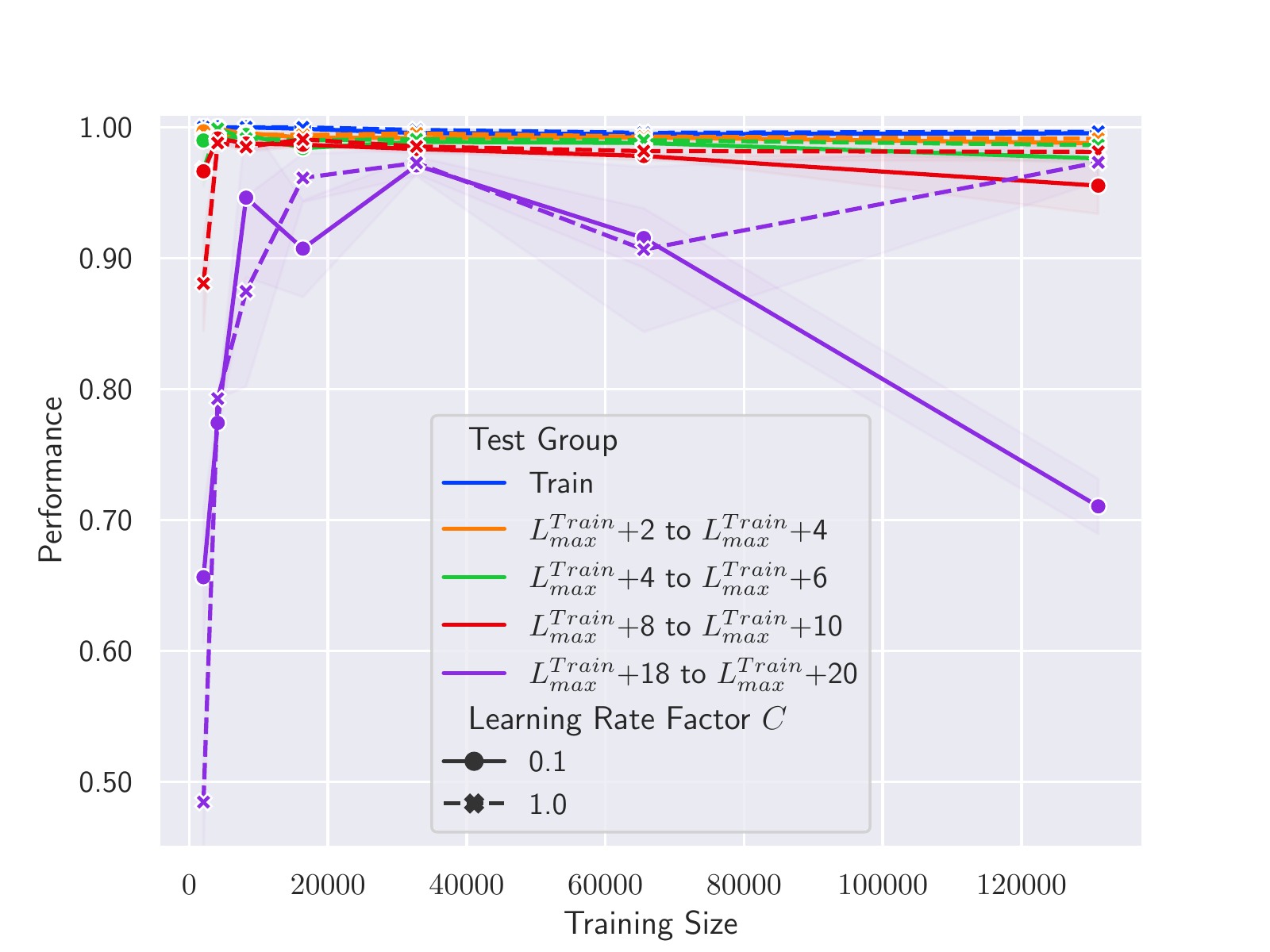}
         \caption{Natural Order}
         \label{fig:perf_nat}
     \end{subfigure}
    %  \hspace{6mm}
     \begin{subfigure}[t]{.32\columnwidth}
         \centering
         \includegraphics[width=\columnwidth]{recursive_func_transformer_neurips/plots/depth/depth3-perf_num_example_by_depth_bin_basic.pdf}
         \caption{Natural Order-Constrained.}
         \label{fig:perf_nat_depth3}
     \end{subfigure}
     \begin{subfigure}[t]{.32\columnwidth}
         \centering
         \includegraphics[width=\columnwidth]{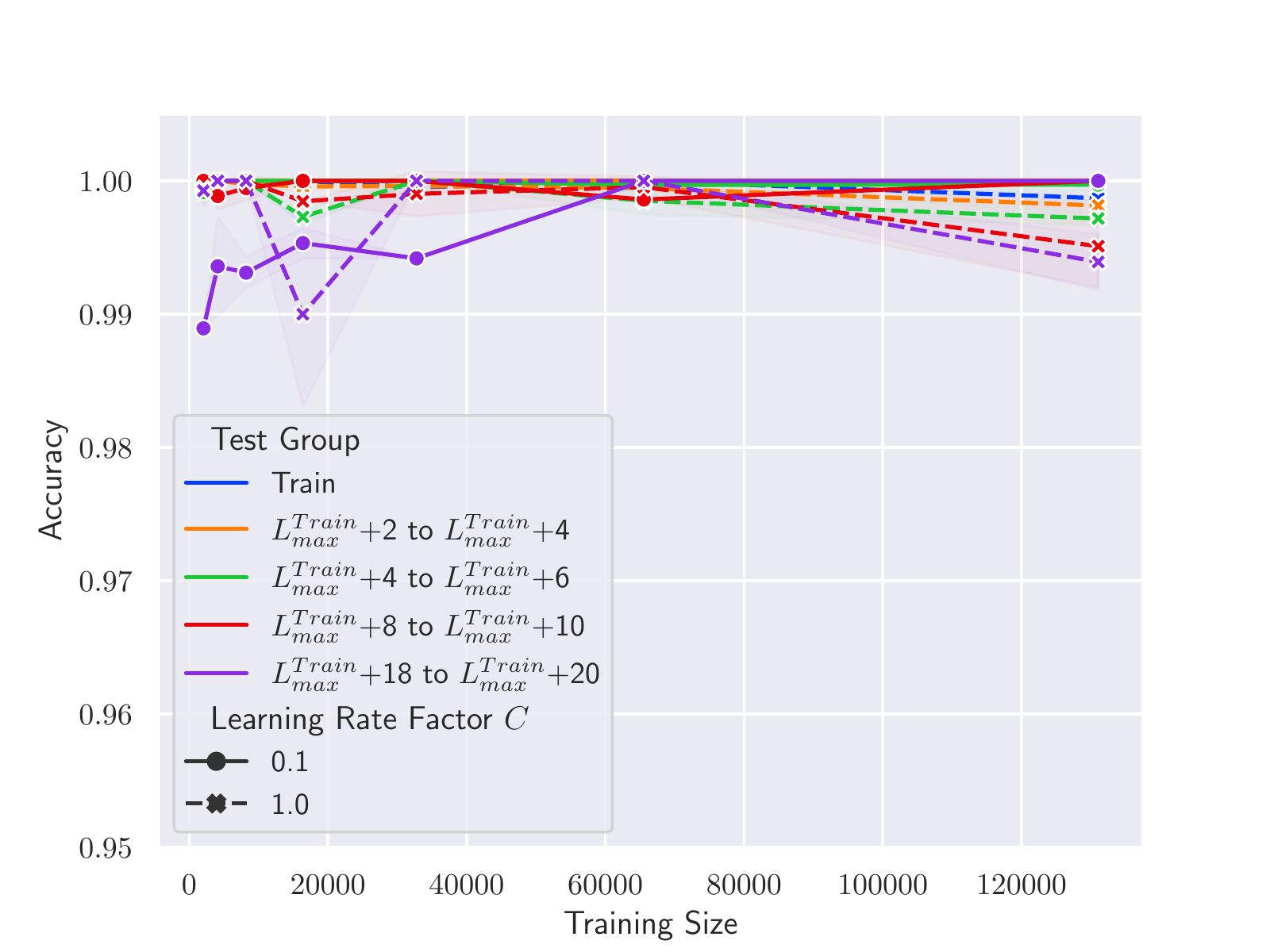}
         \caption{Reverse Order}
         \label{fig:perf_rev}
     \end{subfigure}

        % \vspace{-3mm}
    \caption{Performance versus number of training examples. The performance is the average of samples with all possible recursion depths with in that length range. The error bar indicates the standard deviation across total of 3 runs with different random seeds. Figure~\ref{fig:perf_nat_depth3} is trained with recursion depth up to 3.} 
\end{figure*}
\begin{figure*}[!tb]
     \centering
     \begin{subfigure}[b]{.15\columnwidth}
         \centering
         \includegraphics[width=\columnwidth]{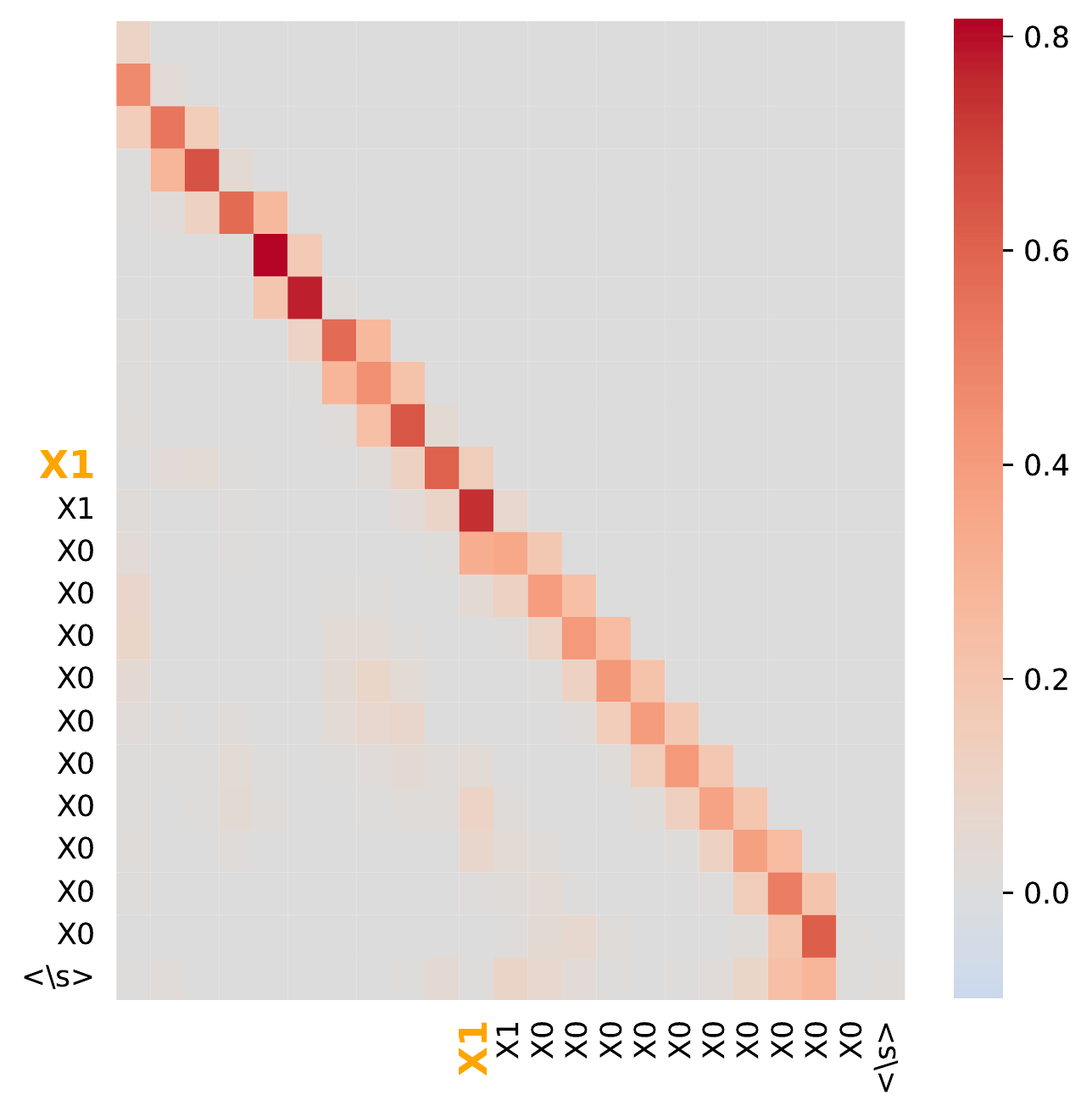}
         \caption{$C=0.1$}
         \label{fig:dec_nat_01}
     \end{subfigure}
         \begin{subfigure}[b]{.15\columnwidth}
         \centering
         \includegraphics[width=\columnwidth]{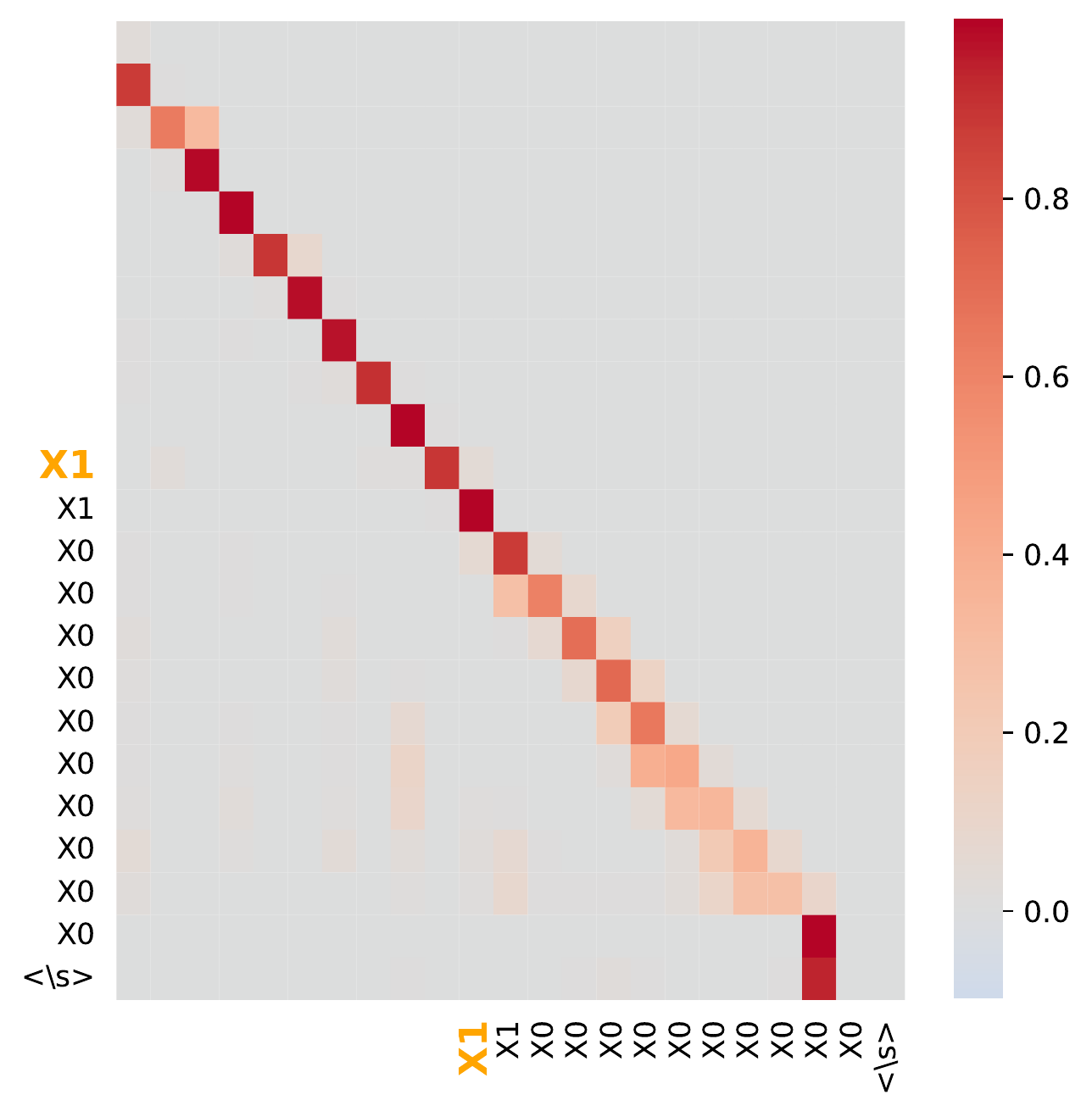}
         \caption{$C=0.3$}
         \label{fig:dec_nat_03}
     \end{subfigure}
    %  \hspace{6mm}
     \begin{subfigure}[b]{.15\columnwidth}
         \centering
         \includegraphics[width=\columnwidth]{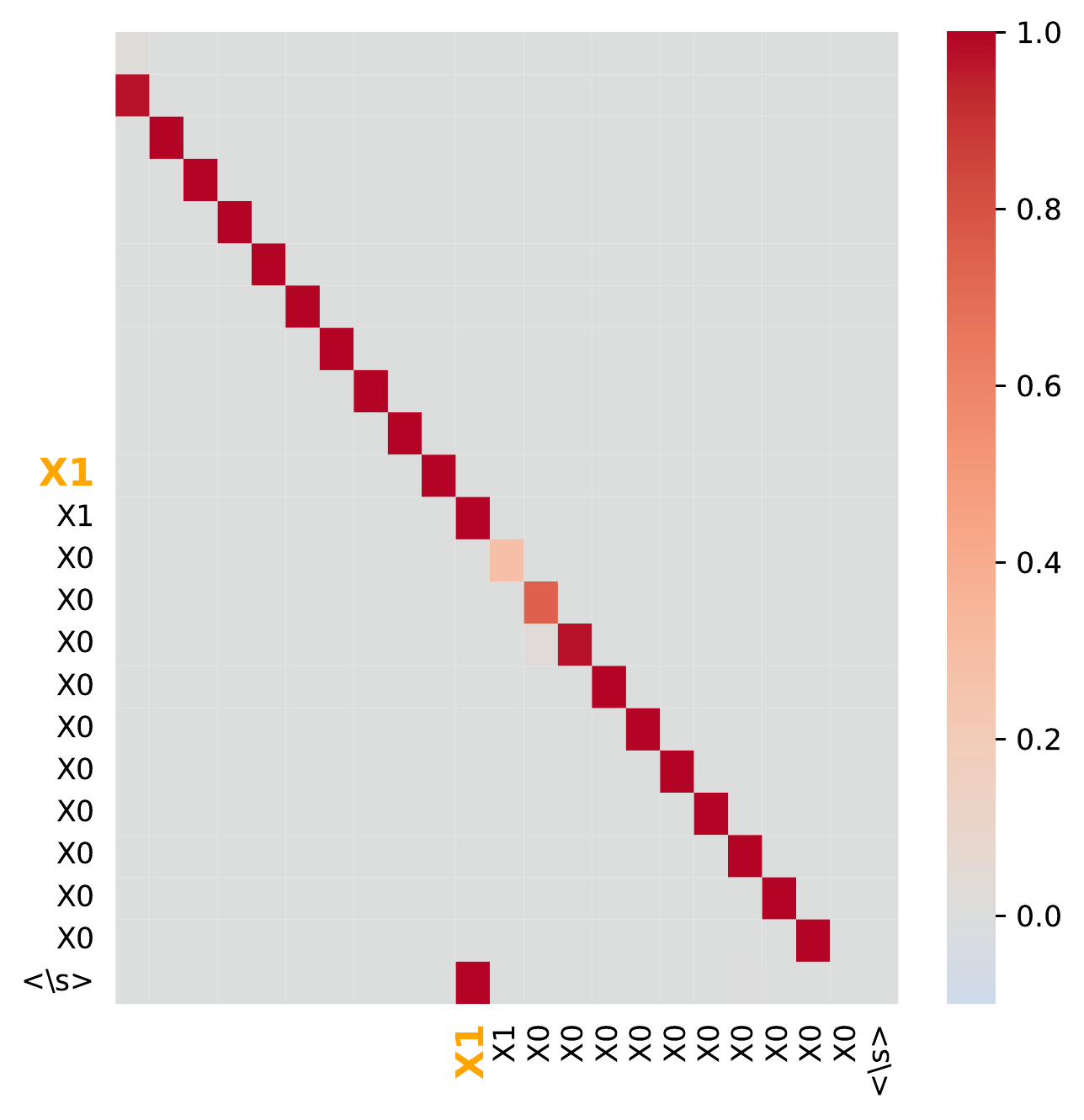}
         \caption{$C=0.5$}
         \label{fig:dec_nat_05}
     \end{subfigure}
     \begin{subfigure}[b]{.15\columnwidth}
         \centering
         \includegraphics[width=\columnwidth]{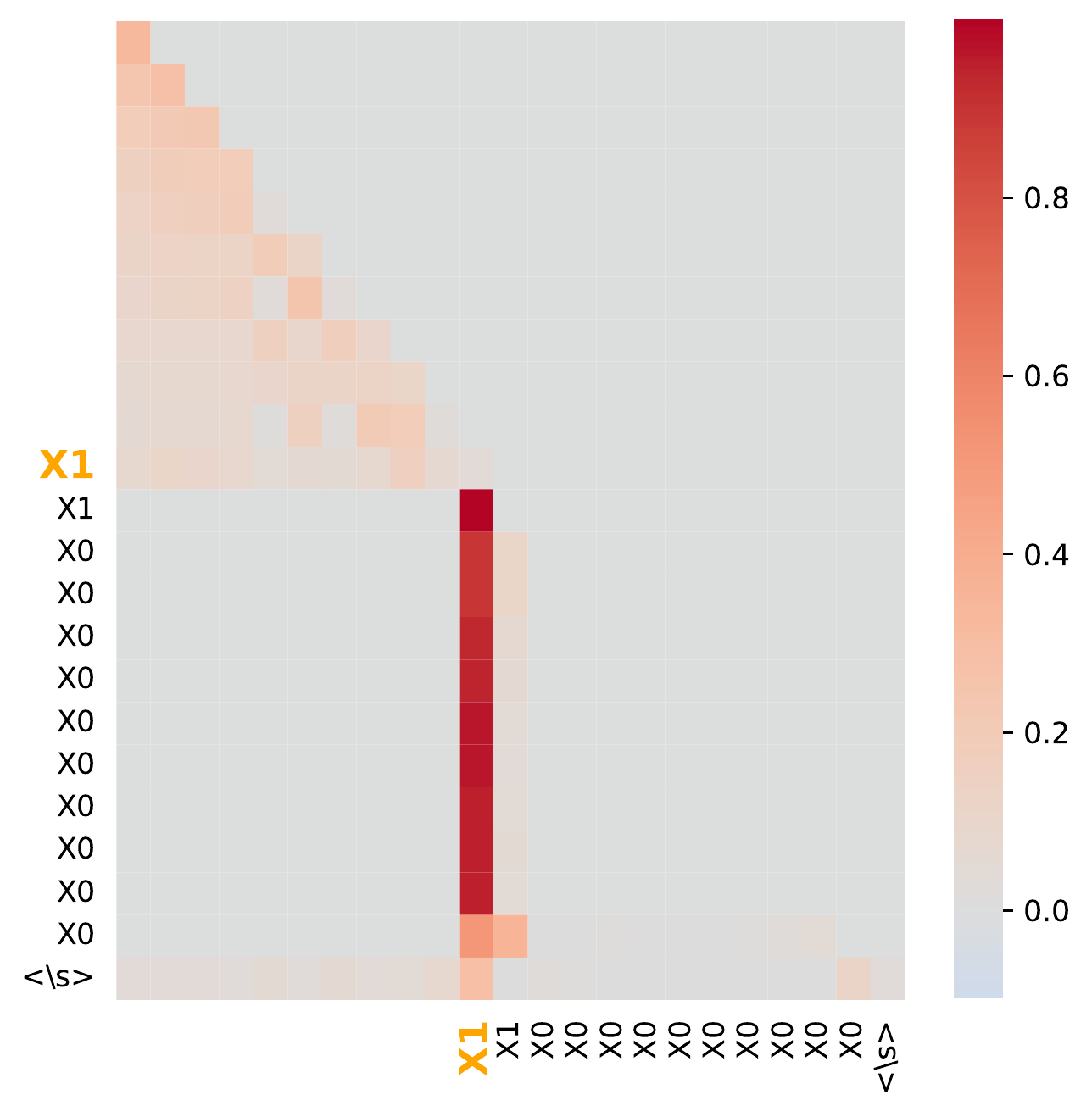}
         \caption{$C=0.7$}
         \label{fig:dec_nat_07}
     \end{subfigure}
     \begin{subfigure}[b]{.15\columnwidth}
         \centering
         \includegraphics[width=\columnwidth]{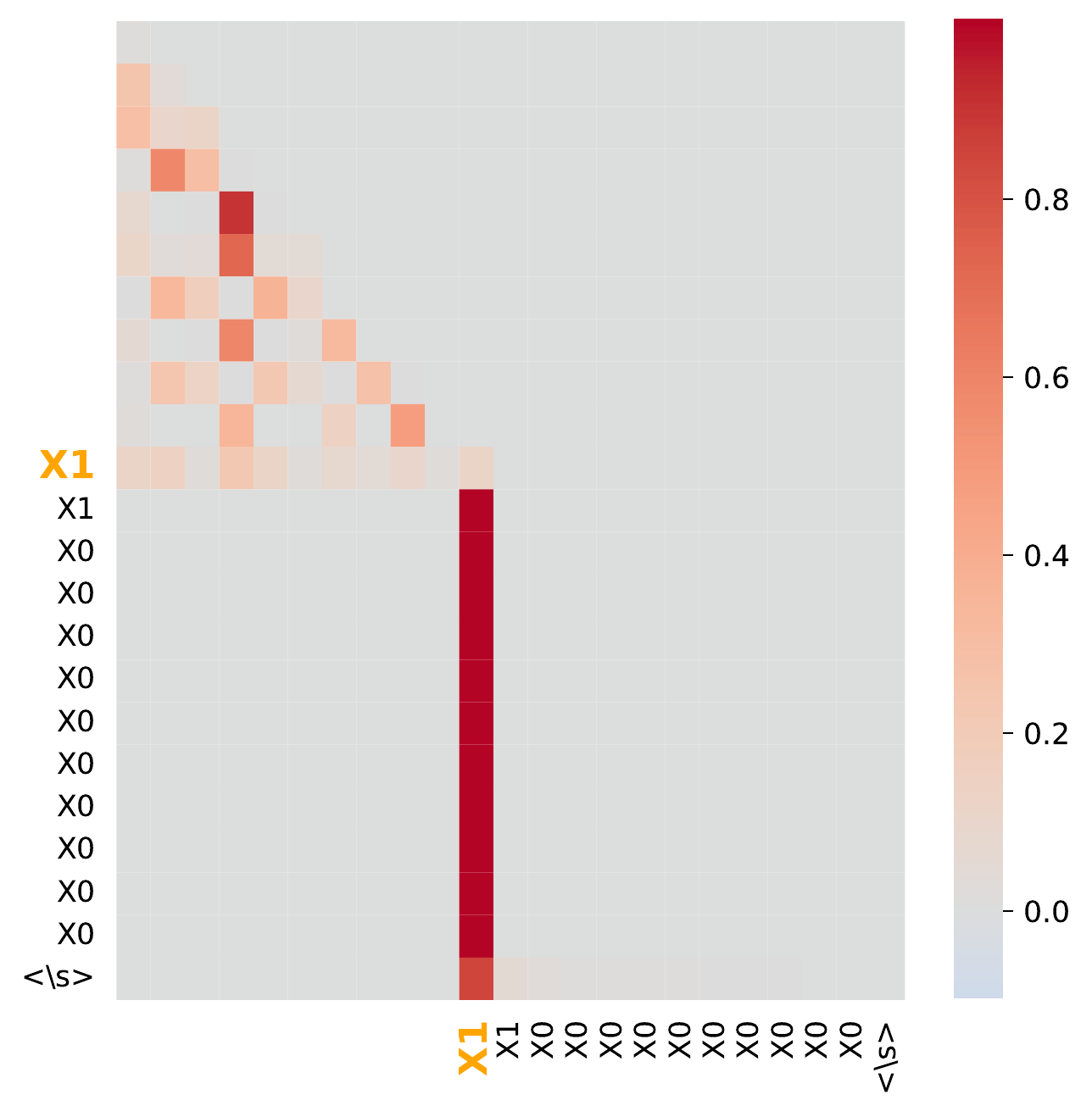}
         \caption{$C=1$}
         \label{fig:dec_nat_1}
     \end{subfigure}
        % \vspace{-2mm}
        \caption{Self-attention of the last decoder layer under different LR factor $C$'s on natural order.}
        \label{fig:lr_rec_head}
        % \vspace{-3mm}
\end{figure*}

Our analysis suggestions that the model may be learning \emph{different algorithms} under different learning rates, and that this can have implications for out-of-domain generalization.
\iffalse
In our study, we observed that the learning rate affects the attention patterns.
%By identifying this relationship between learning %rates and attention pattern,
From this, we gain a better understanding of the factors that contribute to length-generalization problems in transformers when solving recursive tasks. 
We hope this will help make sense of previous studies that have shown
the impacts of hypeparameter choice on generalization for numerical, programming, and reasoning tasks~\cite{anil2022exploring, First22icse}.
\fi
We followed the original transformer learning rate scheduling scheme~\cite{allyouneed}: $
\alpha = C*d^{\frac{1}{2}}*\min\{s^{-\frac{1}{2}},s*S{_w^{-\frac{3}{2}}}\} 
$
where $d$ is the embedding size of the model, $s$ is the current number of update steps, $S_w$ is the predefined warmup step number, and $C$ is the constant controlling the magnitude. 
%We experimented with both the Adam and AdamW optimizers.
To our surprise, we observed a significant difference in attention patterns when trained with different values of $C$ on the natural order task. 
The ``recursion head'' phenomenon emerged when we held $C$ close to 1, while it disappeared when the learning rate was small. As the learning rate grew, the model began to specialize one head into a recursion head gradually, as shown in Figure~\ref{fig:lr_rec_head}. In fact, smaller LR models learn weaker notion of executing the algorithm in Section~\ref{sec:perturb} for longer sequences, as shown in Figures~\ref{fig:perf_perturb_sor} and~\ref{fig:perf_perturb_eos}. Also, when further constraining the recursion depths required to compute the successor during training (Figure~\ref{fig:perf_nat_depth3}), the model trained on a small LR sees a steeper drop in test performance while still attaining near-perfect training accuracy. However, for the reverse order, such a depth constraint does not severely affect the model's performance on either learning rate (Figure~\ref{fig:more_depth} in Appendix~\ref{app:reconstruct}).

% As shown in Figure \ref{fig:nat_perf_depth}, the model is prone to failing in the maximum possible recursion depths for each test group. This is because these cases are constructed by applying consecutive \lstinline{X1} operators immediately after the base case \lstinline{XO}, but the model with large LR learns the shortcut of identifying the location before recursion starts and generates an \lstinline{X1} followed by \lstinline{XO}'s. In contrast, the correct answer should be applying \lstinline{XO}'s immediately after \lstinline{01}. In these cases, the shortcut is not applicable. The model fails on these cases $\mathbf{100\%}$.  

\iffalse
\subsubsection{Step-wise unrolling}
Moved to appendix - delay 1 week
\fi

\subsection{Tree Traversal}
\label{sec:evaltree}

We applied counter-factual patching to identify the components that are the most crucial to models' performance.
We focused on analyzing the model's behavior as it performed tree traversal subtasks we designed (see Appendix~\ref{app:curt} for details). These subtasks involved different stages such as copying initiation, inserting root nodes, and resuming copying after insertion. 
Below are our findings:
\begin{enumerate}
    \item \textbf{Full traversals are hard; tricks are not} (Section~\ref{sec:tricks}). Models poorly learn full traversal, but pick up tricks if possible.
    % We found that the transformer uses simple heuristics to perform the task, including possible parenthetical depth counting and bracket closure tracking. These worked for full pre-order traversal, but the model was unable to learn full in-order traversal.
    % \item \textbf{Transformers can learn to perform step-wise reductions.} (Section~\ref{sec:lr}). The model learns an algorithm specialized for a certain number of reduction steps to perform reduction on unseen tree structures with seen tree depths. 
    \item \textbf{Models learn simple parenthesis and bracket tracking rules to perform reduction} (Section~\ref{sec:parens}). Models learn to track depth this way.
    \item \textbf{Models learn depth-specific tricks} (Section~\ref{sec:depth}). Models find depth-specific shortcuts during reduction, resulting in better performance for certain recursive depths.
    % \item \textbf{Counter-factual patching reveals the key components of the algorithmic circuit} (Section~\ref{sec:perturb}). We are able to reverse engineer how the transformer performs the task by replacing neural activations from counterfactual inference runs and examining attention patterns.
\end{enumerate}
\begin{figure}
\begin{minipage}{.35\textwidth}
 \centering
     \begin{subfigure}[b]{\columnwidth}
         \includegraphics[width=\columnwidth]{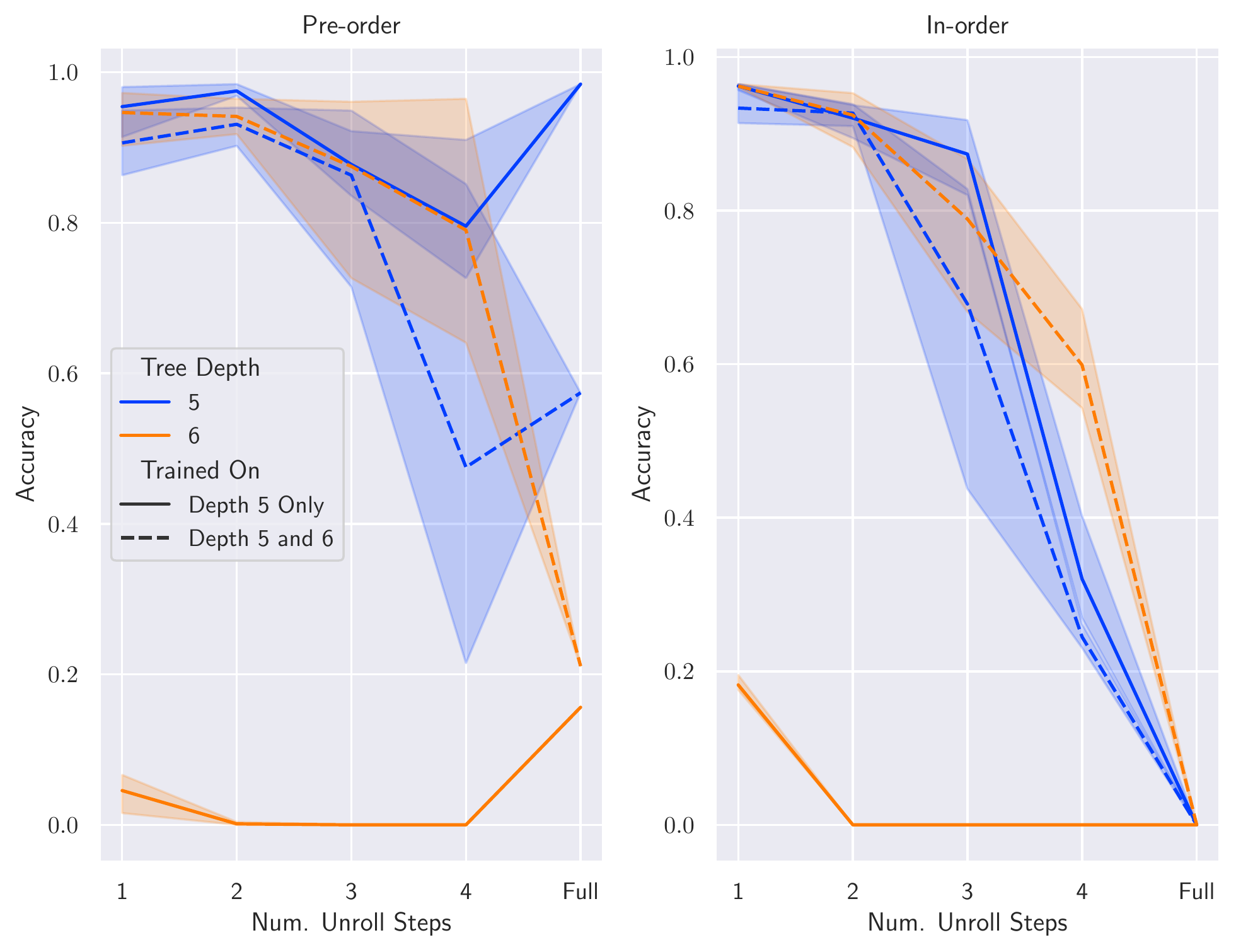}
     \end{subfigure}
        \caption{Accuracy of tree traversal.}
        \label{fig:tree_acc}
\end{minipage}
\begin{minipage}{.6\textwidth}
        \centering
        \begin{subfigure}[b]{\columnwidth}
         % \hspace{2mm}
         \centering
         \includegraphics[width=\columnwidth]{recursive_func_transformer_neurips/plots/attn_traversal/cross_attn_layer_1_preorder_full.pdf}
         \subcaption{Pre-order full traversal.}
         \label{fig:traverse6}
     \end{subfigure} 
            \begin{subfigure}[b]{\columnwidth}
         \centering
         \includegraphics[width=\columnwidth]{recursive_func_transformer_neurips/plots/attn_traversal/1-cross_attn_layer_0_inorder.pdf}
         \subcaption{In-order reduce.}
         \label{fig:traverse1}
     \end{subfigure}
            \caption{Traversal Cross-attention. We looked at snapshots of attention at a particular time step as the decoding proceeds.}
\end{minipage}
\end{figure}
\subsubsection{Full Traversals are Hard; Tricks are Not}
\label{sec:tricks}
     % \begin{subfigure}[b]{.18\columnwidth}
     %     \centering
     %     \includegraphics[width=\columnwidth]{recursive_func_transformer_neurips/plots/attn_traversal/1-cross_attn_layer_0_inorder.pdf}
     %     \caption{Layer 0 cross-attention during in-order task. The model attends to distant parens and EMPTY tokens as well as the next tokens to be copied.}
     %     \label{fig:traverse1}
     % \end{subfigure}
As shown in Figure~\ref{fig:tree_acc}, models can perform full preorder traversals on unseen trees, but fail completely for inorder traversals. Examining the attention behavior of the model, we observed that the model primarily focuses on numerical values and disregards brackets and EMPTY tokens in preorder traversals, as observed through its cross-attention shown in Figure~\ref{fig:traverse6}.
We hypothesize that there is no clear shortcut for sequence models to perform inorder traversals without using a stack. Unlike preorder traversals, which can be done linearly, inorder traversals demand ``understanding'' and capturing recursive relationships between nodes.

\subsubsection{Models Learn Simple Parenthesis and Bracket Tracking Rules to Perform Reduction}
\label{sec:parens}
In addition to heads that write-out node values, those heads that are the most impactful for task performance are responsible for tracking brackets and parentheses---that is, attempting to track recursion depth---both looking ahead to future closures and looking back at the existing output. For example, in the preorder traversal task, among the four cross-attention heads, those at Layer 0 displayed a clear separation of tasks where one head looked ahead in the encoded sequence and attended most to forward parenthesis, brackets, and EMPTY tokens, while the other attention head tended to attend to the encoder sequence tokens in a fairly linear fashion. Depending on the task, closure or opening of brackets acted as signals for the network to change its behavior. In steps when behavior change was required (e.g., completing the copy of a subtree and inserting a non-consecutive symbol or node from the encoder input), we see that the attention heads pay particular attention to the brackets and symbols. For example, decoder self-attention heads will attend to a previous UNROLL symbol when determining whether an EMPTY token should be copied or omitted.

\subsubsection{Models Learn Depth-Specific Tricks}
\label{sec:depth}
Our observations showed that models learned specific tricks for certain depths. For example, for simple two-step reductions in the inorder case, the model can simply copy the root node from the beginning of a parenthetical sequence once that subtree has been copied into an UNROLL statement. But in deeper trees with three reductions, the model needs to track the difference between parent nodes and the base root node. As such, we observed that when performing these deeper reductions the model relies on decoder self-attention and will attend to completed UNROLL phrases which we use to symbolize application of one-step reduction on the subtree inside of it (See Section~\ref{sec:tree} and the appendix), composing this input with cross-attention to the parentheses and key parent nodes---a phenomenon we did not see for more shallow reductions. Conceivably, decoder cross-attention heads could use this as a signal to attend to and copy the first node prior to the initial node inside the UNROLL statement, similar to the induction heads found in GPT-2 \cite{elhage2021mathematical}.

%% file: related.tex
\section{Related Work}

\textbf{Understanding Transformers} $\phantom{k}$ Work on understanding the underlying mechanisms
of transformers spans many angles,
from categorizing computational capabilities~\cite{nn_chomsky,zhang2022pointer,lego,self_attn_language} by measuring performance on synthetic tasks,
to deriving theoretical arguments~\cite{transformer_shortcut_aut,transformer_first_order,attn_turing,nn_turing,theoretical_lim},
to analyzing the functionalities of parameters~\cite{linear_layer_memory},
to reverse engineering learned algorithms from the perspective of mechanistic interpretability~\cite{mechanic_intepret_grokking,reverse_engineering_group_op}. 
%Also, some probe. () Yet, 
%These research works have provided valuable insights into understanding the transformer architecture's strengths and limitations, which, in turn, can guide modifications and improvements to the architecture. 
Our work in particular focuses on the ways in which
transformer models fail on the very tasks they are
trained for, with an emphasis on an important class of algorithms that can be modeled by structural recursion.

\textbf{Mechanistic Interpretability} $\phantom{k}$ Our analyses of the algorithms performed by our trained models was inspired by existing work in the relatively new field of mechanistic interpretability, and includes methods such as counterfactual patching\cite{meng2023locating}, circuit tracing\cite{wang2022interpretability}, automatic circuit discovery \cite{conmy2023automated}, and component ablation. In our work we use a  number of these techniques to reverse-engineer the critical components of the model and how they carry out the algorithm that solves the tasks in question.

\textbf{Program and Proof Synthesis} $\phantom{k}$
The tasks we choose are inspired by work in
program and proof synthesis~\cite{PGL-010, lee2023popl, myth, symlens, ringer2021pldi, ringer2021proof, magaudbertot, chaudhuri2021neurosymbolic}, % {\color{purple} cite more; see lee2023popl for citations}.
and are also an important class of functions
for inductive logic programming~\cite{ilp30}. 
Transformer-based large language models have brought significant progress to program synthesis~\cite{palm, chen2021codex},
but current tools still struggle
to emulate function behavior without prompt engineering techniques like chain of thought prompting~\cite{wei2022chain} or scratchpad reasoning~\cite{nye2021show}. Our work pursues a better understanding of how transformer models fail to represent recursive function behavior. %, with an eye on later improving their performance without prompt engineering techniques. 
We also hope that our work will help open the door to later working making sense of why these prompt engineering techniques may help to begin with.
% Are there non-arxiv sources for CoT and scratchpad? :/

%Our hope is to pave a path to building tools that can approach this task neurally, so that we can build tools for adapting and repairing programs and proofs without the need
%for these symbolic tools. % ugh not happy with this but just trying to wrap something up since deadline is soon

%\paragraph*{Something} {\color{purple} repeat some of the maxim stuff here}

%% file: conclusions.tex
\section{Conclusions, Limitations, and Future Work}
\label{sec:conclusions}

Transformer models can approximate the behavior of
important structurally recursive functions,
but the shortcuts they learn fall short.
We have shown that, by reconstructing the
algorithms corresponding to these shortcuts,
it is possible to understand and even predict
\textit{how} and \textit{why} they fall short. In this work, our main focus was on toy transformer models trained from scratch, while we deferred the understanding of large pretrained language models to future work.
One next step is to use similar analyses
to understand the shortcomings of large pretrained language models on programming and reasoning tasks, and to make sense of why tricks like chain of thought reasoning and scratchpadding help on those tasks. Beyond that, we are excited to use our
newfound understanding to drive future improvements to training and prompting techniques, neural architectures, and evaluation methodologies.

% \section*{Acknowledgement}
% We would like to thank Chris Olah, Jason Rute, Nadia Polikarpova and Neel Nanda for their valuable inputs.